\documentclass[journal,onecolumn]{IEEEtran}

\usepackage[colorlinks=true, linkcolor=blue, citecolor=blue, urlcolor=blue]{hyperref}

\usepackage{cite}

\usepackage{amsmath} 
\usepackage{amsfonts}
\usepackage{amssymb}
\usepackage{bm}

\usepackage{svg}

\usepackage{graphicx} 
\usepackage{caption}  
\usepackage{subcaption}

\usepackage{enumitem}

\usepackage{arydshln}

\usepackage{booktabs}
\usepackage{multirow}

\begin{document}

\title{See It Before You Grab It: Deep Learning-based Rebound Anticipation in Basketball}

\author{%
\begin{tabular}{c@{\hspace{2cm}}c}
Arnau Barrera-Roy &
Albert Clapés \\
Universitat de Barcelona &
Universitat de Barcelona \\
Computer Vision Center &
Computer Vision Center \\
Email: arnau6baroy@gmail.com &
Email: aclapes@ub.edu
\end{tabular}
}

\markboth{}%
{Barrera Roy: Deep Learning-based Rebound Anticipation in Basketball}

\maketitle

\begin{abstract}
Computer vision and video understanding have transformed sports analytics by enabling large-scale, automated analysis of game dynamics from broadcast footage. Despite significant advances in player and ball tracking, pose estimation, action localization, and automatic foul recognition, anticipating actions before they occur in sports videos has received comparatively little attention. This work introduces the task of action anticipation in basketball broadcast videos, focusing on predicting which team will gain possession of the ball following a shot attempt. Rebound prediction presents unique challenges compared to other basketball actions due to multiple players simultaneously contesting in tight spaces, frequent occlusions, and irregular ball bounces off the rim. To benchmark this task, a new self-curated dataset comprising 100,000 basketball video clips, over 300 hours of footage, and more than 2,000 manually annotated rebound events is presented. Two experimental setups for rebound anticipation are proposed. In the offline rebound anticipation setting, video sequences are truncated $\tau_a$ seconds before the action occurs, and the task is to predict the upcoming event. In the online setting, the objective is to determine whether a rebound will occur in the immediate future based on a short, fixed-length video clip, with the key difference that the exact timestamp of the action is unknown beforehand. Comprehensive baseline results are reported using state-of-the-art action anticipation methods, representing the first application of deep learning techniques to basketball rebound prediction. Additionally, two complementary tasks — rebound classification and rebound spotting — are explored, demonstrating that this dataset supports a wide range of video understanding applications in basketball, for which no comparable datasets currently exist. Experimental results highlight both the feasibility and inherent challenges of anticipating rebounds, providing valuable insights into predictive modeling for dynamic multi-agent sports scenarios. By forecasting team possession before rebounds occur, this work enables applications in real-time automated broadcasting and post-game analysis tools to support decision-making. All code is available in this \href{https://github.com/arnalytics/labeled_plays_NBA}{repository}, and the dataset will be made publicly available upon NBA permission approval.

\end{abstract}


\IEEEpeerreviewmaketitle


%
%

\section{Introduction} \label{sec:intro}

\IEEEPARstart{V}{ideo} understanding has emerged as a fundamental area within computer vision, aiming to extract meaningful spatiotemporal information from video streams. Unlike static image analysis, video understanding involves modeling temporal dependencies and dynamic interactions across frames. A variety of tasks fall under this umbrella, such as action recognition, object tracking, action localization, temporal segmentation, captioning, and many others. Among these, one particularly challenging and impactful task is \textit{action anticipation}, which goes beyond recognizing past or ongoing events and instead focus on inferring when and what actions will happen in the future. This predictive capability is particularly valuable in dynamic environments where proactive responses are needed. Applications include surveillance \cite{majhi2025guess}, where potentially harmful behavior can be detected before it occurs; autonomous driving \cite{girase2021loki, liu2020spatiotemporal, rasouli2020pedestrian}, where anticipating pedestrian or vehicle intentions can help avoid accidents; and human-robot interaction \cite{german_robots}, where systems must respond promptly and naturally to user behavior.

In recent years, the intersection between video understanding and sports analytics has received increasing attention. Datasets and challenges such as SoccerNet \cite{soccernet2025challenge} have driven progress in tasks such as player and ball tracking, event detection, and action spotting. In contrast, action anticipation has received comparatively little attention in the sports domain, with only a few works addressing it directly \cite{bball_predictions, bball_waterpolo_predictions, wei2014forecasting, dalal2025fantra}, despite its potential to provide valuable strategic insights and enhance real-time decision-making. The inherent uncertainty and rapid dynamics of human behavior in sports make anticipation an especially challenging problem. In contrast to more scripted domains, such as cooking or instructional videos — where future steps typically follow a predefined sequence — sports are characterized by open-ended interactions among multiple agents, evolving strategies, and constantly changing external conditions. In the case of team sports, this complexity is further amplified, since individual behavior often emerges from collective decision-making and coordinated strategies, rather than from isolated actions. This lack of determinism means that even minor contextual changes, such as the positioning of a player or a slight shift in the trajectory of the ball, can drastically alter the course of play, making reliable anticipation substantially more challenging. Indeed, experimental results from expert human observers, conducted as part of this work, reinforce the idea that accurately predicting sports events is challenging in such non-scripted scenarios.

\begin{figure}[htbp]
    \centering    
    \begin{subfigure}[b]{0.98\textwidth}
        \centering
        \includegraphics[width=\textwidth]{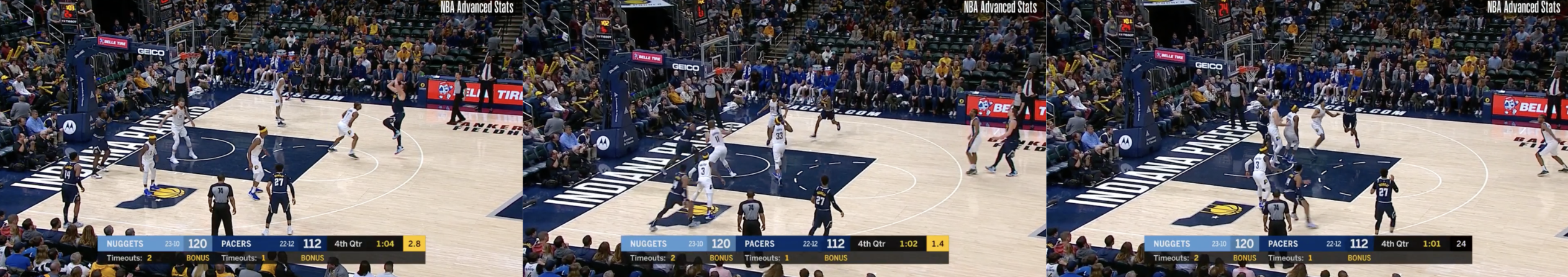}
        \caption{Offensive rebound: team \textit{dark blue} shoots the ball and they regain the possession after the shot is missed.}
        \label{fig:OREB_example}
    \end{subfigure}

    \vspace{1em}

    \begin{subfigure}[b]{0.98\textwidth}
        \centering
        \includegraphics[width=\textwidth]{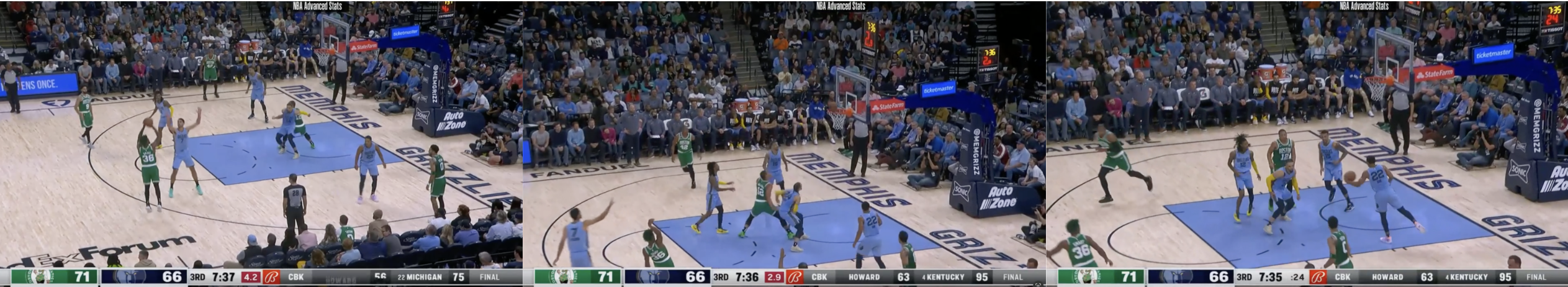}
        \caption{Defensive rebound: team \textit{green} shoots the ball and the opposing team \textit{light blue} gets the ball after the shot is missed.}
        \label{fig:DREB_example}
    \end{subfigure}
    
    \caption{Example sequences of the two studied human actions: \textit{a)} defensive rebound and \textit{b)} offensive rebound.}
    \label{fig:oreb_and_dreb_example}
\end{figure}

In particular, this work focuses on the task of action anticipation within the domain of basketball. Specifically, aiming to predict which \textbf{team} will gain possession of the ball after a shot is attempted at the basket. In basketball, a rebound is defined as the action in which a player gains control of the ball following a missed shot. There are two possible rebound categories. If the attacking team — the one that takes the shot — regains possession, the action is classified as an Offensive Rebound (OREB; see Figure~\ref{fig:OREB_example}). Otherwise, it is a Defensive Rebound (DREB; see Figure~\ref{fig:DREB_example}), meaning that the defending team secures the ball. Predicting rebounds is particularly challenging compared to other actions such as shots or passes. Rebounding situations typically involve multiple players simultaneously contesting the ball within a very tight space, which often results in partial or full occlusions and subtle factors such as positioning, timing, and body momentum critically influence the outcome. In addition, the irregular bounces of the ball off the rim introduce further uncertainty, as small variations in the shot’s trajectory or power can drastically alter the direction in which the ball rebounds. Moreover, the high intra-class variability — particularly in Offensive Rebounds — further increases the complexity of the task.

Two main setups will be considered for this task, both falling under the broader umbrella of action anticipation, since in both cases the objective is to predict the future given the past temporal context: Offline Rebound Anticipation and Online Rebound Anticipation. The key difference lies in how this temporal context is constructed. In the offline setting, the video sequence is truncated exactly $\tau_a$ seconds before the action occurs, and the task is to predict the upcoming action. In the online setting, the exact timestamp of the action is unknown in advance. Instead, the video is divided into fixed-length clips, and the goal is to anticipate whether an action will take place within an anticipation window, which corresponds to the immediate future frames following the observed context.

\begin{figure}[htbp]
    \centering
    \includegraphics[width=1.0\textwidth]{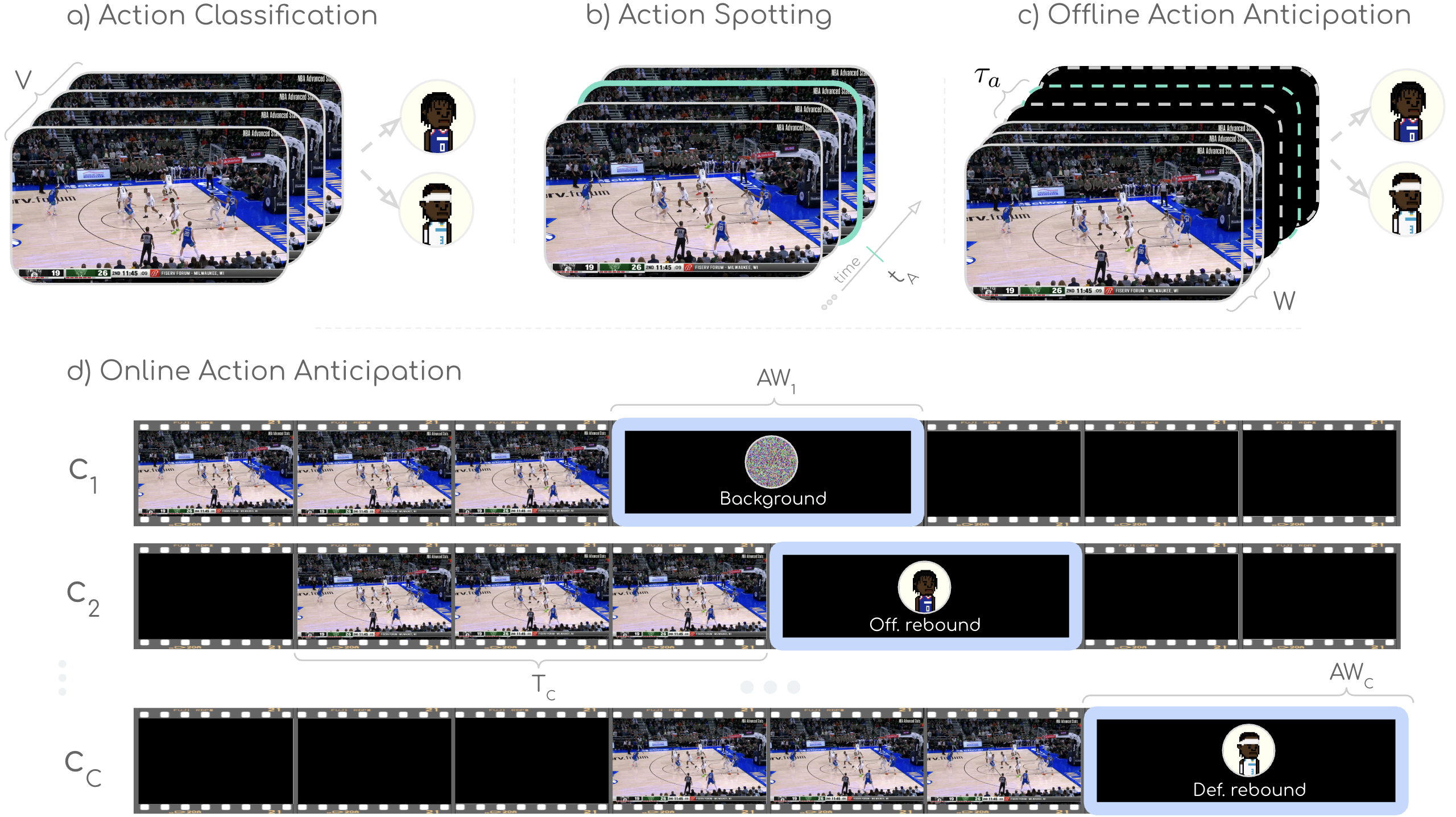}
    \vspace{0.07cm}
    \caption{Illustration of the different tasks solved in this project: \textit{a) Action Classification}, \textit{b) Action Spotting}, \textit{c) Offline Action Anticipation} and \textit{d) Online Action Anticipation}. The notation used in this figure ($V, t_A, \tau_a,$ etc.) will be later explained in detail in Section \ref{sec:problem_def}.}
    \label{fig:tasks}
\end{figure}

Finally, given the characteristics and limitations of the dataset, it is useful to also consider two auxiliary tasks that support the main objectives. The first is action classification, which consists of assigning each video sequence to one of several predefined categories depending on the action being performed. The second is action spotting, which involves identifying the exact frame in which an action takes place. It is worth noting that, unlike anticipation tasks, these auxiliary tasks have access to the frames where the actions actually occur. A schematic overview of both the main and auxiliary tasks is provided in Figure~\ref{fig:tasks}.

The main contributions of this work include the creation of a self-curated dataset for various video understanding tasks in the domain of basketball, comprising 100,000 videos and over 300 hours of footage\footnote{The dataset is not publicly available as explicit permission from the NBA has not yet been granted; the request has already been submitted.}, along with more than 2,000 manually annotated events. Additionally, this work applies action anticipation techniques to \textbf{rebounds}, a particularly challenging action to predict due to the number of players involved and, to the best of the authors' knowledge, one that has not previously been addressed using only video data. Finally, baseline results are provided on this dataset and benchmark, employing two different methods — one of which represents the current state-of-the-art — for action anticipation in sports.

\vspace{1em}
This manuscript is organized as follows: Section~\ref{sec:SOTA} introduces the general architectural principles of current state-of-the-art action anticipation models, followed by a more detailed analysis of those approaches specifically applied to sports scenarios. Section~\ref{sec:problem_def} provides a formal definition of the target tasks along with the evaluation metrics used for each of them. Section~\ref{sec:dataset} describes the dataset employed in this work, including the data collection process, the manual annotation protocol, and the splits defined for each of the video understanding tasks addressed: action anticipation, action classification, and action spotting. Section~\ref{sec:method} details the main method used for online action anticipation, which, among the two action anticipation setups considered, is the most widely studied in the literature. Section~\ref{sec:experiments} presents the experiments conducted to analyze the impact of several factors, such as the effect of varying the temporal context provided to the model or the influence of increasing the number of training samples on overall performance, among others. It also includes a comparison with the performance of human experts and an attempt to interpret the basis of the model's predictions. Section~\ref{sec:implementation_details} outlines the hyperparameters and implementation details of the models used in each one of the experiments. Section~\ref{sec:results} reports and analyzes the results obtained. Finally, Section~\ref{sec:limitations} discusses the limitations of this work and possible directions for future research, while Section~\ref{sec:conclusions} summarizes the main findings.

%
%

\section{State of the Art} \label{sec:SOTA}

This section begins by providing a simplified taxonomy of the action anticipation task (\ref{sec:SOTA_taxonomy}), in order to contextualize the setup adopted in this work. Next, a comprehensive overview of the design principles and architectural components that characterize action anticipation models is presented (\ref{sec:SOTA_architecture}). Additionally, some background (\ref{sec:SOTA_background}) is provided on the action anticipation task specifically in the sports domain, which is the context of the present work. Finally, some of the most widely used action anticipation datasets are introduced (\ref{sec:SOTA_datasets}).

\subsection{Taxonomy of action anticipation}
\label{sec:SOTA_taxonomy}
There are multiple ways to categorize the different approaches to the action anticipation task. Here, a simplified taxonomy is presented, which allows us to justify some of the design decisions made in addressing this task. 

First, action anticipation tasks can be categorized based on the temporal horizon of the prediction. In \textbf{short-term action anticipation}~\cite{dalal2025fantra, kim2021temporally}, models predict actions within a small future window, typically only a few seconds ahead of the observed context. In contrast, \textbf{long-term action anticipation}~\cite{abu2021long, gammulle2019forecasting, mascaro2023intention} aims to predict the sequence of future actions, potentially looking several minutes ahead. In general, the sports domain typically focuses on short-term anticipation due to the high variability and uncertainty inherent in these activities, as is also the case in this work.

Second, action anticipation tasks can also be categorized based on the type of prediction produced. \textbf{Non-temporal anticipation} approaches~\cite{kim2021temporally,gammulle2019forecasting, mascaro2023intention} predict only the action class (or sequence of classes) within the anticipation horizon, without providing any temporal information about when the action will occur. \textbf{Temporal anticipation} approaches ~\cite{bball_predictions, abu2021long, dalal2025fantra} predict not only the class but also some form of temporal information, such as a time-to-event estimate or a temporal interval indicating \textit{when} the action is expected to take place, which increases the complexity of the task. In the present work, the first approach is adopted, as the available data for the anticipation task is relatively limited compared to other datasets in the literature that address temporal anticipation, which prevents the use of the most challenging setup.

\subsection{General architectural principles} \label{sec:SOTA_architecture}

State-of-the-art methods for action anticipation tend to follow a similar architectural structure: a backbone to extract features from the input data, a neck to further process these features — often enhancing their discriminability for the specific anticipation task and potentially introducing temporal modeling if the backbone does not — and finally a head, which is responsible for producing the final prediction.

The \textbf{backbone} is responsible for extracting features from video data. Depending on the type of backbone used, the resulting features can be per-frame, where each feature vector encodes information from a single frame only — without incorporating any temporal context (i.e., information from preceding or subsequent frames) — or they can include some form of temporal modeling, where each feature vector represents information aggregated from multiple frames. The most commonly used backbones in video understanding are 2D CNNs, 3D CNNs, and Vision Transformers.

2D CNNs~\cite{gammulle2019forecasting, kim2021temporally}, such as \textit{ResNet} \cite{ResNet} or \textit{EfficientNet} \cite{EfficientNet}, are typically the most computationally efficient, but they lack temporal modeling capabilities. Other 2D CNN-based architectures, like \textit{TSN} \cite{TSN}, introduce temporal modeling at a coarse level by sampling frames from different temporal segments and aggregating their features. However, in all these cases, the backbone still processes each frame independently, meaning that either no temporal dynamics are captured or they are modeled only at a very coarse granularity. 

On the other hand, 3D CNNs~\cite{mascaro2023intention, bball_predictions}, such as \textit{SlowFast}~\cite{SlowFast} or \textit{X3D}~\cite{feichtenhofer2020x3d}, explicitly introduce temporal modeling by using 3D convolutional kernels that operate simultaneously on the spatial dimensions of the frames (height $H$ and width $W$) and on the temporal dimension $T$. This allows the model to capture local motion patterns and temporal dependencies directly across multiple consecutive frames. However, this comes at the cost of significantly higher computational and memory requirements, as 3D convolutions involve an additional dimension compared to their 2D counterparts. This not only increases the number of parameters and floating-point operations (FLOPs), but also leads to higher GPU memory usage and slower training and inference times, especially for long video sequences.

Another increasingly popular family of backbones for video understanding are Transformer-based architectures~\cite{gong2022futur}. These models rely on the self-attention mechanism, which enables them to model long-range dependencies across spatial and temporal dimensions more flexibly than convolutional networks. Unlike CNNs, which have a fixed receptive field determined by kernel size and network depth, Transformers can attend to any part of the input sequence, making them particularly effective for capturing global context and complex temporal dynamics at different scales. However, these benefits come at a computational cost. Transformer-based models typically require large amounts of data to train effectively, are more memory-intensive due to the quadratic complexity of self-attention with respect to sequence length, and often require pre-training on large datasets to achieve competitive performance. As a result, although they are gaining traction in the literature, they are still less prevalent than CNN-based alternatives, especially in domains with limited training data.

To finish with the backbone, it is important to note that there are two main strategies to train these backbones: end-to-end (E2E) training and pre-training, each with its own advantages. In end-to-end training~\cite{girdhar2021anticipative, zhong2018unsupervised}, the entire model — including the backbone, temporal module, and prediction head — is optimized jointly from scratch (or from a given initialization) using the target task's loss signal. This allows the model to adapt all components specifically to the final objective, often leading to improved performance, especially when sufficient labeled data and computational resources are available. In contrast, pre-training~\cite{mascaro2023intention, kim2021temporally} involves training the backbone separately on a large-scale dataset or a related pretext task — such as action classification in this work — and then using the resulting model to extract features for the target task (e.g., action anticipation or spotting). Pre-trained backbones can significantly reduce training time and memory usage, since fewer parameters need to be updated during the target task training. While end-to-end training generally yields better or comparable performance, it often demands considerably more computational resources — particularly in terms of GPU memory — as well as a larger amount of training data to be effective. This requirements make pre-training a more practical alternative in many scenarios, offering a good balance between efficiency and accuracy when trained on a related domain or task.

In terms of \textbf{neck} architectures, this stage plays a central role in further modeling the temporal dynamics of video sequences, refining the local spatiotemporal (or per-frame) features extracted by the backbone into representations suitable for downstream tasks. Three common architectural families are used to capture temporal dependencies: \textit{Recurrent Neural Networks (RNNs)} \cite{furnari2020rolling, abu2021long, gammulle2019forecasting, kim2021temporally, zhong2018unsupervised}, \textit{Temporal Convolutional Networks (TCNs)}~\cite{abu2021long}, and \textit{Transformers} \cite{yi2021asformer, gong2022futur, girdhar2021anticipative}. RNNs — typically Long Short-Term Memory (LSTM) or Gated Recurrent Unit (GRU) — are sequential by nature and well-suited to modeling ordered data, though they often suffer from vanishing gradients and compounding errors in long sequences, which is particularly problematic for action anticipation. TCNs offer an efficient alternative using 1D convolutions (often dilated) to capture long-term dependencies with parallel computation. Transformers have emerged as the most flexible and widely adopted architecture recently, especially in anticipation tasks, due to their global attention mechanism that can capture both local and long-range dependencies. Transformer encoders are commonly used to process the observed context window, while decoders generate anticipated representations, often using cross-attention and learnable queries.

A critical design consideration across these architectures is the \textit{temporal receptive field}, which defines how far into the past or future the model can effectively observe. To accommodate both short- and long-term dependencies, several multi-scale designs are employed. In multi-scale architectures without temporal downsampling, the temporal resolution is preserved across layers, and receptive fields grow progressively, e.g., via dilated convolutions in TCNs or progressively wider attention windows (or global attention) in Transformers. Alternatively, \textit{encoder-decoder} \cite{xarles2024tdeed, mascaro2023intention, gong2022futur} structures reduce temporal resolution in the encoder (e.g., via pooling or strided convolutions) and restore it in the decoder (e.g., via interpolation or transposed convolutions), using skip connections to preserve information at different temporal resolutions or scale.

Finally, the \textbf{head} of an anticipation model — responsible for generating the final predictions — is typically implemented using multi-layer perceptrons (MLPs) and can follow two main design strategies. \textit{Anchor-based} approaches~\cite{Zhao2024AntGPT, xu2021long} consider predefined temporal positions within the anticipation window, predicting the action occurring at each position. Most short-term anticipation methods, adopt this strategy, focusing on actions within fixed temporal intervals. In contrast, \textit{query-based} approaches~\cite{gong2022futur, girdhar2021anticipative} use a set of learnable representations that are not tied to specific temporal positions, predicting both the action and its precise localization.

In anticipation specifically, some approaches explicitly address temporal uncertainty by modeling action duration probabilistically. For instance, certain models \cite{bball_predictions} predict a Gaussian distribution (mean and standard deviation) over the anticipated action’s occurrence. More recent work \cite{zhong2023diffant, mascaro2023intention} explores generative models such as VAEs or diffusion models to sample multiple plausible futures, acknowledging the inherently stochastic nature of anticipation.

Although it is not strictly an architectural component, it is worth noting that some approaches have explored multimodality to build stronger anticipation models. For example, in~\cite{wang2025multimodallargemodelseffective} they use a multimodal LLM to combine text labels with video footage to predict future actions.

\subsection{Background - Action Anticipation in Sports} \label{sec:SOTA_background}

The main problems studied by the computer vision community in the sports domain are event detection and localization (also referred to as action spotting)~\cite{xarles2024tdeed, tridet}, as well as player and ball detection or tracking~\cite{arbues2019singlecamera,bball_action_detection, andrews2024footyvision} or 3D pose estimation~\cite{Bridgeman_2019_CVPR_Workshops, ludwig2021selfsupervised}. Although addressed to a lesser extent, action anticipation in sports scenarios has also been explored in the literature; this section provides a detailed analysis of prior research on these tasks specifically within the sports domain.

\subsubsection{\textit{“What will Happen Next? Forecasting Player Moves in Sports Videos”} --- Felsen et al., ICCV 2017~\texorpdfstring{\cite{bball_waterpolo_predictions}}{}}
They present a generic framework for forecasting future events in team sports videos directly from visual inputs. Their goal is to anticipate actions such as player movements, ball location, and other events of interest in sports like basketball and water polo. They use an ``overhead representation of the game" derived from 2D player and ball tracking data. From this representation, they extract hand-crafted features (player and ball coordinates and velocities, distances, angles, etc.) and feed them into a random forest. They also evaluate two CNN-based methods: (1) inputting the overhead representation along with the hand-crafted features into a VGG-16, and (2) inputting raw RGB images into a VGG-16. Results show that the random forest outperforms the CNNs, while the Overhead CNN outperforms the Image CNN, indicating that extracting forecasting-relevant features directly from raw images is challenging.

\subsubsection{\textit{“Future Event Prediction: If and When”} --- Neumann et al., CVPR Workshops 2019~\texorpdfstring{\cite{bball_predictions}}{}}

They propose a framework to predict Time to Event (TTE) in sports video sequences, estimating probability distributions to answer whether an event will occur and, if so, when. To handle uncertainty and events that may never happen, the model predicts the complementary cumulative distribution function (CDF) rather than the probability density function, allowing for a natural representation of event likelihoods over time. The approach used is a 3D ResNet-34 backbone with a soft-attention module~\cite{3D_ResNet, SoftAttention} and evaluates several network architectures, with the best-performing model being the Gaussian Mixture Model Heatmap (GMMH), which discretizes the future time window and predicts the likelihood of events at each step. Results show that GMMH significantly outperforms other models, achieving a mean TTE error of 1.42s, while attention map analysis indicates that the network focuses on players and team formations as cues for predicting actions. Predictions on the basketball dataset also show higher uncertainty, reflecting the inherent unpredictability of sports events.

\subsubsection{What to Do and Where to Go Next? Action Prediction in Soccer Using Multimodal Co-Attention Transformer (Goka et al., 2024) \texorpdfstring{\cite{goka2024and}}{}}

The paper proposes a method for predicting both the next action type and its destination in soccer matches, addressing not only ``what will happen next’’ but also ``where it will happen.’’ Although this is not a vision-based model, it is included here to illustrate alternative approaches beyond direct visual inputs. This multimodal approach uses action words (e.g., pass, dribble), along with 2D ball-tracking data and other hand-crafted features (such as the match time when the action started or the angle from the opponent’s goal center, among others). It creates a separate embedding for each data type, which is then fed into the model. The core of the method is a co-attention Transformer, consisting of a Multi-Modal Co-Attention (MCAT) encoder that learns relationships across modalities, followed by a Self-Attention Transformer (SAT) that refines each modality individually. The model outputs the probability distribution over action types and the predicted end position of the next action. 

\subsubsection{Action Anticipation from SoccerNet Football Video Broadcasts (Dalal et al., 2025) \texorpdfstring{\cite{dalal2025fantra}}{}}

Dalal et al.\ propose the \textit{Football
Action ANticipation TRAnsformer (FAANTRA)}, the first method dedicated to action anticipation in football broadcast videos. FAANTRA adapts the state-of-the-art anticipation model FUTR \cite{gong2022futur} to this new domain, introducing architectural changes such as local self-attention in the encoder and an auxiliary segmentation branch. The model follows an encoder-decoder transformer architecture with a RegNetY backbone, and predicts future actions within an anticipation window of 5 or 10 seconds. To enable this task, the authors introduce the SoccerNet Ball Action Anticipation (SN-BAA) dataset, derived from the SoccerNet Ball Action Spotting annotations. FAANTRA is trained with a multi-task loss combining action detection, classification, timestamp regression, and auxiliary action segmentation. Experimental results show that while FAANTRA provides a strong baseline, there remains a significant gap to upper bounds that have access to future information, highlighting the difficulty of the task.

\subsection{Datasets} \label{sec:SOTA_datasets}

A wide range of datasets has been developed for action anticipation, typically adapted from action localization or segmentation datasets and covering diverse scenarios such as daily activities, cooking, or sports. As shown in Table~\ref{tab:unified_action_datasets}, these datasets differ not only in scale and annotation protocols but also in the nature of the domains they represent. In particular, \textit{scripted} domains, such as cooking or instructional videos, follow a predefined sequence of actions constrained by a recipe or tutorial, whereas \textit{non-scripted} domains, such as sports, involve open-ended interactions among multiple agents, evolving strategies, and constantly changing external conditions. This latter setting introduces greater uncertainty and variability, making action anticipation considerably more challenging.

As there are no publicly available datasets for action anticipation in basketball — the \textit{NCAA Basketball} dataset~\cite{bball_action_detection} is no longer accessible — I collected and curated my own dataset, \textit{NBA Rebounds}, to fill this gap. Furthermore, the fast-paced and unpredictable nature of basketball, similar to other sports, introduces challenges compared to datasets where action sequences tend to follow a more scripted pattern. Compared to \textit{SoccerNet BAA}~\cite{dalal2025fantra}, although it contains fewer annotated actions, the \textit{NBA Rebounds} dataset exhibits greater variability in terms of the number of games, players, courts, and uniforms, making it a strong alternative for research in the domain of basketball.

Finally, the NBA Rebounds dataset is divided into two versions. As will be explained later, the original scraped data includes only the class of the action occurring within each video. To adapt the dataset for more complex tasks, such as action anticipation, 2,000 of the original 100,000 samples were manually annotated, also including the timestamp of the action.

\begin{table}[htbp]
\centering
\renewcommand{\arraystretch}{1.2}
\caption{Summary of datasets for action anticipation. $^{\dagger}$Dataset no longer available.}
\begin{tabular}{cccccc}
\hline
\textbf{Dataset} & \textbf{Year} & \textbf{Domain} & \textbf{\#Videos/Duration} & \textbf{\#Classes} & \textbf{Actions type} \\
\hline
50Salads~\cite{stein2013combining} & 2013 & Cooking & 50 / 4.5h & 17 & Scripted \\
Breakfast Actions~\cite{kuehne2014language} & 2014 & Cooking & 1,989 / 77h & 48 & Scripted \\
THUMOS14~\cite{idrees2017thumos} & 2014 & Web/YT & 412 / 20h & 20 & Non-scripted \\
MPII Cooking 2~\cite{rohrbach15ijcv} & 2015 & Cooking & 273 / 27h & 222 & Scripted \\
TVSeries~\cite{de2016TVseries} & 2016 & TV & 27 eps / 16h & 30 & Scripted \\
Charades~\cite{sigurdsson2016hollywood} & 2016 & Daily Activities & 9,848 / 82h & 157 & Scripted \\
NCAA Basketball$^{\dagger}$~\cite{bball_action_detection} & 2016 & Basketball & 257 / 385h & 11 & Non-scripted \\
EGTEA Gaze+~\cite{li2018eye} & 2018 & Cooking & 86 / 28h & 106 & Scripted \\
EpicKitchens-100~\cite{damen2018epicKitchens} & 2021 & Cooking & 100 / 100h & 300 & Non-scripted \\
Ego4D~\cite{grauman2022ego4d} & 2022 & Daily Activities & - / 3,670h & 4756 & Non-scripted \\
Assembly101~\cite{sener2022assembly101} & 2022 & Daily Activities & 4,321 / 513h & 202 & Scripted \\
SoccerNet BAA~\cite{dalal2025fantra} & 2025 & Soccer & 7 games / 10h & 12 & Non-scripted \\
\hdashline
\textbf{NBA rebounds (Action Class)} & \textbf{2025} & \textbf{Basketball} & \textbf{100k} / \textbf{300h} & \textbf{2} & \textbf{Non-scripted} \\
\textbf{NBA rebounds (Action Class+Timestamp)} & \textbf{2025} & \textbf{Basketball} & \textbf{2k} / \textbf{7h} & \textbf{2} & \textbf{Non-scripted} \\
\hline
\end{tabular}
\label{tab:unified_action_datasets}
\end{table}

%
%

\section{Problem Definition}\label{sec:problem_def}

The primary focus of this thesis is human action anticipation, with particular emphasis on rebound prediction in basketball. As introduced in Section~\ref{sec:intro}, two distinct types of action anticipation tasks are considered: \textit{online} and \textit{offline} anticipation, which differ in the amount of prior information available about the action's timestamp. In addition, two secondary or pretext tasks are presented that complement the main objectives, namely action classification and action spotting (see Fig.~\ref{fig:tasks} for a visual overview). This section provides a rigorous definition of all these tasks to ensure conceptual clarity moving forward.

\subsection{Action Classification}

Given a video sequence \( V = \{t_1, t_2, \dots, t_A, \dots, t_{T_V}\} \) composed of \( T_V \) frames, in which a single action \( A \) (the rebound) of the class $y_A$ occurs at frame \( t_A \), the goal of action classification is to determine the \textbf{class} of the action occurring in the sequence. Formally, the model receives as input the hole video $V$ and then it must classify the upcoming action into one of two categories: (1) Offensive rebound or (2) Defensive rebound. In this task, the model has full access to the action frame and its surrounding temporal context. This simplifies the task, as the relevant cues for disambiguating the action are fully observable.

The evaluation metric used for action classification is \textit{accuracy}, which measures the proportion of correctly predicted labels over the total number of samples.

\subsection{Action spotting}

The goal of action spotting is to temporally localize an action within a video sequence, that is, to detect the exact frame in which the action occurs and to assign it the corresponding action class. Formally, the model receives the entire sequence \( V \) (i.e., it has access to the action frame) and predicts both a frame \( \hat{t}_A \), corresponding to the estimated timestamp of the action, and a class label \( \hat{y}_A \), identifying the type of action.

For the evaluation, a prediction is considered correct if it both matches the ground-truth action class and falls within a temporal window of length \(\delta\) centered on the ground-truth timestamp \( t_A \); formally, this requires \( \hat{y}_A = y_A \) and \( \hat{t}_A \in [t_A - \tfrac{\delta}{2}, \, t_A + \tfrac{\delta}{2}] \). Performance is then measured using the mean Average Precision at this tolerance \(\delta\), denoted as mAP@\(\delta\). In this work, \(\delta\) is expressed in seconds, although it can equivalently be defined in frames by considering the frame rate of the sequence \( V \).

\subsection{Offline Action anticipation}
The goal of the action anticipation task is to anticipate which team will secure the rebound before the action takes place, i.e., without access to the action frame or any neighboring frames. In the offline setting, we leverage the ground-truth information about the presence or absence of a rebound action to sample data instances. In this case, the temporal context $W$ is \textit{trimmed}, meaning that it always ends exactly $\tau_a$ frames (or seconds) before the rebound action occurs. As a result, the classification task becomes strictly binary, which constitutes a key difference compared to the online setup, as will be discussed later.

Formally, the model is provided with a partial observation window (also referred to as the temporal context) \( W = \{t_1, t_2, \dots, t_W\} \), with \( W \subset V \), where \( t_W \) is the last frame observed by the model, such that \( t_W = t_A - \tau_a \). Here, \( \tau_a \) is a predefined anticipation time expressed in frames (or seconds), with the constraint \( t_W < t_A < t_{T_V} \). This temporal context serves as the input to the model, which must classify the upcoming action into one of two categories: (1) Offensive rebound or (2) Defensive rebound. The action itself, along with any subsequent frames, remains unobserved during both training and inference. Thus, the task is formulated as a binary classification problem conditioned on partial temporal context, explicitly determined by the anticipation time \( \tau_a \). As in action classification, the evaluation metric used for this task is \textit{accuracy}.

\subsection{Online Action Anticipation}
The task of online anticipation consists in predicting whether an action will occur in the near future based on a short, fixed-length temporal context. Differently from offline action anticipation, in the online scenario we perform an exhaustive sampling of clips within the video without assuming the presence of nearby future actions. This makes the task inherently more challenging, as the context windows are \textit{untrimmed} and uniformly sampled, meaning they may not necessarily precede a rebound, unlike in offline anticipation where the temporal context is \textit{trimmed} and guaranteed to precede the action. As a result, a third class (no rebound) must be introduced to represent these background windows. Since each video $V$ contains only a single rebound, non-action windows are much more frequent than action windows, leading to a class imbalance that further increases the difficulty of the task.

Formally, given a video $V$ in the dataset, we construct a set of consecutive (possibly overlapping) clips \( C = \{c_1, c_2, \dots, c_{|C|}\} \). Each clip $c_i$ has a fixed duration of \( T_C \) frames (or equivalently, seconds), and for each clip we define a corresponding anticipation window $AW_i$ of fixed length \( \Delta \) (see Equation~\ref{eq:online_anticipation_clip_AW}).
\begin{equation} \label{eq:online_anticipation_clip_AW}
    \begin{minipage}{0.45\textwidth}
    \[
    c_i = \{ t_{k}, t_{k+1}, \dots, t_{k+T_C-1} \} \subset V
    \]
    \end{minipage}
    \hfill
    \begin{minipage}{0.45\textwidth}
    \[
    AW_i = \{ t_{k+T_C}, t_{k+T_C+1}, \dots, t_{k+T_C+\Delta-1} \}
    \]
    \end{minipage}
\end{equation}

Here, $k$ is a dummy index indicating that, in principle, the first frame of a clip can start at any position within the video sequence. Intuitively, as illustrated in Fig.~\ref{fig:tasks} d), the clips can be visualized as generated by a sliding window over the whole video. The model observes a single clip \( c_i \) and must predict the type of action (if any) that will occur within \( AW_i \) — which remains unseen — choosing from three categories: (1) Offensive rebound (OREB), (2) Defensive rebound (DREB), or (3) No rebound (or background). This formulation defines a multi-class classification problem conditioned on a limited, local temporal context.

To properly evaluate performance under these uneven conditions, we adopt the \textit{macro-averaged F1-score} as the primary metric. This metric computes the F1-score independently for each class and then averages them, ensuring that all classes — including the dominant background class — are weighted equally, thus providing a fair assessment of the model's capability to anticipate rebounds despite the challenging online scenario.

%
%

\begin{figure}[htbp]
    \centering
    \begin{subfigure}[t]{0.33\textwidth}
        \centering
        \includegraphics[width=\textwidth]{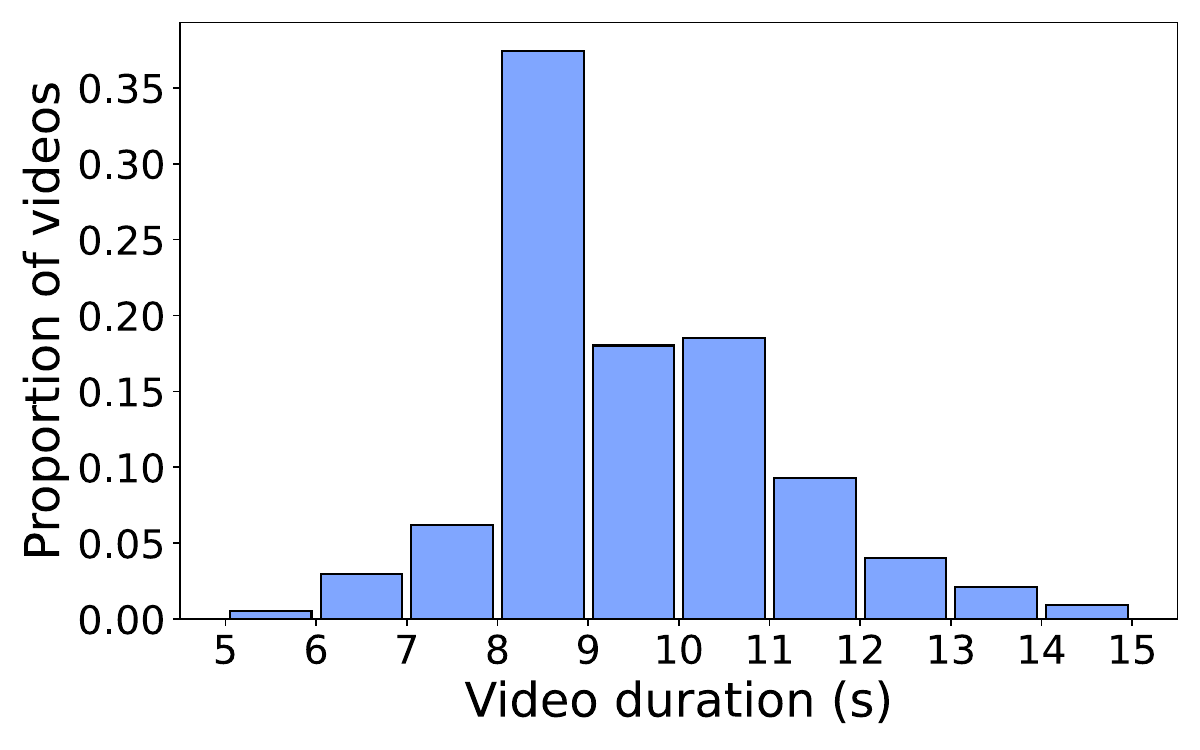}
        \caption{Video durations ditribution}
    \end{subfigure}%
    \hfill
    \begin{subfigure}[t]{0.33\textwidth}
        \centering
        \includegraphics[width=\textwidth]{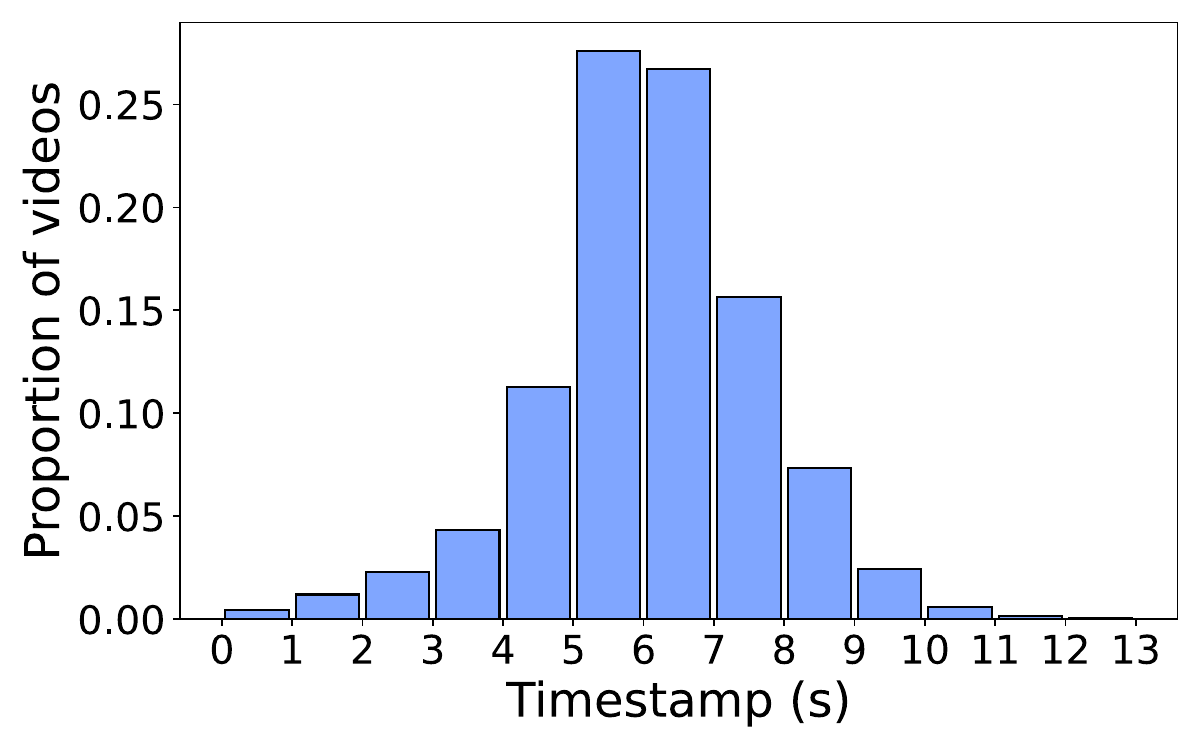}
        \caption{Tiemstamp distribution}
    \end{subfigure}%
    \hfill
    \begin{subfigure}[t]{0.33\textwidth}
        \centering
        \includegraphics[width=\textwidth]{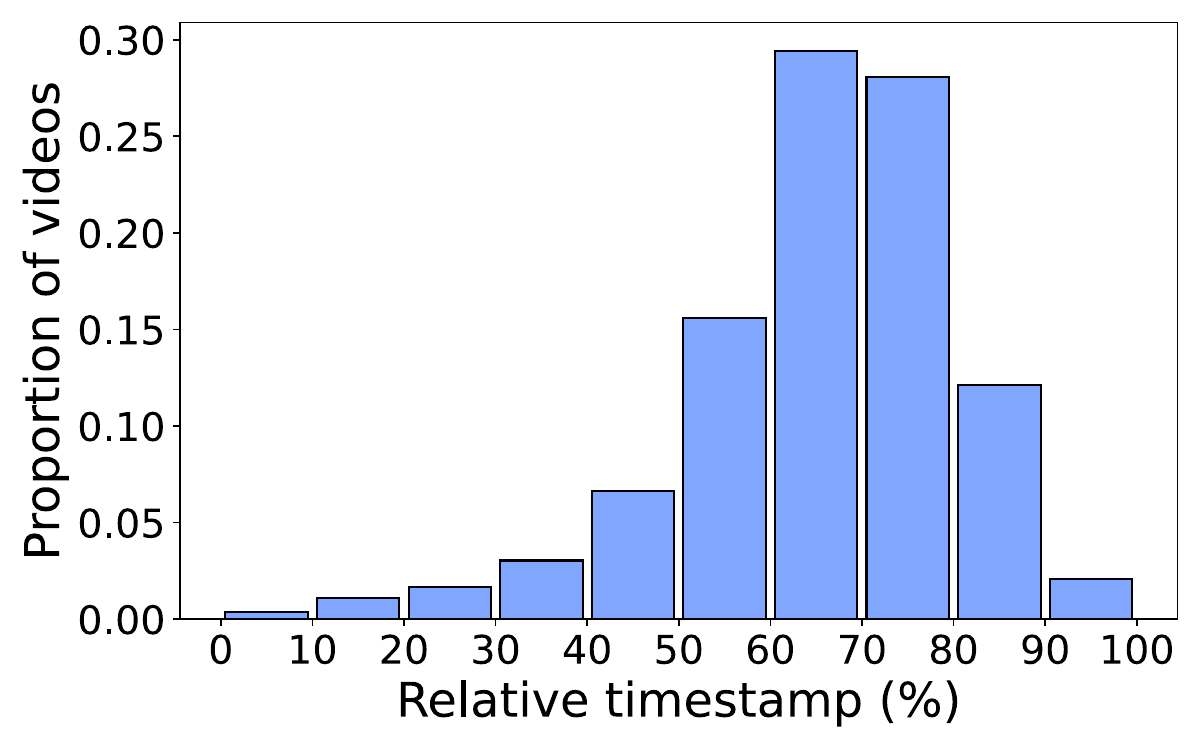}
        \caption{Relative timestamp distribution}
    \end{subfigure}
    
    \caption{Dataset $D_{100\text{K}}$ statistics: a) Histogram showing the distribution of video durations along the hole dataset, b) histogram of the absolute timestamp distribution of the actions across the videos and c) histogram of the action's timestamp relative to the hole video duration.}
    \label{fig:dataset_stats}
\end{figure}

\section{Dataset}\label{sec:dataset}
The dataset used in this work, $D_{100\text{K}}$, was created by myself by building a web crawler and scraping the NBA Stats website \cite{nba_stats}. It consists of over 100,000 videos recorded at a resolution of \(1280 \times 720\) pixels and a frame rate of 60 frames per second, with durations ranging from 5 to 15 seconds (see Figure~\ref{fig:dataset_stats}), spanning 10 different seasons, 30 NBA teams, and more than 800 distinct players. Each video contains exactly one action, corresponding to a single rebound event, and is annotated with a binary class label: either an offensive rebound (OREB) or a defensive rebound (DREB). However, due to resource constraints and training time limitations, we only work with a subset of 2,000 videos, $D_{2\text{K}}$, randomly selected from the complete dataset, $D_{2\text{K}} \subset D_{100\text{K}}$, while ensuring a balanced distribution between the two classes. Note that for each video, we have the label indicating the type of rebound (i.e., the class), but we do not have the timestamp of \emph{when} the rebound occurs within the video. For the action anticipation task, such temporal annotation is essential, so I had to manually annotate this information for the selected videos. After the annotation process, each of the 2,000 samples in $D_{2\text{K}}$ contains two pieces of information: the \emph{class} of the action occurring in the video and the \emph{timestamp} indicating when that action takes place. To ensure transparency and clarify potentially ambiguous cases, the details of the annotation process are specified below.

Rebound timestamps were annotated with frame-level precision (at 60 fps), following the criterion of selecting the first frame in which the ball makes contact with the hand of the player who eventually gains possession. For ambiguous or special cases, the following guidelines were applied:

\begin{itemize}
    \item Blocked shots: This is an exception to the main criterion. If the player who blocks the shot is also the one who secures the rebound, I annotated the moment when the player gains control of the ball, not the frame when the block occurs.

    \item Occluded rebounds: The timestamp was inferred based on visual cues such as the trajectory and speed of the ball in the last visible frame, as well as the body movement of the players.

    \item Mislabeled cases: A total of 84 videos were originally labeled as either OREB or DREB, but did not contain an actual rebound event. These videos were removed, resulting in a final dataset $D_{2\text{K}}$ of 1,916 valid samples. However, for clarity and simplicity, we still refer to the dataset as containing 2,000 videos.
\end{itemize}

As shown, some annotations required subjective judgment. However, since all annotations were carried out by myself, the criteria — though subjective — were applied consistently throughout the dataset. This helped ensure internal consistency across annotations, and as a result, the annotations are coherent and suitable for training and evaluating data-driven models rigorously.

As introduced in Section \ref{sec:intro}, this work addresses several video understanding tasks, each of which will rely on different adaptations of the original dataset $D_{100\text{K}}$.

\subsection{Action Anticipation dataset} 

From the 2,000 samples of $D_{2\text{K}}$, three class-balanced splits are created: 1,500 videos for training ($D^{train}_{2\text{K}}$), 250 for validation ($D^{val}_{2\text{K}}$), and 250 for testing ($D^{test}_{2\text{K}}$). The same splits are used for the offline and online action anticipation tasks, however, note that the pre-processing of the videos will differ between tasks.

\subsection{Action Classification dataset} 

For the action classification task, the models were trained on a larger dataset, $D_{25\text{K}}$, which corresponds to a subset of the full collection $D_{100\text{K}}$. Unlike action spotting and anticipation, classification only requires a video-level label and does not need action timestamp annotations. This is why a much larger dataset can be used for this task. While the entire $D_{100\text{K}}$ dataset was eligible for classification, a subset of 25,000 samples ($D_{25\text{K}}$) was selected to find a good balance between sample variability and feasible training times. This subset was divided into training (23,000 videos, $D^{train}_{25\text{K}}$), validation (1,000 videos, $D^{val}_{25\text{K}}$), and test (1,000 videos, $D^{test}_{25\text{K}}$) splits.

Importantly, $D_{25\text{K}}$ and $D_{2\text{K}}$ were defined as disjoint sets ($D_{25\text{K}} \cap D_{2\text{K}} = \varnothing$) in order to prevent data leakage and guarantee a fair evaluation. This separation is essential because classification models are pre-trained on complete videos, where the action frames are included, while action anticipation models are trained on trimmed clips preceding the action, i.e., excluding the action frame and subsequent frames. The critical point is that the classification model must not be exposed during pre-training to the same action instances that will later appear in the anticipation dataset. In other words, it is not problematic if the classification model sees action frames from other videos, but it would compromise the anticipation setup if it had already seen the action frames from the exact videos used for anticipation. Such leakage would artificially simplify the task and bias the evaluation. By enforcing disjoint datasets, we ensure that no action frames from the anticipation videos are present during classification pre-training, thereby preserving the validity of the experimental protocol.

\subsection{Action Spotting dataset} 

To train the spotting model, we use the same dataset and splits as those employed for action anticipation,  $D_{2\text{K}}$. Once the model is trained, inference is performed on an unlabeled dataset, $D_{U}$, which contains the samples that were not used for training either the spotting model or the classification model. Formally, $D_{U} = D_{100\text{K}} \setminus D_{25\text{K}} \setminus D_{2\text{K}}$, where $\setminus$ denotes the set difference, i.e., the elements that belong to the first set but not to the second. Therefore, $D_{U}$ satisfies $D_{U} \cap (D_{25\text{K}} \cup D_{2\text{K}}) = \varnothing$.

\subsection{Data pre-processing}
This section concludes by detailing the pre-processing steps applied to the video data. Since working with videos is significantly more computationally demanding than images, a set of steps is crucial for ensuring an efficient training process. Additional steps are required for certain tasks, such as action anticipation -- as we will see next. Among task-agnostic pre-processing steps, the following were applied:

\paragraph{Scaling} The first step, common to all tasks, consists of reducing the resolution of the video frames. Depending on the application or dataset, this operation may have critical consequences, as excessive downscaling can result in the loss of small but relevant visual cues — such as the ball in sports scenarios, which often plays a crucial role. Unless otherwise specified, the resolution is reduced to \(455 \times 255\) pixels, approximately three times smaller than the original.

\paragraph{Temporal striding} Another common strategy is to exploit the temporal redundancy present in video data. Raw videos are typically recorded at high frame rates, usually 30 or 60 frames per second (fps). Depending on the nature of the actions in the dataset — for example, fast-moving elements such as cars in Formula 1 or the ball in tennis — it may be necessary to retain the original frame rate. However, in the case of basketball videos, the objects don't move that fast and visual information between consecutive frames is often redundant. For this reason, it is save to work with temporally downsampled videos. The downsampling is controlled by a parameter called \textit{temporal stride}, and the effective frame rate after sampling can be computed as: $fps_{\text{down}} = \frac{fps_{\text{orig}}}{\text{stride}}$. Note that the stride is a parameter that can be easily adjusted depending on the task or experiment, as different temporal resolutions may be better suited for different objectives. For instance, in action classification, a larger stride can be used without significantly affecting performance. In contrast, tasks such as action spotting require higher temporal precision, and therefore a smaller stride is required.

\paragraph{Temporal padding} Another general pre-processing requirement is to pad videos to a consistent length. In this work, padding is applied at the batch level: each video is extended with black frames to match the longest video in its batch. Furthermore, this padding is added to the beginning of the sequence. While this choice is not critical for action classification or spotting, it is important for offline action anticipation as it helps to temporally align the videos.
\paragraph{Data augmentation} As a common strategy in the literature, data augmentation was applied to mitigate overfitting and improve the generalization capabilities of the models. The transformations considered include color jitter, since different teams wear different uniforms and basketball courts can vary in color; horizontal flip, as the game of basketball is symmetric with respect to the horizontal direction of play; a slight Gaussian blur, to simulate the effect of motion blur caused by fast-moving objects; and a slight random resized crop, applied carefully so as not to lose track of important players or the ball, to simulate broadcast camera zooms. All these transformations were applied with a probability of 50\% and were performed video-wise: although the choice of which transformations to apply is random, once selected, they are applied consistently across all frames of a video. This is critical, as applying transformations independently to each frame would disrupt the temporal coherence of the video and make the learning process significantly harder for the model.

The following paragraphs address the \textit{task-specific} pre-processing steps. In general, no additional pre-processing is required for action classification or action spotting beyond those already described. However, for both offline and online action anticipation, additional steps are necessary:

\paragraph{Clip sampling} For offline anticipation, fixed-length clips will be trimmed from the original videos. Each clip has to be trimmed up to the anticipation time so that the action itself is not visible. A potential issue arises when the action occurs within the first $t \leq \tau_a$ frames of the video, as no temporal context frames are then available for the network; these videos are therefore removed. The number of videos removed depends on the selected anticipation time, but the impact is minimal, with less than $5\%$ of the dataset removed for the highest anticipation time.

Finally, for online anticipation, a more elaborate sampling is required. To generate the clips, a clip length must first be defined, and then a sliding window of this length is applied over each video to produce a set of clips (see Figure~\ref{fig:tasks}d). In this case, the sampling process is exhaustive, not using the groundtruth information (class and temporal location of the actions). This introduces clips that do not precede any action and, thus, a no-action (
or background class). Moreover, the clips can be sequential or overlapping. Unless otherwise specified, a 50\% overlap is used to reduce the probability of a rebound occurring right at the beginning or end of a clip.

\section{Method} \label{sec:method}
This section provides a detailed explanation of the main method used to address the online anticipation task, a Transformer-Encoder Anticipation Model (\textit{TEAM}), while the specific implementation details are presented in Section~\ref{sec:implementation_details}. It represents an adaptation of state-of-the-art online anticipation approaches in the sports domain, tailored to the characteristics of the dataset used here. In particular, the original FAANTRA method \cite{dalal2025fantra}, designed for SoccerNet, employed a more complex architecture, which was adapted to suit the specific characteristics and constraints of the current dataset.

The adaptations applied to the model in this work include: first, a single classification objective is used instead of combining the classification loss with an auxiliary segmentation loss, since each clip contains exactly one action, rendering the additional segmentation signal unnecessary. Second, the model focuses solely on predicting the action class, without incorporating a time-to-event estimation, due to the limited number of manually annotated samples available. This restriction helps ensure reliable training and evaluation while simplifying both predictions and the computation of performance metrics. Finally, transformer decoder layers are omitted, as multiple learnable queries are not required for clips containing only one action. This design choice will be further discussed in \ref{sec:method_temporal_module}.

\begin{figure}[htbp]
    \centering
    \includegraphics[width=1.0\textwidth]{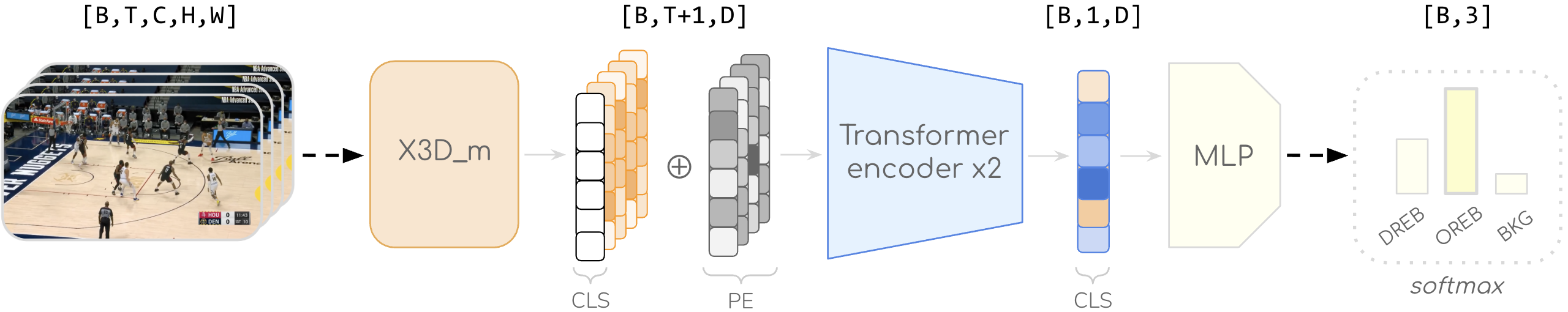}
    \caption{Schematic view of the method \textit{TEAM}, showing the dimensions of the input and output tensors for each module. The input to the model consists of a batch of videos, from which a 3D CNN backbone extracts local spatiotemporal features. Positional encoding (PE) is then added to these features, and a CLS token is appended. This tensor is subsequently passed through a double Transformer encoder layer, which enriches the CLS token. Finally, the enriched CLS token is fed into an MLP to produce the final prediction, represented as a per-class probability.}
    \label{fig:general_model}
\end{figure}

This section is divided into three sub sections, each one dedicated to one of the modules that comprehend the general skeleton of a video understanding model: the backbone to extract visual features, a temporal module to model temporal dependencies and a head to produce the final outputs of the model. In Figure~\ref{fig:general_model}, a detailed diagram is presented, illustrating the main architectural components of the method.

\subsection{Backbone}
After preprocessing, the data is represented as a tensor of dimensions $B \times T \times C \times H \times W$, corresponding to a batch of videos, which is first passed to the backbone. Here, $B$ denotes the batch size, $T$ the temporal resolution of the videos (i.e., the number of frames per video), $C$ the number of color channels, and $H$ and $W$ the spatial dimensions (height and width) of the video frames.

In this work, the backbone employed is the well-established 3D Convolutional Neural Network (X3D) \cite{feichtenhofer2020x3d}, which is widely recognized for its compactness and efficiency. Unless otherwise specified, the intermediate-sized variant  ``X3D\_m" is used, as it offers a favorable trade-off between model size and performance. Notably, it maintains the same number of parameters as the smallest versions (3.79M) while being able to process longer input sequences than its less parametric X3D variants.

As previously mentioned, 3D CNNs extend 2D CNNs by introducing 3D convolutional and pooling kernels that operate simultaneously over the spatial dimensions ($H, W$) and the temporal dimension $T$ of the input video. This design enables the model to capture motion patterns and local temporal dependencies directly across multiple consecutive frames, making them well-suited for video understanding tasks. The use of this model as a baseline is not motivated by its age or performance limitations, but rather by its simplicity and flexibility, given the fact that it can later serve as a backbone for extracting visual features for more complex models, thereby ensuring a fairer comparison among the different approaches.

A particularly important characteristic of X3D models is that they contain several blocks of 3D convolutions that jointly process the three dimensions $H$, $W$, and $C$, while maintaining the temporal resolution across all blocks. Temporal downsampling is only performed at the final block of the network. This aspect is relevant because, depending on the task, it may be desirable to preserve the same temporal resolution as the input, i.e., to obtain one feature vector per input frame. Therefore, unless otherwise specified, when X3D is used as a backbone, the last block will be omitted. In this configuration, the output features extracted for each video have dimensions $B \times T \times D$, where $D$ results from a combination (and compression) of the original $C, H, W$ channels. It is worth noting that an output of this shape could also be obtained using a 2D CNN, which would similarly extract a per-frame feature vector. However, the key difference lies in the temporal modeling: a 3D CNN incorporates temporal context into these feature vectors, so that each one encodes information from multiple frames, something that a standard 2D CNN could not achieve.

\subsection{Temporal Module} \label{sec:method_temporal_module}
As previously discussed, employing a 3D CNN as a backbone enables the extraction of video features that already incorporate a certain degree of temporal modeling, thanks to the 3D convolutions applied across both the spatial dimensions ($H, W$) and the temporal axis ($T$). However, 3D CNNs share a similar inductive bias with 2D CNNs, as they primarily aggregate information that is local in both time and space. Consequently, the temporal modeling performed by the backbone remains more limited to short-range dependencies. To address this limitation, and following previous work \cite{dalal2025fantra, gong2022futur}, a transformer-based temporal module is incorporated to capture long-range temporal dependencies.

The temporal module operates on the sequence of per-frame features extracted by the backbone. Given an input tensor of dimensions $B \times T \times D$, this module models dependencies across the entire temporal context. It consists of an stack of $l_E$ transformer encoder layers, each one comprising a multi-head self-attention (MHSA) block followed by a position-wise feed-forward network (FFN). The self-attention mechanism enables each temporal position to attend to all others, thereby capturing long-range dependencies.

A central component of the design is the learnable CLS (classification) token, originally introduced in BERT \cite{bert} and later adapted to vision tasks in Vision Transformers \cite{ViT}. This token, a parameter of dimensions $1 \times 1 \times D$, is prepended to each sequence, producing an augmented sequence of length $T+1$. Through self-attention, it aggregates information from all temporal positions, providing a global representation of the video sequence. The CLS token is initialized as a vector of zeros and expanded to match the batch size during the forward pass. To encode temporal order, learnable positional embeddings are added, ensuring that the model captures the sequence of frames. After processing through the transformer encoder stack, the representation at the CLS token position is extracted as the final output and passed to the classification head, yielding a tensor of dimensions $B \times D$ that provides a compact summary of the entire temporal sequence.

Finally, the rationale for employing only the encoder layers of the transformer can be detailed as follows. Transformers may consist of an encoder, a decoder, or both. The encoder operates with bidirectional attention, while the decoder relies on masked (causal) attention and is inherently autoregressive, enabling the sequential generation of a variable number of tokens and determining when to stop decoding. In the present case, only a single output is required — the action class — since each clip contains only a single  action to anticipate. This output can be directly obtained from the CLS token introduced into the encoder input. Consequently, the decoder is unnecessary, and the architecture is restricted to the encoder layers alone.

\subsection{Prediction Head}

Following feature refinement by the temporal module, the representation corresponding to the CLS token is passed to the classification head to produce the final predictions. This feature tensor has dimensions $B \times D$ and serves as a compact summary of the entire temporal sequence. The classification head adopted in this work is an anchor-free fully connected MLP, consisting of a single linear layer that maps the CLS token features to an output dimension chosen according to the task at hand. During training, the model is optimized using the cross-entropy loss, while during inference, a softmax function is applied to the output logits to obtain a probability distribution over the possible classes.

\subsection{Baseline model}
The above-mentioned method is the main approach used for online action anticipation. However, it will not be applied to simpler tasks such as action classification or spotting. Instead, the \textit{baseline} model consists solely of the backbone of \textit{TEAM}. Since X3D is already capable of modeling spatio-temporal dynamics, it serves as a strong baseline for comparison with the more complex model. Accordingly, the term \textit{baseline} model refers to the vanilla X3D\_m architecture.

%
%

\section{Experiments} \label{sec:experiments}
This section presents the conducted experiments and outlines the methodology followed. The description begins with the auxiliary tasks — action classification and action spotting (\ref{sec:auxiliary tasks experiments}) — which serve as a preliminary step before addressing the main task of action anticipation. The core experiments are then introduced, starting with offline action anticipation (\ref{sec:experiments offline anticipation}) and concluding with online anticipation (\ref{sec:experiments online anticipation}).

\subsection{Auxiliary tasks}

\subsubsection{Backbone pre-training via Action Classification} \label{sec:auxiliary tasks experiments}

The first step is to pre-train the X3D\_m backbone in rebound classification. The intuition is that the internal representations learned by this model — particularly those in the earlier layers — will be better adapted to the NBA Rebounds dataset than the generic pre-training available in PyTorch on Kinetics-400~\cite{kay2017kineticshumanactionvideo}. This approach takes advantage of the large amount of labeled data available for this particular task, which can be exploited without additional annotation cost. In this setting, complete sequences are provided to the model, which is tasked with performing a binary classification between OREB and DREB. 

Subsequently, this backbone pre-trained on action classification, serves as the initialization for all downstream tasks, including action anticipation in both the online and offline setups. For this reason, exhaustive optimization of the model is not performed for the classification task, since it may rely excessively on information from frames occurring after the action, which would not constitute a useful starting point for anticipation. An ablation study demonstrating the benefits of this pre-training will be presented in Section~\ref{sec:experiments online anticipation}.

\subsubsection{Timestamp Pseudolabeling via Action Spotting} \label{experiment:spotting}

The spotting auxiliary task is particularly useful for obtaining additional automatically labeled samples, hereafter referred to as pseudo-labeled samples, that later can be used in the main anticipation tasks to increase the number of training samples in a semi-automatic way. Specifically, a spotting model can be trained on the 2,000 manually labeled samples already available and subsequently applied to unlabeled data to generate pseudo-labeled samples. During inference, it is common to apply post-processing to the predictions of the spotting model. The specific post-processing techniques used will be detailed later in the implementation section (Section~\ref{sec:implementation_details}), and the effect of each of these steps will be illustrated in the experimental results.

When the spotting model is trained, inference can be performed on the dataset $D_U$, which contains approximately 73,000 samples that were neither manually labeled nor used to train the classification model, thereby ensuring the absence of data leakage. Once inference is completed, a filtering process is applied to the resulting pseudo-labeled samples to improve their reliability, as the employed method is not highly robust. The final filtering procedure is described below; these specific filters were determined after examining the distribution of detections in the inferred dataset, but are presented here for clarity. The filtering criteria are as follows:

\begin{enumerate}[label=(\alph*)]  
    \item Only consider samples were there is a single detection across the whole sequence, in order to be consistent with the manual existing manual annotations.  
    \item Only consider detection where the confidence is $\geq0.99$.
    \item The dataset must be balanced so that the number of DREB and OREB videos is equal.  
\end{enumerate}

After the inference and filtering process, two different sources of labeled data are obtained. First, a set of 2,000 ground-truth manually labeled videos containing annotations for both the action class and its timestamp. Second, a pseudo-labeled set containing two types of annotations: the action class as ground truth and the action timestamp inferred using the spotting model.

\subsection{Offline Action Anticipation} \label{sec:experiments offline anticipation}

As previously mentioned, offline anticipation can be seen as a natural extension of action classification towards action anticipation. The model is provided with an input video trimmed up to $\tau_a$ seconds before the action occurs. In this way, the model is tasked with performing the classification $\tau_a$ seconds in advance, i.e., predicting which action will happen in the remaining future frames (e.g., $T_V - W$) of the video.

The set of experiments conducted to analyze the effect of varying different parameters on the model's performance are:

\subsubsection{Varying the anticipation time \texorpdfstring{$\tau_a$}{t}} \label{experiment:offline_anticipation_time}

\begin{figure}[htbp]
    \centering
    \begin{subfigure}[t]{0.33\textwidth}
        \centering
        \includegraphics[width=\textwidth]{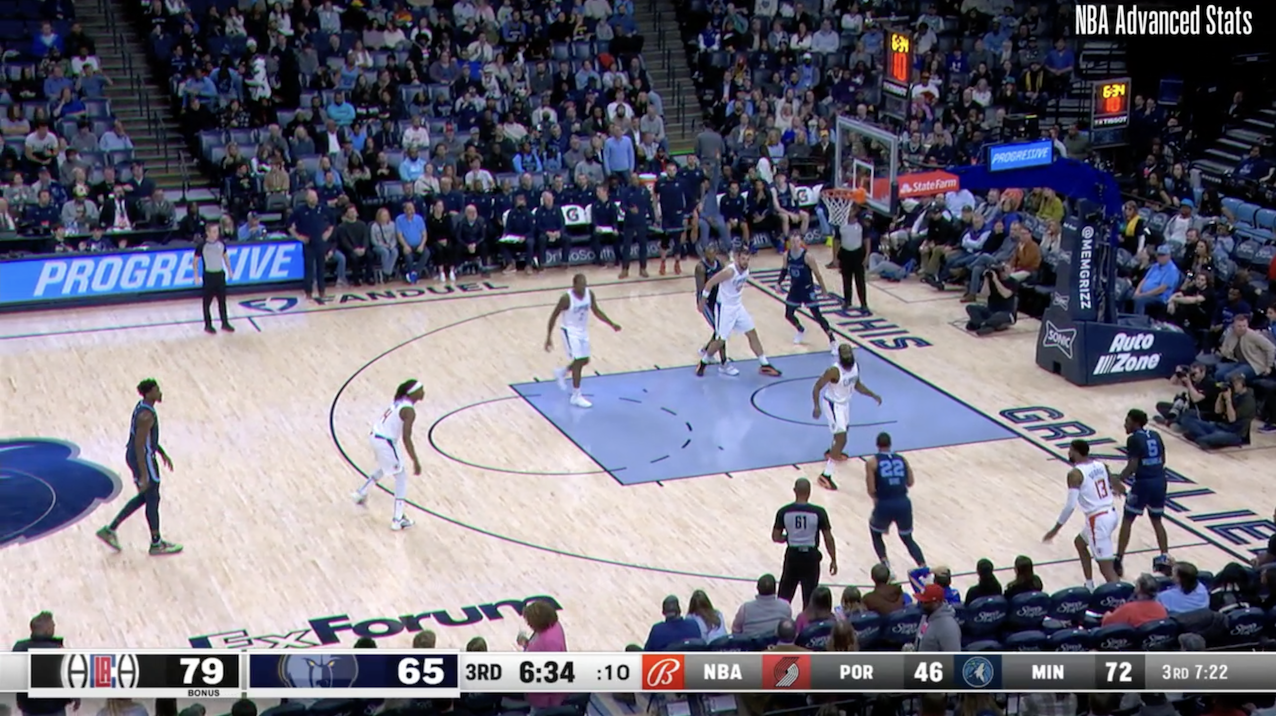}
        \caption{$\tau_a=2.5$s before the action.}
    \end{subfigure}%
    \hfill
    \begin{subfigure}[t]{0.33\textwidth}
        \centering
        \includegraphics[width=\textwidth]{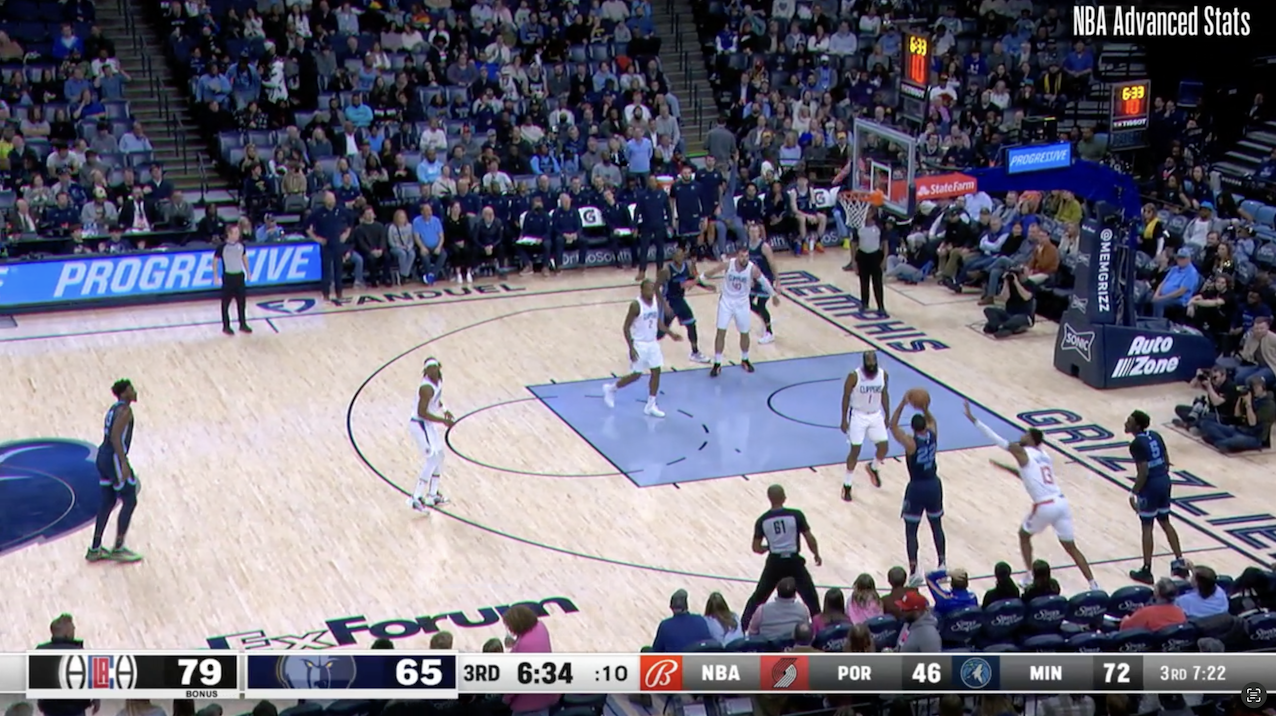}
        \caption{$\tau_a=2.0$s}
    \end{subfigure}%
    \hfill
    \begin{subfigure}[t]{0.33\textwidth}
        \centering
        \includegraphics[width=\textwidth]{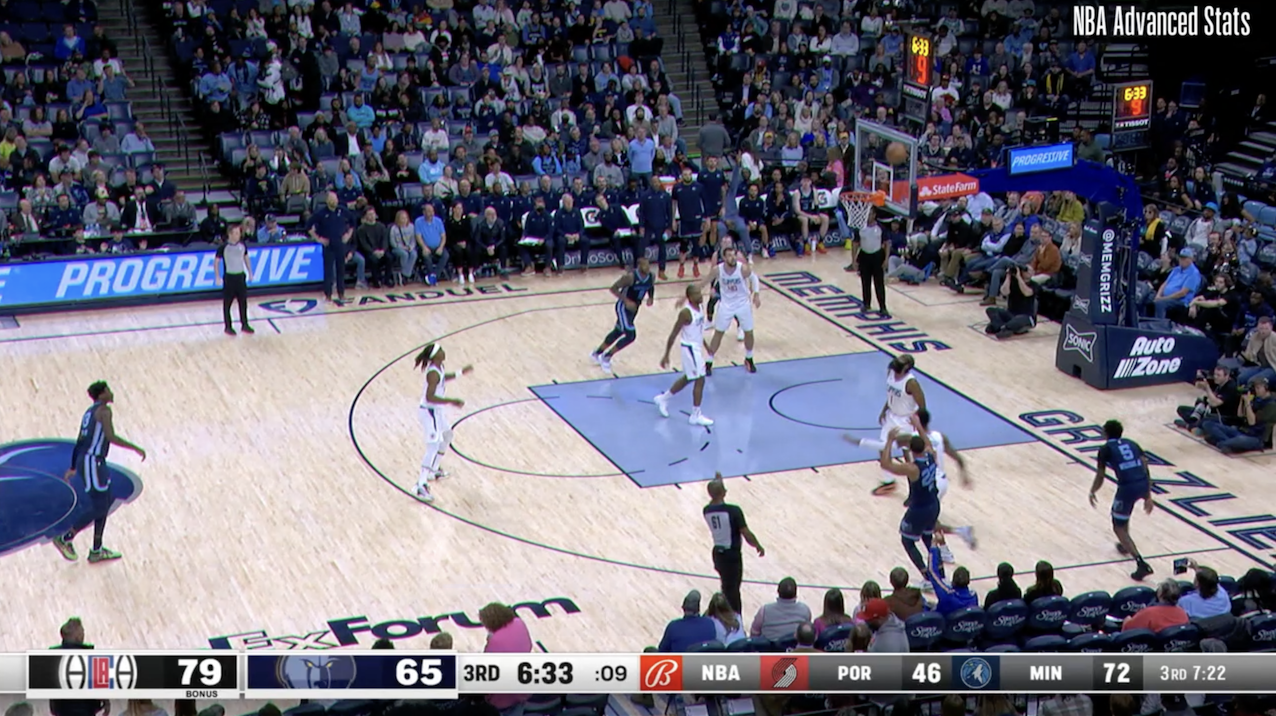}
        \caption{$\tau_a=1.5$s}
    \end{subfigure}
    
    \vspace{0.2cm}
    
    \begin{subfigure}[t]{0.33\textwidth}
        \centering
        \includegraphics[width=\textwidth]{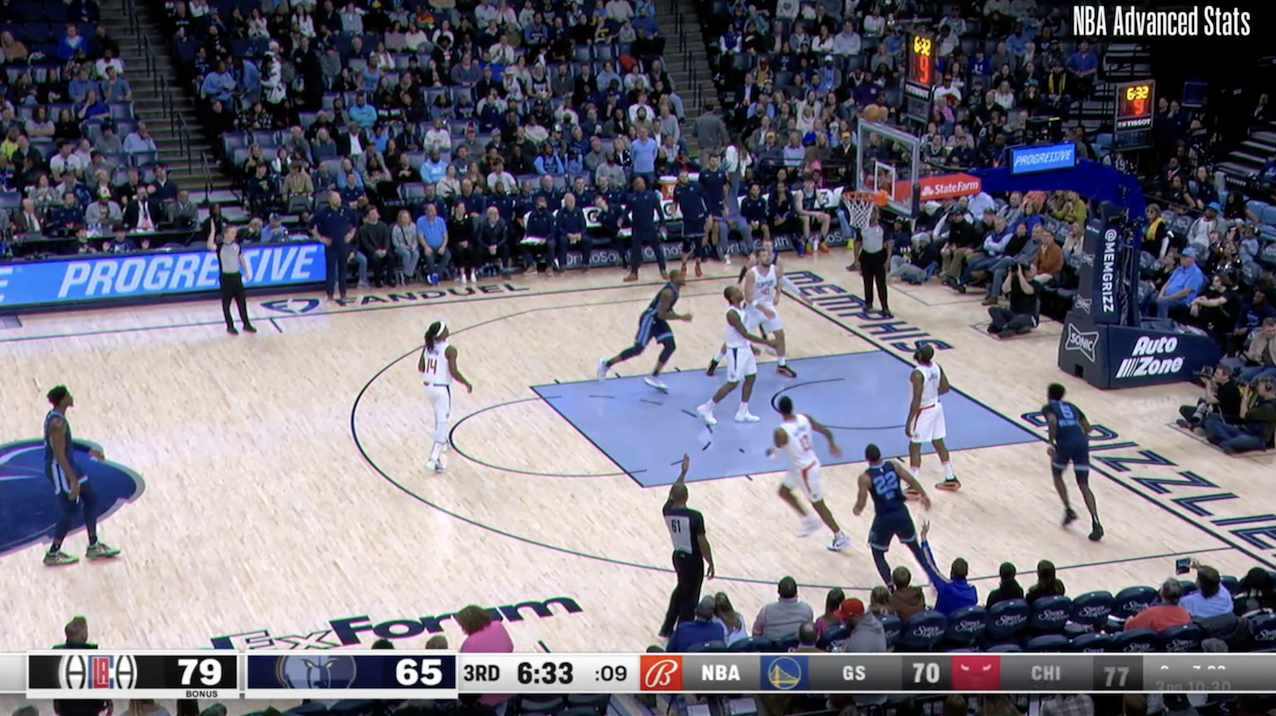}
        \caption{$\tau_a=1.0$s}
    \end{subfigure}%
    \hfill
    \begin{subfigure}[t]{0.33\textwidth}
        \centering
        \includegraphics[width=\textwidth]{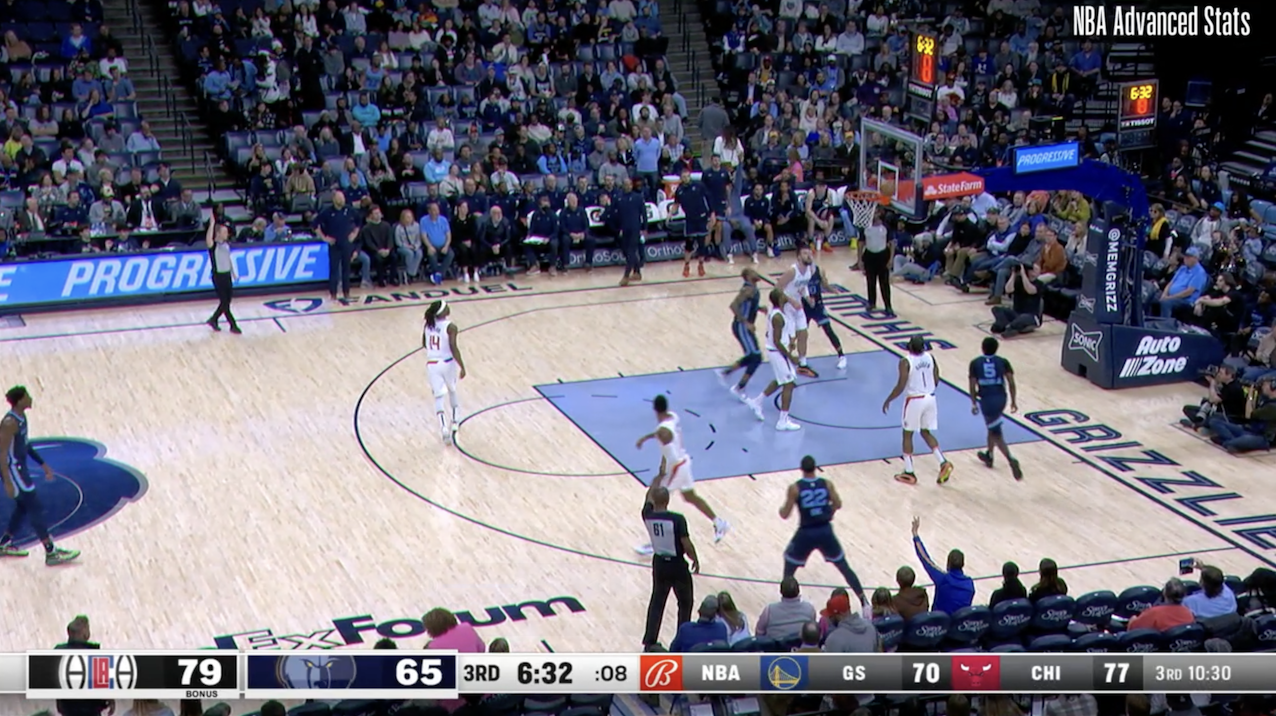}
        \caption{$\tau_a=0.5$s}
    \end{subfigure}%
    \hfill
    \begin{subfigure}[t]{0.33\textwidth}
        \centering
        \includegraphics[width=\textwidth]{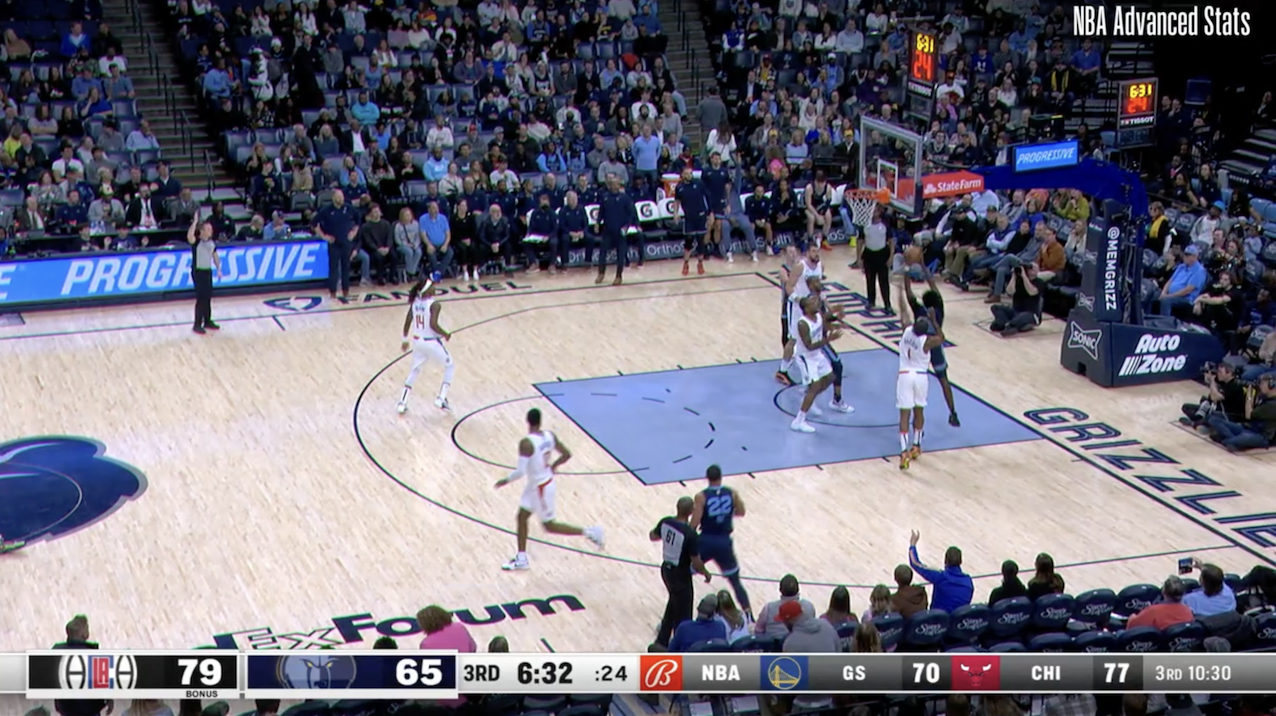}
        \caption{$\tau_a=0$s. Frame where the action occurs.}
    \end{subfigure}
    
    \vspace{0.1cm}
    
    \caption{Illustration of different anticipation times ($\tau_a$) for the same play (OREB). Each subfigure shows the last frame that the model would see when trained at each $\tau_a$.}
    \label{fig:anticipation_time_example}
\end{figure}

The anticipation time parameter $\tau_a$ is varied to evaluate the model's performance when it is given different amounts of temporal context. Four anticipation times are considered: $\tau_a \in \{0.5, 1.0, 1.5, 2.0\}$s. Figure~\ref{fig:anticipation_time_example} illustrates an example play, showing the different anticipation times to provide an intuitive understanding of when the model must make the prediction. For each value of $\tau_a$, a separate model is trained, as models optimized for different anticipation times seem to focus on distinct visual cues. Preliminary experiments indicated that, for example, a model trained with $\tau_a = 1\,\mathrm{s}$ achieved optimal performance for that specific anticipation time, whereas applying it to videos trimmed to other anticipation times (e.g., $\tau_a = 0.5\,\mathrm{s}$) led to suboptimal predictions, even though the model had been trained for a more challenging task.

Finally, it is explicitly noted that only the results of the \textit{baseline} model are presented for the offline anticipation task. This decision is based on two main reasons: 1) space constraints, and 2) experiments with \textit{TEAM} indicated that adding the Transformer encoder does not appear to significantly improve performance. Therefore, the simplest model was retained, as this facilitates the interpretability analysis presented later. For the sake of completeness and transparency, however, the results of all experiments discussed in this section using \textit{TEAM} are provided in the Appendix (Section~\ref{sec:appendix}).

\subsubsection{Augmenting training with pseudo-labeled data} \label{experiment:offline_training_samples}
It is well known that models benefit from a larger number of training samples, as this increases variability and helps prevent overfitting. As previously described, there is a set of 2,000 manually labeled samples for action anticipation and another set of pseudo-labeled samples obtained by performing inference with the spotting model on unlabeled data. In this experiment, a \textit{baseline} model is trained using the manually labeled dataset, $D_{2\text{K}}$, with its corresponding splits $D^{\text{train}}_{2\text{K}}$, $D^{\text{val}}_{2\text{K}}$, and $D^{\text{test}}_{2\text{K}}$. Subsequently, training is repeated with a larger number of samples by incorporating pseudo-labeled data, while evaluation and testing continue to use the manually annotated splits $D^{\text{val}}_{2\text{K}}$ and $D^{\text{test}}_{2\text{K}}$ to ensure a fair comparison. Specifically, two extended training splits are considered: one with $D^\text{train}_{2\text{K}}$ plus 1,500 pseudo-labeled samples (3,000 training samples in total), and another with $D^{\text{train}}_{2\text{K}}$ plus 16,000 pseudo-labeled samples (17,500 in total). Finally, since training with more samples may require a larger number of trainable parameters, these three experiments are repeated with an increased number of trainable parameters (1M) compared to the baseline. This setup allows for an assessment of whether increasing the number of pseudo-labeled samples for training improves performance or whether the manually annotated set alone is sufficient.

\subsubsection{AI vs Human comparison} The third experiment is designed to contextualize the performance of the different models trained for the offline anticipation task. Since the dataset employed is self-curated, no prior studies or models have been applied to this data, which poses a challenge: there is no external reference to determine what constitutes good or poor model performance. To address this, the model’s ability to forecast the future is compared with that of human experts. Five individuals with extensive experience in basketball — each accumulating over 5,000 hours of playing, practicing, watching, or coaching — were asked to perform ``inference" on the test set for the offline anticipation task.

Human inference was conducted as follows. Each expert was presented with complete videos from the test set $D^{test}_{2\text{K}}$, trimmed to a specific anticipation time $\tau_a$. Each sequence could be viewed only once, and the expert had 5 seconds after the sequence ended to make a prediction. The experts were not allowed to communicate with each other and were provided only with the following prior information: (1) each sequence contains a rebound and no sequence ends with a basket (enforcing a binary decision focused solely on the rebound), and (2) the anticipation time $\tau_a$ for which the video was trimmed. To avoid any bias in their decisions, the experts were not informed that the classes were balanced in the test set.

Once each expert provided their predictions, these were aggregated across all five experts using a majority-based criterion. Since there are five experts and only two classes, ties were not possible, and a winning decision was always obtained. After aggregating the experts’ predictions, the comparison between the humans and the AI model was performed using a standard confusion matrix, computing true positives, true negatives, false positives, and false negatives for the two classes. The comparison was conducted for anticipation times of $\tau_a = \{0.5, 1.5\}$ seconds. These two values were selected instead of all four possible anticipation times, as evaluating every case would have been too time-consuming for the humans. Although the choice of $\tau_a$ is somewhat arbitrary, it was intended to cover both a relatively easier scenario ($\tau_a = 0.5$ s) and a more challenging one ($\tau_a = 1.5$ s).

\subsubsection{Interpretability} 
Finally, after training all offline anticipation models, an interpretability analysis will be conducted to identify the regions of the video attended by the model during prediction, providing insights into its decision-making process and motivating the inclusion of such experiments. To this end, LayerCAM~\cite{layercam}, a recent activation-mapping technique, will be used to generate visual explanations for convolutional neural network decisions. LayerCAM computes pixel-level importance weights through the element-wise product of the gradients and activations at each spatial location. Unlike Grad-CAM, which assigns a single channel-wise weight based on globally averaged gradients, LayerCAM produces one weight per pixel, yielding more fine-grained and localized visual explanations at the cost of increased computational effort.

The experiment will focus on comparing anticipation models trained with different $\tau_a$ values. Each model is expected to adapt its learning to maximize forecasting capabilities based on the temporal context provided. The hypothesis is that models trained with different anticipation times may focus on different visual cues.

\subsection{\texorpdfstring{Online Action Anticipation}{Online Action Anticipation}} \label{sec:experiments online anticipation}

The last set of experiments focuses on the action anticipation task in the online setup. Unlike the offline scenario, where fixing the anticipation time also automatically determined the temporal context, in the online setup the temporal context window and the anticipation window are fixed independently. Importantly, the anticipation window may or may not contain an action, depending on whether the event occurs within the temporal horizon of the window. The experiments conducted on the online setup are the following:

\subsubsection{Changing the clip length} \label{experiment:online_clip_length}
The first experiment evaluates the effect of increasing or decreasing the temporal context (i.e., using a smaller or larger sliding window length to create the clips) while maintaining fixed the anticipation window length. The hypothesis is that increasing the temporal context may improve the model's predictions, as providing more prior information about the play allows the model to better understand it and potentially rely on a wider range of visual cues than when the clip duration is limited.

\subsubsection{Changing the anticipation window length} \label{experiment:online_AW_length}

The second experiment examines the effect of changing the anticipation  window (i.e., increasing or decreasing the size of the \textit{sky blue} window in Fig.~\ref{fig:tasks}d) while keeping the temporal context constant. Here, the hypothesis is that a larger temporal window is expected to benefit the model. This is not because a longer prediction window inherently improves understanding of the play, but due to the nature of the clips in the dataset. Since each clip contains a single action, increasing the anticipation window allows the model to be less precise in modeling the time-to-event dynamics. Note that, although the model is not explicitly trained or evaluated to predict the time-to-event, solving the online anticipation task requires it to implicitly encode some temporal information in order to determine whether the action will occur within the next anticipation window or at a later time. Therefore, a larger $AW$ provides a higher margin of error. Intuitively, this means that the model may correctly identify that the upcoming action is an OREB, but the difference between working with $AW = 1.5$s and $AW = 0.5$s is significant. In the first scenario, each prediction corresponds to a longer window, while in the second scenario, the model must make roughly three times more predictions, increasing the probability of errors. Effectively, increasing the $AW$ length simplifies the task by reducing the number of \textit{background} predictions required. In the theoretical limit of $AW \rightarrow \infty$, the task transitions from a multiclass classification problem with three possible classes to a binary classification problem.

\subsubsection{Complementary ablations} 
Finally, although conducting an extensive hyperparameter search was not feasible due to the long training times of video models ($\sim15$–$20$ minutes per epoch), an ablation study is provided to analyze the impact of different hyperparameters related to the task, settings of the model, or even the overall architecture. In particular, for the task-related hyperparameters, the ablation includes: (1) the stride used to sample frames from the videos, where a larger stride results in less dense temporal sampling (fewer frames selected), and (2) the overlap between clips sampled from the original video. For the model-related settings, the analysis considers: (1) the pre-training used to initialize the backbone, comparing the default weights provided for X3D by PyTorch on the Kinetics-400~\cite{kay2017kineticshumanactionvideo} dataset with a custom pre-training on rebound classification, and (2) the resolution of the input frames. Regarding the architecture-related settings, the exploration involves: (1) two different backbones, a 3D CNN (X3D) and a 2D CNN (ConvNeXt\_Tiny), and (2) three different alternatives for the temporal module, namely the vanilla X3D without additional components (the \textit{baseline} model), a Transformer encoder, and an LSTM, thereby covering the three most common families of architectures for temporal modeling in video understanding. Finally, specifically for the Transformer-based model (the main method of this work, \textit{TEAM}), the study examines the impact of varying the number of heads in the MHSA as well as the number of encoder layers.

%
%

\section{Implementation and training details} \label{sec:implementation_details}

In this section, the specific hyperparameters and implementation details of the models used for each experiment and task solved in this project are described. The structure of this section is as follows: first, the general training settings applied across all experiments, regardless of the model, are presented. Subsequently, the two models used in this work are introduced, the \textit{baseline} model and \textit{TEAM}, including detailed descriptions of their architectural implementations and any model-specific training configurations.

\subsection{General training configuration}

\subsubsection{Frame resolution} Regarding the input data, all frames passed to the model — whether from a complete video (action classification or spotting), a trimmed video (offline anticipation), or as an untrimmed video sampled with short clips (online anticipation) — are resized to a resolution of 256×455. 

\subsubsection{Data augmentation} Additionally, as explained in Section~\ref{sec:dataset}, the data augmentations applied include: color jitter (hue $\pm$0.2, brightness/contrast/saturation scaled between 0.7 and 1.2), horizontal flip, a slight Gaussian blur (kernel size between 3 and 7), random resized crop (scale between 0.9 and 1.0, aspect ratio fixed at 455/256), and affine transformations (rotation $\pm$10\textdegree, translation $\pm$5\%, scaling between 0.95 and 1.05). All transformations were applied with a probability of 50\% and performed consistently across all frames of each video (i.e., the exact same set of transformations was applied to every frame within a video).

\subsubsection{Optimizer} All models were trained using the AdamW optimizer with a weight decay of $5\times10^{-5}$. Task-specific learning rates were employed to ensure optimal convergence for each experimental setup.

\subsection{Baseline}
As previously mentioned, the \textit{baseline} model is a 3D CNN, specifically a vanilla X3D\_m. This model is employed in action classification, action spotting, offline action anticipation, and online action anticipation experiments. Nevertheless, each task may require specific adaptations of the model to better align with its objectives. Therefore, along this work, two different implementations of the X3D are used. Note that X3D is used both as a full model for action classification and spotting, and later also as a backbone to extract features for anticipation tasks.

\subsubsection{Implementation for Action Classification, Offline Anticipation and Online Anticipation} \label{sec:baseline_classification_anticipation}

The implementation of the \textit{baseline} model for action classification and offline anticipation is identical, as both tasks share essentially the same inputs and outputs. The specific implementation detailed below is used for experiments \ref{sec:auxiliary tasks experiments}, \ref{experiment:offline_anticipation_time}, \ref{experiment:offline_training_samples}, \ref{experiment:online_clip_length} and \ref{experiment:online_AW_length}.

From the original PyTorch implementation of X3D\_m, only the final classification layer of the head needs to be adapted. The original layer produces 400 outputs, corresponding to the 400 classes of the Kinetics-400 dataset in which X3D is pre-trained. For the current tasks, this output dimension is reduced from 400 to 1. During inference, a \texttt{sigmoid} function is applied to map the output to a probability in $[0,1]$, which allows making a binary decision: values above the threshold 0.5 are interpreted as \textit{offensive rebound}, while values below are interpreted as \textit{defensive rebound}.

Similarly, for online action anticipation, the vanilla X3D\_m is employed, with the only modification being also the adjustment of the final classification head. However, in this case, the task involves three possible output classes: \textit{defensive rebound}, \textit{offensive rebound}, and the additional background class \textit{no rebound}. Consequently, the output dimension of the classification head is adapted from 400 to 3, and during inference a \texttt{softmax} activation function is applied to obtain a normalized probability distribution over the classes.

Although the architecture implementation is nearly identical across tasks, the training settings differ. These are described below:

\paragraph{Training configuration Action Classification} The model is initialized with the PyTorch default weights pre-trained on Kinetics-400 and is subsequently fine-tuned for classification, with approximately 3M of its parameters unfrozen (80\% of the total trainable parameters). The final training setup included: 
batch size = 4,  
learning rate = 8e-4 with cosine annealing scheduler,   
stride = 5,  
and a maximum of 25 epochs with an early stopping patience of 6 epochs. Training was performed with 25,000 sequences, evaluated on 1,000 sequences for early stopping, and tested on a separate set of 1,000 sequences. The average training time per epoch was $\sim$2h on a single GPU Nvidia GeForce RTX 3090. 

The model is trained using the standard Binary Cross-Entropy (BCE) loss, as defined in Equation~\ref{eq:BCE_loss}, where $N$ denotes the number of training samples, $y_i \in \{0,1\}$ represents the ground-truth label for the $i$-th sample, and $\hat{y}_i \in [0,1]$ corresponds to the predicted probability assigned to the positive class by the model.

\begin{equation} \label{eq:BCE_loss}
    \mathcal{L}_{\text{BCE}} = - \frac{1}{N} \sum_{i=1}^{N} \Big( y_i \log \hat{y}_i + (1 - y_i) \log (1 - \hat{y}_i) \Big)
\end{equation}

\paragraph{Training configuration Offline Anticipation} The model was initialized with the best weights from action classification, with only the last 100,000 parameters set as trainable (note that a greater number of parameters can be frozen than in the classification task, as pre-training on a large dataset has already adapted the earlier layers to the visual features of basketball videos), batch size = 10, learning rate = $1 \times 10^{-4}$ with a cosine annealing scheduler, binary cross-entropy loss, stride = 5, and a maximum of 100 epochs with an early stopping patience of 10 epochs. Training was performed on 1,500 sequences, validated on 250 for early stopping, and tested on a separate set of 250. Each epoch took around 5 minutes on a single Nvidia GeForce RTX 3090 GPU.

\paragraph{Training configuration  Online Anticipation} Initialization on best weights of action classification with the last 500,000 parameters of the backbone unfrozen ($\sim13$\% of the total parameters in the backbone),
batch size = 15,  
learning rate = $1 \times 10^{-4}$ with a cosine annealing scheduler, 
stride = 3,  
overlap = 0.5,  
Clip length = 2s,  
AW length = 1s,  
and a maximum of 100 epochs with an early stopping patience of 10 epochs. Each epoch took around 15 minutes on a single Nvidia GeForce RTX 3090 GPU.

The model is trained using a multi-class cross-entropy loss (see Equation~\ref{eq:CE_loss_online}) with class weights [0.49, 0.49, 0.02] for the OREB, DERB and no rebound classes respectively. Here, $V$ denotes the number of videos in the batch, $C$ is the number of classes (3 in this case), $y_{v,c} \in \{0,1\}$ is the ground-truth indicator for class $c$ of video $v$, and $\hat{y}_{v,c} \in [0,1]$ represents the predicted probability for class $c$ for the same video. Since online anticipation exhibits class imbalance — with relatively few action clips compared to background clips — a class-dependent weight \(w_c\) is introduced in the loss function to ensure that action classes contribute more strongly to the optimization. Without this weighting, the signal from action clips could be easily overshadowed by the overwhelming number of background clips. In particular, the class weights for online spotting are set to \([0.49, 0.49, 0.02]\) for DREB, OREB, and background, respectively. Note that no normalization term is required, as all clips have the same length, being created with a fixed sliding window.

\begin{equation} \label{eq:CE_loss_online}
\mathcal{L} = - \sum_{v=1}^{V} \sum_{c=1}^{C} w_c \cdot y_{v,c} \, \log \hat{y}_{v,c}
\end{equation}

\subsubsection{Implementation for Action Spotting}
The main particularity of action spotting, compared to action classification or anticipation, is that temporally precise predictions are required. This is achieved by producing a prediction for each input frame, since the task is formulated as a frame-wise classification problem. If the same approach as in \ref{sec:baseline_classification_anticipation} were followed — adapting only the final classification layer to match the number of input frames — the model would be forced to first compress the features and then artificially expand them again. To avoid this, the last block of the vanilla X3D\_m, which is responsible for compressing the features along the temporal dimension, is removed. As a result, after applying adaptive average pooling over the spatial dimensions, the model outputs feature maps of size $T \times 192$, where spatial information is aggregated per frame while the temporal resolution is preserved (i.e., the final value of $T$ depends on the temporal resolution of the input video). Finally, since action spotting involves three possible classes — \textit{defensive rebound}, \textit{offensive rebound}, and the background action \textit{no rebound} — the classification head is adapted accordingly. This implementation of the \textit{baseline} model is only used for the experiments in \ref{experiment:spotting}. 

Before continuing, it should be noted that the action spotting method used in this work (a vanilla X3D\_m) is relatively rudimentary, and numerous improvements could be implemented to achieve state-of-the-art performance. However, this task alone constitutes an active field of research and could serve as the focus of a dedicated master's thesis. As this is not the central topic of the project, a minimal viable solution was implemented to enable experiments involving training with a larger number of samples for action anticipation without requiring manual annotation. This choice naturally affects the quality of the resulting pseudo-labeled samples, which must be considered as a limitation when analyzing the results presented in Section~\ref{sec:results}.

\paragraph{Training configuration}

The last two blocks of the network were set as trainable parameters (representing an 80\% of the total parameters),
batch size = 4,  
learning rate = 1.8e-3 with a cosine annealing scheduler, 
stride = 3, and a maximum of 150 epochs with an early stopping patience of 8 epochs. Training was performed on 1,500 sequences, validated on 250 for early stopping, and tested on a separate set of 250. Each epoch took around 10 minutes in a single GPU Nvidia GeForce RTX 3090.

Since each frame can belong to one of three possible categories — DREB, OREB, or background — the model is trained using the standard multi-class cross-entropy loss (see Equation~\ref{eq:CE_loss}), which is applied to the frame-wise multiclass classification problem. Here, $V$ denotes the number of videos in the batch, $F_v$ is the number of frames in video $v$, $C$ is the number of classes, $y_{v,f,c} \in \{0,1\}$ is the ground-truth indicator for class $c$ at frame $f$ of video $v$, and $\hat{y}_{v,f,c} \in [0,1]$ represents the predicted probability for class $c$ at the same frame. Action spotting also exhibits a strong class imbalance — with hundreds of background frames for each action frame — therefore a class-dependent weight $w_c$ is also introduced in the loss function to ensure that action classes contribute more strongly to the optimization. In particular, the imbalance in spotting is much larger than in online anticipation and the weight classes are: [0.49999, 0.49999, 0.00002] for DREB, OREB, and background classes respectively. The normalization term $\sum_{v=1}^{V} F_v$ corresponds to the total number of frames across the entire batch, where $F_v$ denotes the number of frames in video $v$. This normalization ensures that the loss is averaged over all frames in the batch, rather than depending on the absolute batch size or the individual video lengths. Without this averaging, batches containing more frames would contribute disproportionately to the optimization, thereby biasing the training process. By normalizing in this way, each frame contributes equally to the loss, leading to a more balanced and stable optimization procedure.

\begin{equation} \label{eq:CE_loss}
    \mathcal{L} = - \frac{1}{\sum_{v=1}^{V} F_v} 
    \sum_{v=1}^{V} \sum_{f=1}^{F_v} \sum_{c=1}^{C} 
    w_c \cdot y_{v,f,c} \, \log \hat{y}_{v,f,c}
\end{equation}

\paragraph{Inference}

During inference, it is common in action spotting models to apply post-processing to the raw predictions produced by the model. In particular, a three-step post-processing pipeline was applied:

\begin{enumerate}
    \item Temporal smoothing of the probabilities: A moving average filter is applied to the class probabilities over time to reduce prediction noise and enforce temporal consistency. Specifically, for each class independently, a one-dimensional convolution with a uniform kernel (i.e., all weights equal to $1/\textit{window}$) is applied to the sequence of probabilities, preserving the original sequence length. The optimal value for the \textit{window} parameter was determined experimentally as the value that maximized the mAP on the validation set; this value was 7. This means that, for each frame, the predicted probability is smoothed by averaging over a temporal window that includes the current frame, the three preceding frames, and the three following frames.

    \item Probability thresholding: The next step filters out low-confidence detections. For each frame, the model checks the maximum predicted probability across all classes. If this value is below 0.7, the frame is automatically assigned to the background class with a probability of 1. As before, this threshold was selected empirically as the value that maximized the mAP on the validation set.
    
    \item Non-maximum suppression (NMS): Finally, we apply a vanilla non-maximum suppression filter, is applied to avoid multiple detections of the same action, very common in the literature very common in the literature~\cite{xarles2024tdeed,Rohrbach2012_nms_postprocessing,tang2019_nms_postprocessing}. A temporal window of 2.5 seconds is used, determined empirically. For each detection that passes the previous two filters, 1.25 seconds forward and 1.25 seconds backward are inspected, and only the detection with the highest confidence within that window is retained. This temporal window is chosen to ensure that the detection is robust while avoiding the need to scan the entire video, which is unnecessary given that each video contains a single action.
\end{enumerate}

The effect that each one of this steps has on the model probabilities is illustrated in Figure~\ref{fig:spotting_post_processing}.

\subsection{Transformer-Encoder Anticipation Model (TEAM)}
The second model that will be used in this work has already been introduced in Section~\ref{sec:method}, and it is an adaptation of the current state-of-the-art for online action anticipation. Therefore, it will only be used in the experiments belonging to the online anticipation task: \ref{experiment:online_clip_length} and \ref{experiment:online_AW_length}. 

The \textit{TEAM} architecture employs an X3D\_m backbone pre-trained on action classification. The only modification introduced to the backbone consists of the removal of the last block, following the approach used in action spotting, in order to preserve the temporal resolution of the input clip. As a result, the backbone produces one feature vector per frame, yielding a tensor of dimensions $T \times D$ for each sample in a batch, where $T = 40$ (corresponding to a clip of 2 seconds at 60 fps with a stride of 3), and $D = 192$ denotes the dimensionality of each per-frame feature vector.

For the temporal module, two transformer encoder layers with MHSA are used. By default, each layer employs 8 heads, a feed-forward network with an inner dimension of 512, a dropout rate of 0.1, and GELU activation. Positional encoding, implemented as a randomly initialized learnable embedding, is added to the features before the transformer encoder. It has a maximum length of $N_{\text{max}} = 900$, corresponding to the number of frames in a 15-second clip at 60 fps (i.e., the maximum possible duration of a clip in the dataset used), and is added element-wise to the input features. Finally, a CLS token, initialized to zeros, is prepended to the feature tensor produced by the backbone, effectively increasing the temporal dimension by one (i.e., $T = 41$).

\subsubsection{Training configuration} \label{sec:TEAM_online_training_config}

The \textit{TEAM} Backbone initialization on best weights of action classification with the last 500,000 parameters of the backbone unfrozen ($\sim13$\% of the total parameters in the backbone),
batch size = 15,  
learning rate = $1 \times 10^{-4}$ with a cosine annealing scheduler, cross-entropy loss with class weights [0.49, 0.49, 0.02] for DREB, OREB, and background classes respectively,
stride = 3,  
overlap = 0.5,  
Clip length = 2s,  
AW length = 1s,  
Frame resolution = 455 × 256,  
and a maximum of 100 epochs with an early stopping patience of 10 epochs.

%
%

\section{Results} \label{sec:results}
In this section, the results of the experiments described in Section~\ref{sec:experiments} are presented and discussed. Accordingly, the subsection structure mirrors that of Section~\ref{sec:experiments}.

\subsection{Auxiliary tasks}
\subsubsection{Supervised Pre-training - Action Classification}
\begin{figure}[htbp]
    \centering
    \begin{subfigure}[t]{0.45\textwidth}
        \centering
        \includegraphics[width=\textwidth]{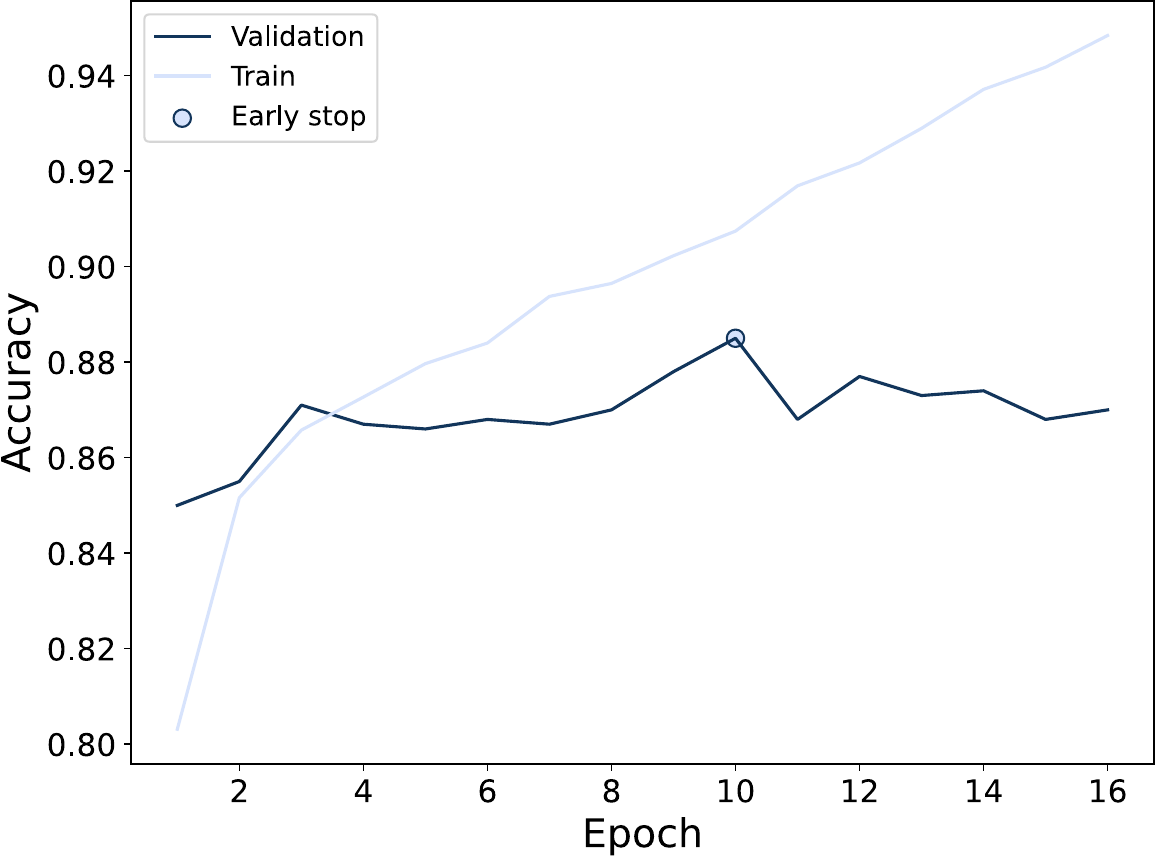}
        \caption{Evolution of training (light blue) and validation (dark blue) accuracy across epochs. The marker denotes the best validation epoch at which early stopping was applied.}
        \label{fig:classification_train_curves}
    \end{subfigure}%
    \hfill
    \begin{subfigure}[t]{0.40\textwidth}
        \centering
        \includegraphics[width=\textwidth]{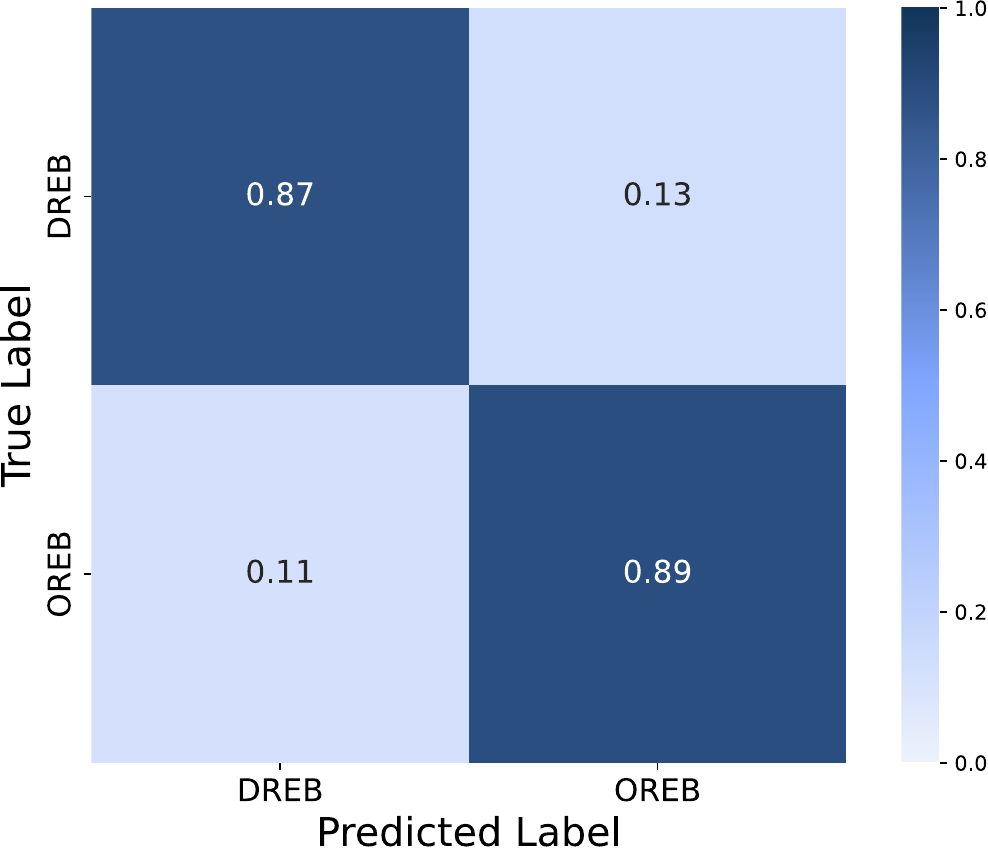}
        \caption{Confusion matrix on the test split, using the weights of the early stop epoch (10). It is normalized across the ``True Label" axis (horizontally).}
        \label{fig:classification_CM}
    \end{subfigure}
    
    \vspace{0.1cm}
    
    \caption{Results for the binary rebound classification model: \textit{a)} training and validation curves; \textit{b)} confusion matrix of the test split.}
    \label{fig:classification_results}
\end{figure}

As described in Section~\ref{sec:method}, the \textit{baseline} model was pre-trained on the rebound classification task (OREB vs. DREB) using complete video sequences. Figure~\ref{fig:classification_train_curves} shows the training and validation accuracy curves. The validation accuracy stabilizes on the first few epochs, and peaks at the 10th epoch at 0.88, while the training accuracy continues to improve up to 0.91\%. This indicates a mild overfitting tendency, mitigated by the early stopping mechanism. Figure~\ref{fig:classification_CM} and Table~\ref{tab:classification_metrics} summarize the classification metrics on the held-out test set. The model achieves an overall accuracy of 0.88, matching the performance on the validation set, which indicates that, despite optimizing for validation accuracy, the model did generalize well. Furthermore, performance is well-balanced between the two classes, with F1-scores of 0.88 for both DREB and OREB. The absence of a strong bias toward either class suggests that the model has learned discriminative spatio-temporal features relevant to both rebound types.

Overall, the results confirm that the backbone effectively captures patterns distinguishing offensive and defensive rebounds, denoting no significant difference in difficulty to identify both classes and providing a solid foundation for subsequent fine-tuning on anticipation and spotting tasks.

\begin{table}[htbp]
\centering
\renewcommand{\arraystretch}{1.2}
\caption{Classification metrics per class and overall accuracy for the binary classification model on the test split.}
\begin{tabular}{lcccc}
\hline
\textbf{Class} & \textbf{Precision} & \textbf{Recall} & \textbf{F1-score} & \textbf{Support} \\
\hline
DREB & 0.89 & 0.87 & 0.88 & 500 \\
OREB & 0.87 & 0.89 & 0.88 & 500 \\
\hline
\multicolumn{4}{r}{\textbf{Overall accuracy}} & 0.881 \\
\hline
\end{tabular}
\label{tab:classification_metrics}
\end{table}

\newpage
\subsubsection{Timestamp Pseudolabeling via Action Spotting}

\begin{figure}[htbp]
    \centering
    \begin{subfigure}[t]{0.24\textwidth}
        \centering
        \includegraphics[width=\textwidth]{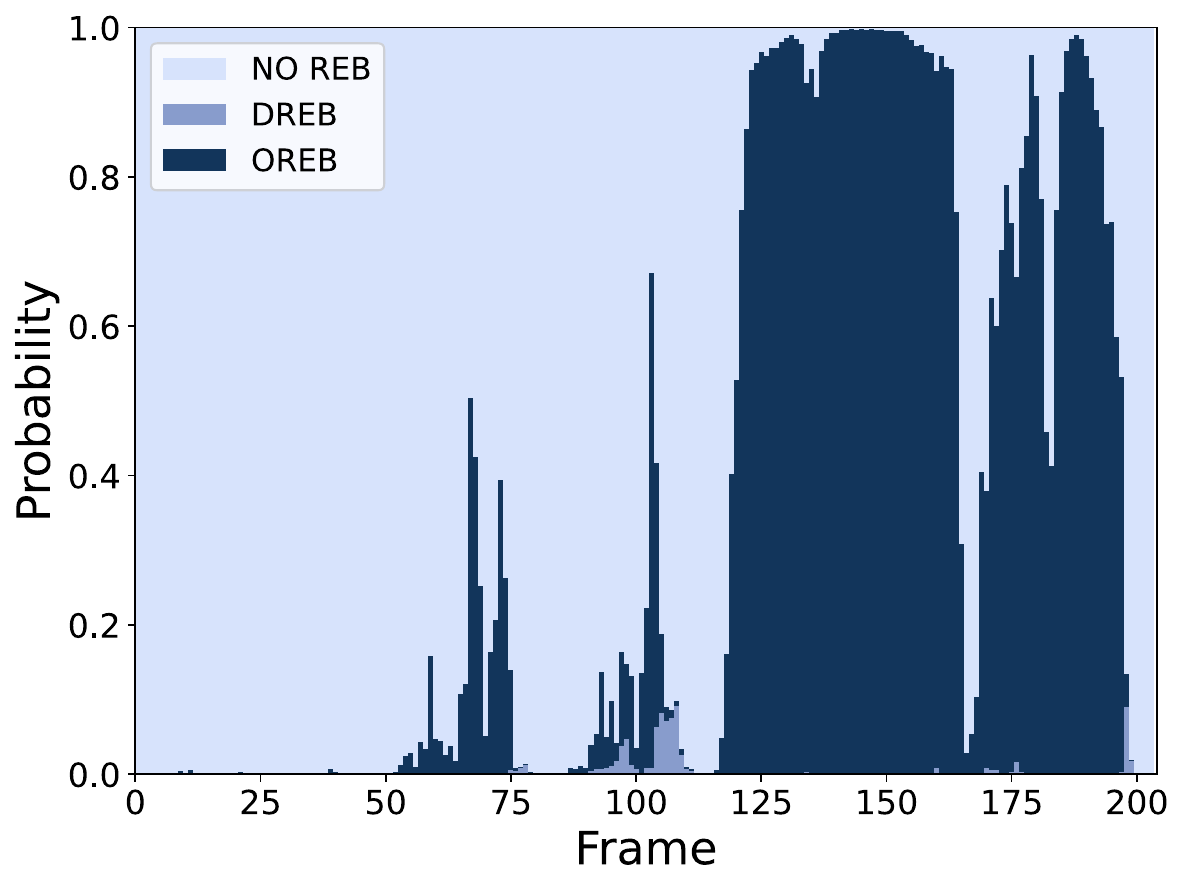}
        \caption{Raw probabilities.}
    \end{subfigure}%
    \hfill
    \begin{subfigure}[t]{0.24\textwidth}
        \centering
        \includegraphics[width=\textwidth]{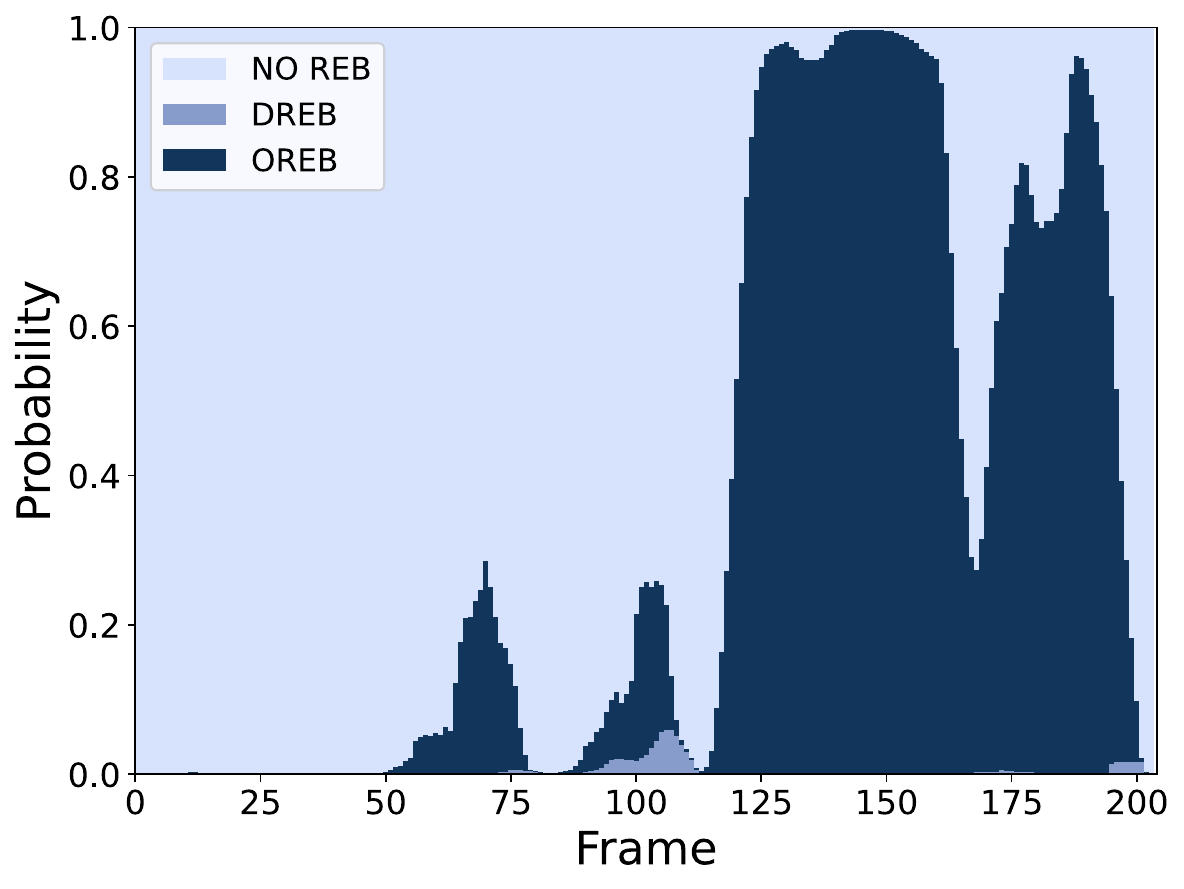}
        \caption{After temporal smoothing.}
    \end{subfigure}%
    \hfill
    \begin{subfigure}[t]{0.24\textwidth}
        \centering
        \includegraphics[width=\textwidth]{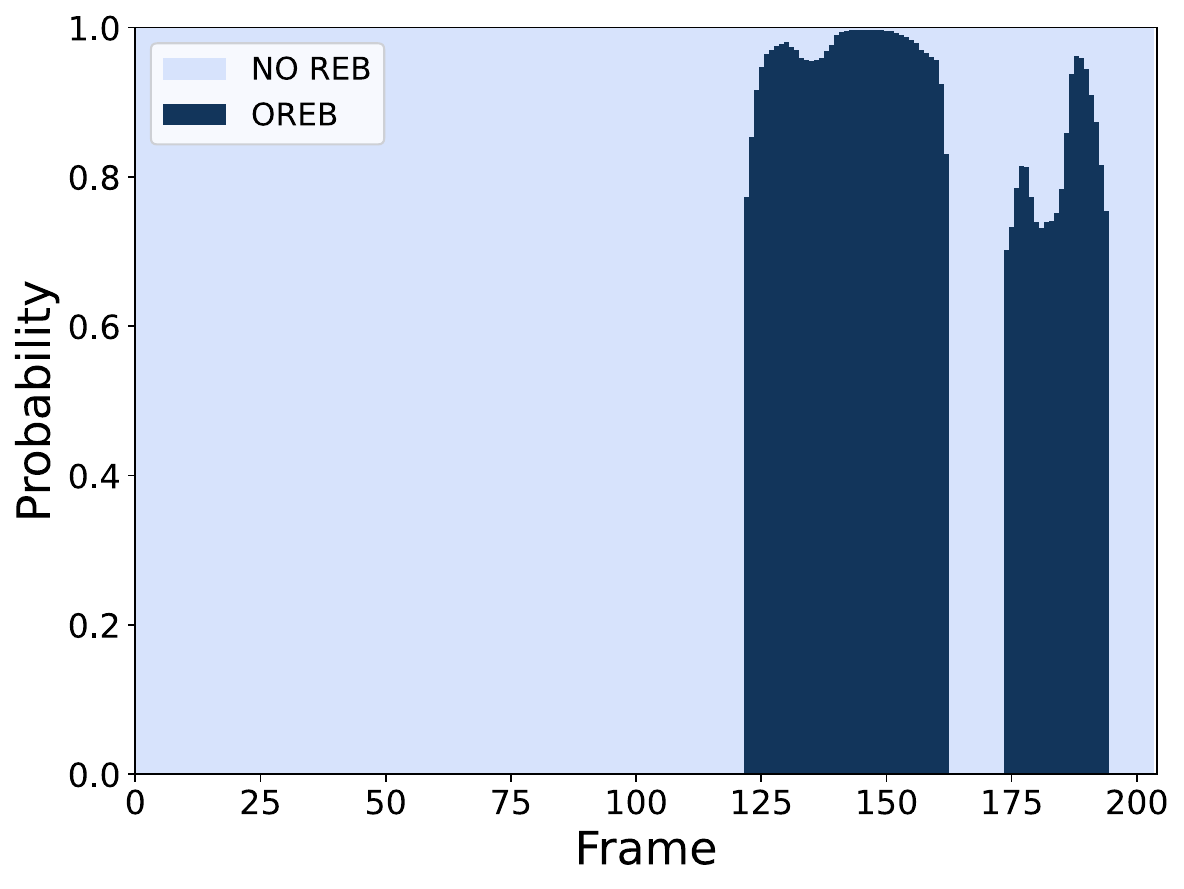}
        \caption{After thresholding.}
    \end{subfigure}
    \hfill
    \begin{subfigure}[t]{0.24\textwidth}
        \centering
        \includegraphics[width=\textwidth]{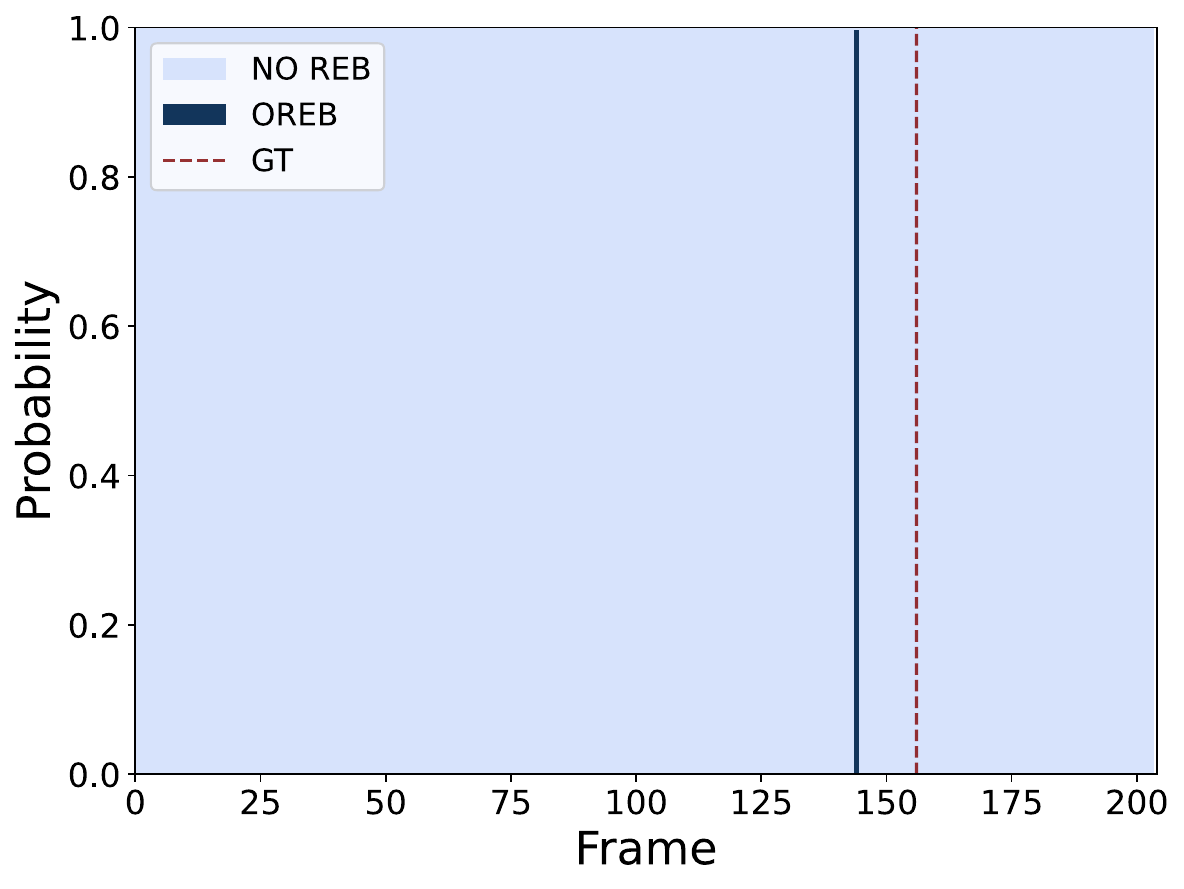}
        \caption{Final detections after NMS and Ground Truth.}
    \end{subfigure}
    
    \vspace{0.1cm}
    
    \caption{Post-processing pipeline applied to the raw probabilities generated by the model in the rebound spotting task.}
    \label{fig:spotting_post_processing}
\end{figure}

\paragraph{Effect of the post-processing}
As described in Section~\ref{sec:implementation_details}, the raw outputs of the model are processed through a post-processing pipeline designed to filter redundant detections. Before presenting the overall performance results, it is important to highlight the impact of each step of this pipeline, as they substantially affect the final predictions. The pipeline consists of three sequential operations applied to the raw probability scores: (1) temporal smoothing with a 7-frame window, (2) thresholding at 0.7 on the smoothed probabilities, and (3) non-maximum suppression with a 2.5-second window (corresponding to 50 frames, since a stride of 3 is used). Figure~\ref{fig:spotting_post_processing} illustrates the distribution of probabilities for each of the three classes after the successive application of these steps, thereby showing the individual effect of each operation.

\begin{table}[htbp]
\centering
\renewcommand{\arraystretch}{1.2}
\caption{Ablation study of post-processing configurations evaluated on the validation set. Results present mAP computed across tolerance windows $\delta \in \{1, 2, 3\}$ seconds and subsequently averaged, reported both per-class (DREB, OREB) and as the overall average across both classes.}

\begin{tabular}{cccccc}
\hline
\textbf{Temporal Smoothing} & \textbf{Thresholding} & \textbf{NMS} & \textbf{DREB mAP@$\{1,2,3\}$} & \textbf{OREB mAP@$\{1,2,3\}$} &  \textbf{Average mAP@$\{1,2,3\}$} \\
\hline
\texttimes & \texttimes & \texttimes & 0.30 & 0.21 & 0.26 \\
\texttimes & \checkmark & \checkmark & 0.54 & \textbf{0.35} & 0.44 \\
\checkmark & \checkmark & \checkmark & \textbf{0.58} & 0.34 & \textbf{0.46} \\
\hline
\end{tabular}
\label{tab:postprocessing_ablation}
\end{table}

As shown in Table~\ref{tab:postprocessing_ablation}, the ablation study demonstrates the impact of different post-processing steps on rebound spotting performance. The notation mAP@$\{\delta_1, \delta_2, \delta_3\}$ represents the arithmetic mean of mAP values across the specified temporal tolerance thresholds. The results show that without any post-processing (\texttimes\ \texttimes\ \texttimes), the model achieves the lowest mAP for both DREB and OREB classes, as well as the lowest average mAP. Using thresholding and NMS (\texttimes\ \checkmark\ \checkmark) leads to a substantial improvement, particularly for OREB, indicating that removing low-confidence predictions and suppressing multiple detections helps reduce false positives. Finally, adding temporal smoothing (\checkmark\ \checkmark\ \checkmark) further increases the DREB mAP but slightly worsens the results for the OREB class, highlighting that smoothing the predictions over time provides more consistent and accurate detection of rebounds.

\begin{figure}[htbp]
    \centering
    \begin{subfigure}[t]{0.43\textwidth}
        \centering
        \includegraphics[width=\textwidth]{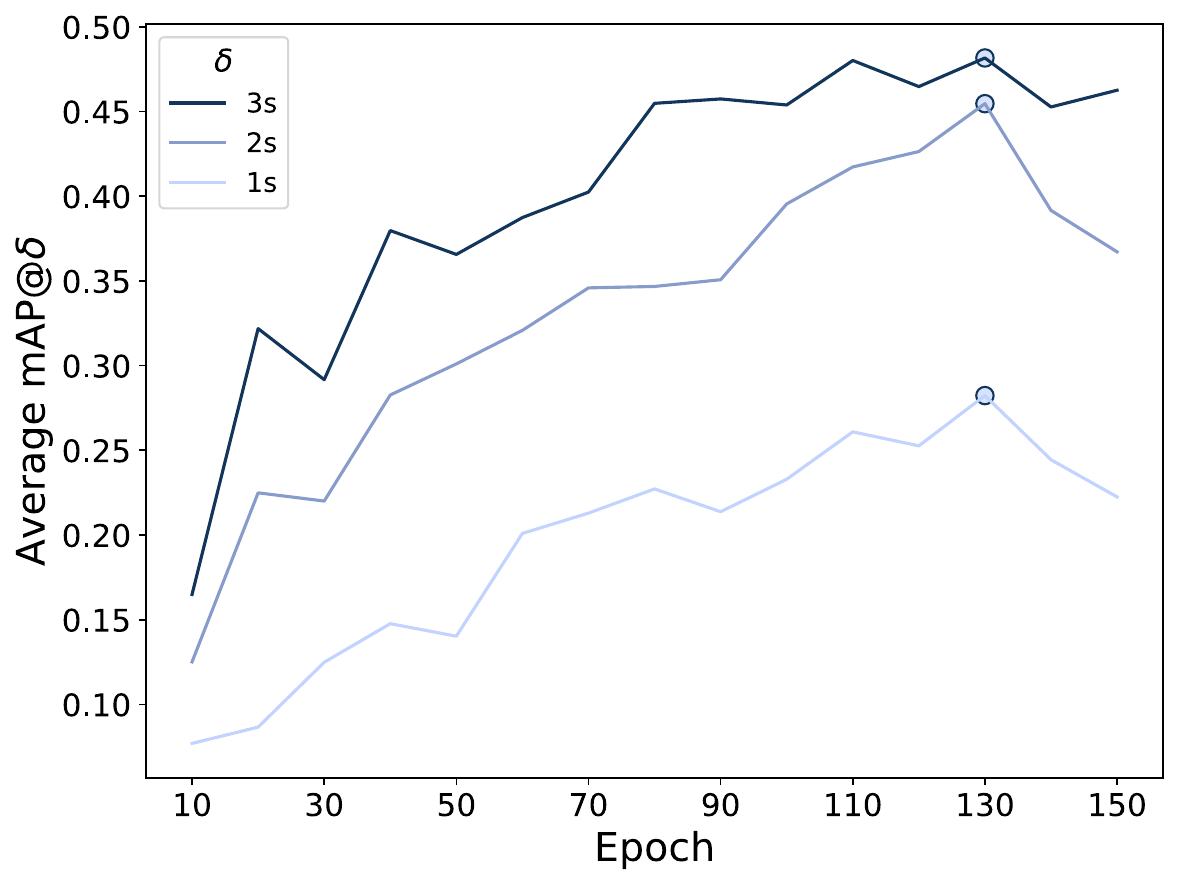}
        \caption{Evolution of the mAP@$ \delta$ at different thresholds, averaged over the DREB and OREB classes, on the validation split. The circle indicates the epoch where early stopping was triggered.}
        \label{fig:spotting_val_evol}
    \end{subfigure}%
    \hfill
    \begin{subfigure}[t]{0.535\textwidth}
        \centering
        \includegraphics[width=\textwidth]{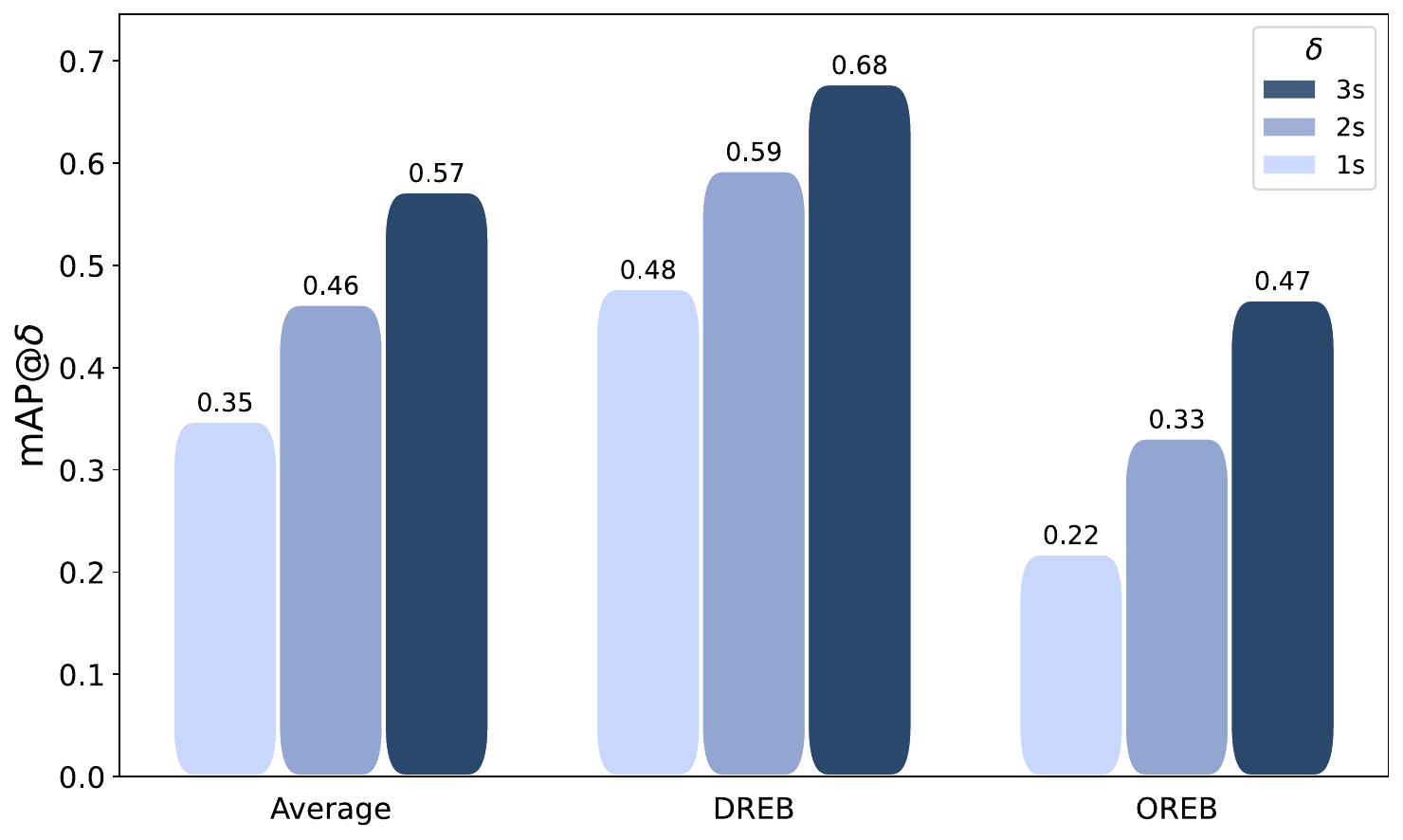}
        \caption{Average mAP@$ \delta $ across both classes, as well as the mAP@$ \delta $ for each individual class, on the test split at different thresholds.}
        \label{fig:spotting_test_barplot}
    \end{subfigure}
    
    \vspace{0.1cm}
    
    \caption{Results for rebound spotting model on the \textit{a)} validation split and \textit{b)} test split.}
    \label{fig:spotting_results}
\end{figure}

\paragraph{Ablation on tolerances window sizes ($\delta$)}
After analyzing the specific impact of the post-processing pipeline, we now turn to a broader evaluation of the model’s performance. Figure~\ref{fig:spotting_results} presents the results for the rebound spotting model. 

First, let's look at Figure~\ref{fig:spotting_val_evol}, where the evolution of the mAP@$\delta$ for thresholds of 1s, 2s, and 3s on the validation split is presented. The circle indicates the epoch early stopping was triggered. We see that the mAP increases consistently across all $\delta$ values until it peaks at epoch 130. Furthermore, we observe that the higher the $\delta$, the higher the mAP. This behaviour is expected, as a larger $\delta$ tolerates a greater temporal misalignment between the predicted and ground truth events while still considering the prediction correct. Note that we do not represent the evolution of the training curves, because post-processing is applied to the validation split but not when evaluating the training metrics. Therefore, it would not be a fair comparison.

\begin{figure}[htbp]
    \centering
    \begin{subfigure}[b]{0.98\textwidth}
        \centering
        \includegraphics[width=\textwidth]{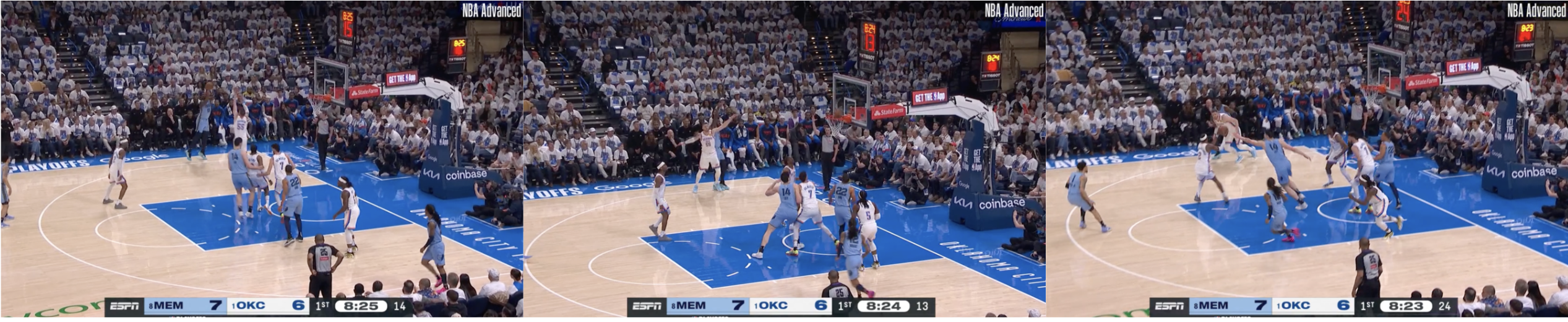}
        \caption{\href{https://youtu.be/T1gsvP9BDuQ}{\underline{Example}} of an isolated player getting a DREB, easier to spot.}
        \label{fig:DREB_isolated_example}
    \end{subfigure}
    
    \vspace{1em}  
    
    \begin{subfigure}[b]{0.98\textwidth}
        \centering
        \includegraphics[width=\textwidth]{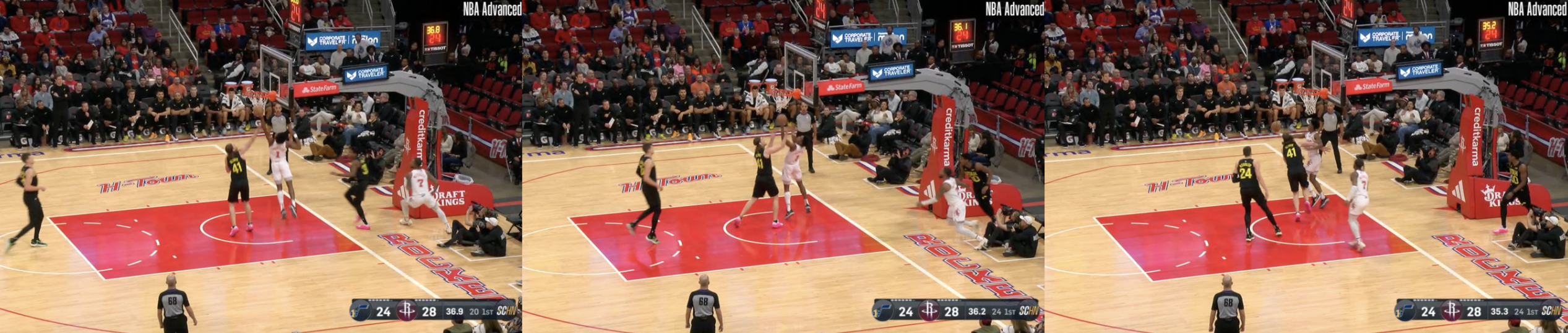}
        \caption{\href{https://youtu.be/hCz_hlabfGc}{\underline{Example}} of an OREB where the ball is tipped two times (frames 1 and 2) before being controlled (frame 3). In addition, the action is partially occluded, all of this making it more difficult to spot.}
        \label{fig:OREB_tips_example}
    \end{subfigure}
    
    \caption{Example sequences to showcase the difficulty to spot a \textit{a)} defensive rebound and \textit{b)} offensive rebound.}
    \label{fig:dreb_easy_oreb_difficult}
\end{figure}

If we now examine Figure~\ref{fig:spotting_test_barplot}, we observe the same trend when comparing the mAP across all $\delta$ values: the higher the $\delta$, the better the performance. Here, we can even see a very consistent pattern, with the mAP increasing by approximately 0.1 for each additional second in $\delta$. However, a new pattern emerges when analyzing the results by class: the DREB class consistently outperforms the OREB class. This difference may be explained by the fact that, in OREB situations, there are typically more players involved in the play, increasing the likelihood that the ball will be tipped once or twice before a player gains full control. Such intermediate touches may be detected by the model as false positives, negatively impacting the performance for this class. Furthermore, the greater number of players in OREB plays increases the probability that one of them will partial or totally occlude the action, making it harder for the model to identify the exact moment of the rebound compared to DREB plays. Although these situations are not guaranteed to occur in every OREB and may also appear in some DREBs, they are intuitively more frequent in OREB scenarios. Conversely, in certain DREB situations, the ball may go directly to an isolated defensive player, resulting in a clear and uncontested possession. This facilitates the model’s ability to pinpoint the timestamp or frame where the rebound occurs with greater precision. A qualitative example of the situations before described can be found in Figure~\ref{fig:dreb_easy_oreb_difficult}.

\begin{figure}[htbp]
    \centering
    \begin{subfigure}[t]{0.45\textwidth}
        \centering
        \includegraphics[width=\textwidth]{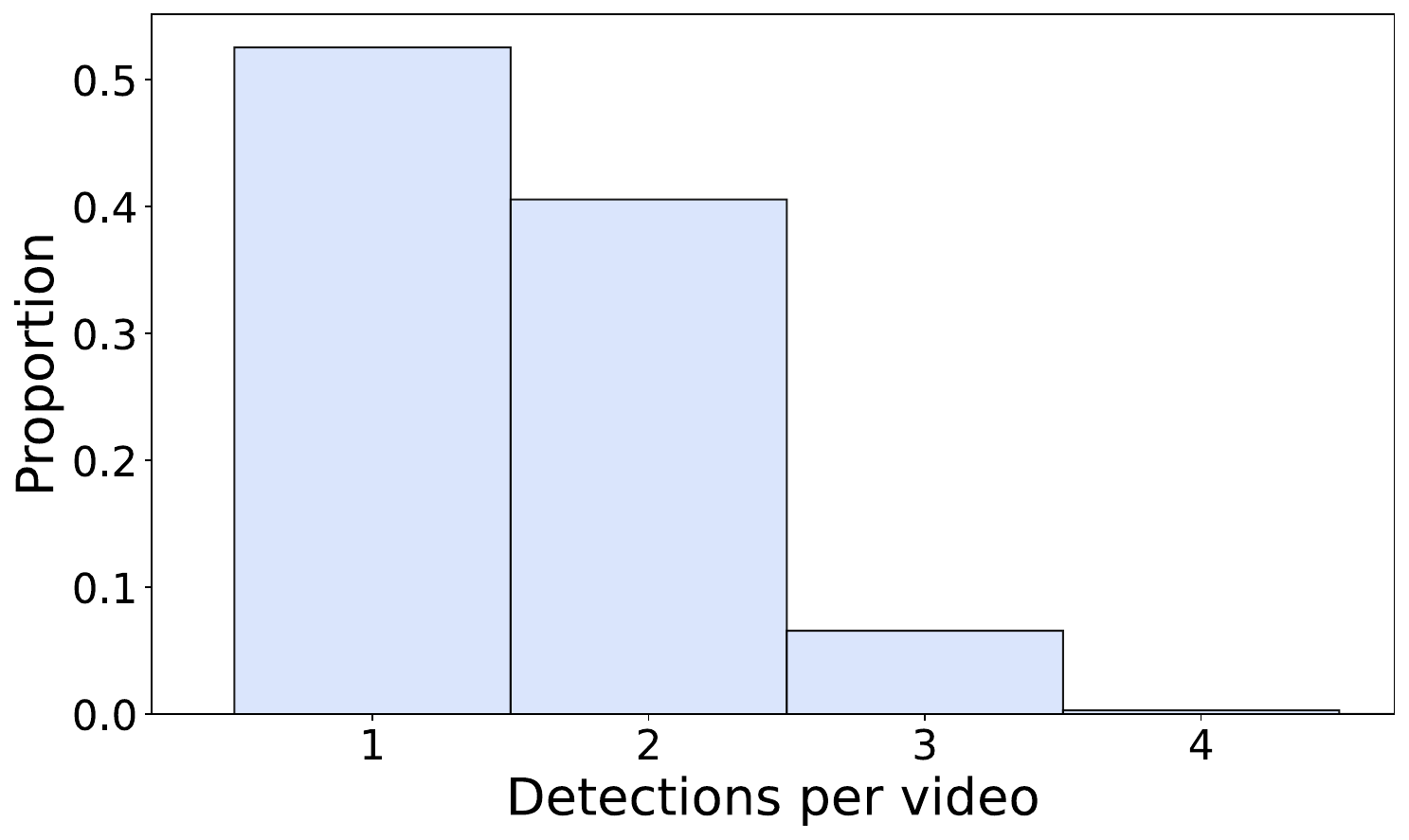}
        \caption{Histogram showing the distribution of detection counts per video in the validation set.}
        \label{fig:spotting_inference_confidences}
    \end{subfigure}%
    \hfill
    \begin{subfigure}[t]{0.45\textwidth}
        \centering
        \includegraphics[width=\textwidth]{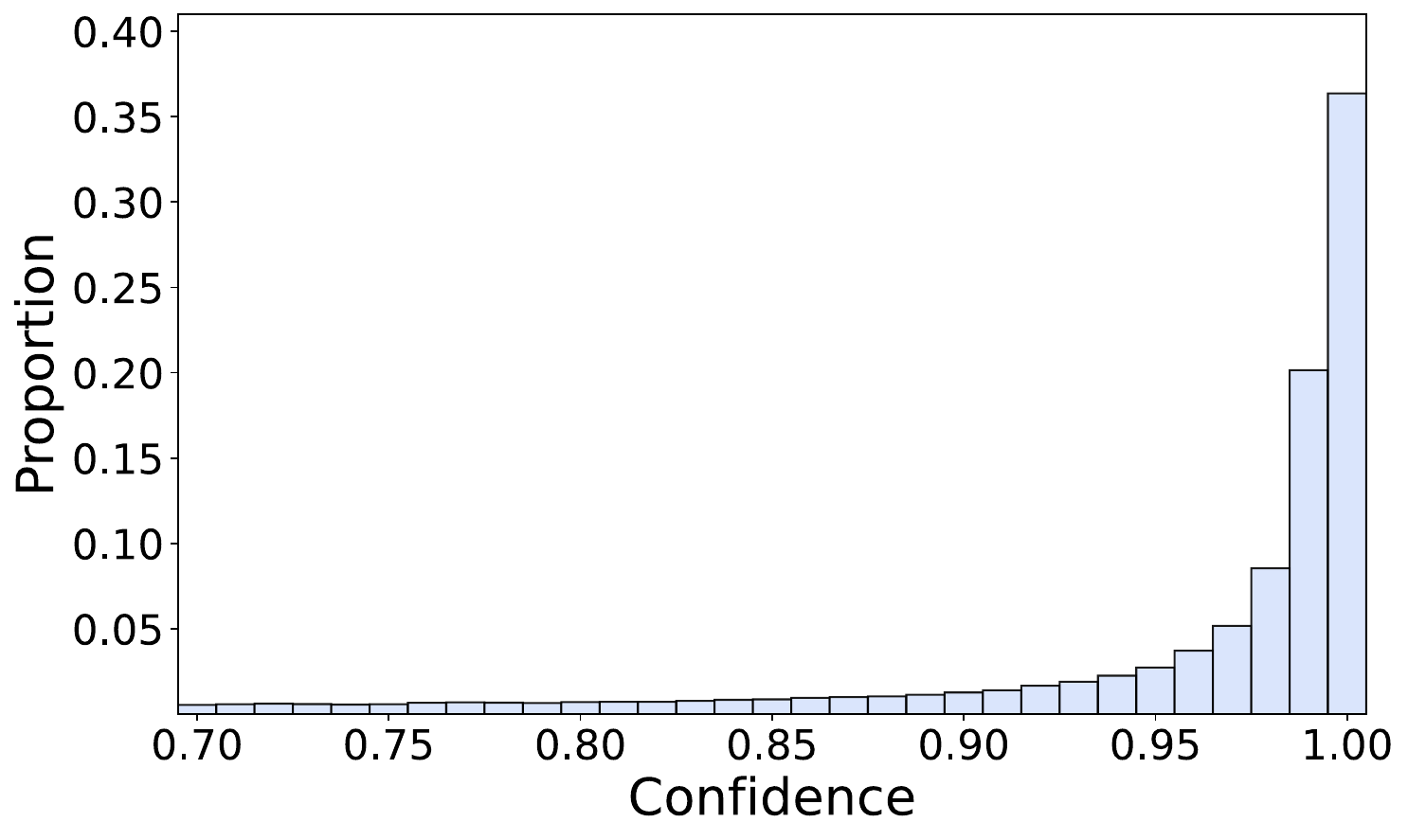}
        \caption{Distribution of model confidence scores across all predicted detections.}
        \label{fig:spotting_inference_num_detections}
    \end{subfigure}
    
    \vspace{0.1cm}
    
    \caption{Statistical analysis of the rebound detections inferred using the fine-tuned spotting model.}
    \label{fig:spotting_inference_stats}
\end{figure}

\paragraph{Inference on unlabeled data} 
Finally, the spotting section concludes with inference on the video sequences that are unlabeled. These sequences are unlabeled in terms of the action timestamp but not regarding the action occurring in the video. Inference is performed on the approximately 75,000 samples that were not used to train the classification models presented in the previous section, ensuring that no data leakage occurs. In Figure~\ref{fig:spotting_inference_stats}, statistics of the inferred samples are presented. First, Figure~\ref{fig:spotting_inference_num_detections} shows the distribution of the number of detections per video across the entire inferred dataset. The first observation is that approximately 50\% of the samples contain a single detection within the whole sequence. Similarly, around 40\% of the samples contain two detections in the same sequence. This may be attributed to false positives, as the raw sequences were initially trimmed by the NBA to include a single rebound.

However, it is also possible that some of the sequences with multiple detections indeed contain more than one rebound, for instance when several missed shots near the basket lead to consecutive rebounds in the same play, or when residual rebounds from a different play are included because the trimming was not precise enough to fully isolate the action. 
Information on how the different play clips available on the NBA website are generated is not publicly available; however, given the large number of videos and the variety of play types, it is highly likely that this is done in a fully automatic manner from raw broadcast footage, probably using a methodology or model similar to the one employed in this section, but naturally more advanced and robust. Therefore, it is not surprising that the trimming is not always perfectly precise or error-free. Finally, samples with three or more detections are rare, representing less than 10\% of the total.

Second, Figure~\ref{fig:spotting_inference_confidences} shows the distribution of confidences for the different detections across all samples. The distribution exhibits a clear pattern: high-confidence detections dominate, with approximately 80\% of the detections having a confidence above 0.95. This indicates that the model is highly decisive in most cases, assigning high probability scores to its predictions. Nevertheless, the long tail of lower-confidence detections reveals that ambiguous situations still occur, which may correspond to visually challenging plays, occlusions, or cases where the trimming of the clips introduces noise.

Since these samples will be combined with manually annotated ones, consistency between both sources of manually labeled and pseudo-labeled data is required. In the manually annotated samples, only a single action per video was considered; if more than one action occurred, only the first was annotated. To ensure consistency, the inferred samples are filtered to remove videos with multiple detections. This step also reduces the likelihood of including false positives. Furthermore, a confidence-based filter is applied, retaining only those samples with a confidence score $\geq 0.99$. After applying both filters, 25{,}000 pseudo-labeled samples are obtained from the original 75{,}000 inferred ones. An analysis of the class distribution within this subset reveals a marked imbalance: 17{,}000 belong to the DREB class, while only 8{,}000 correspond to OREB. Given that the original 75{,}000 unlabeled samples were evenly distributed, this outcome suggests either that DREB actions are easier to detect (resulting in higher confidence scores) or that OREB samples are more prone to multiple detections. In any case, the objective is to construct a balanced pseudo-labeled dataset. To this end, all 8{,}000 OREB samples are retained, and 8{,}000 are randomly selected from the 17{,}000 DREB samples, as all have already passed strict filtering criteria. This process yields a final set of \textbf{16,000 pseudo-labeled samples}, equally balanced across both classes, each containing a single annotated action with its class label as ground truth and its timestamp inferred by the trained spotting model.

For now, these pseudo-labeled samples will be set aside, and we will get back to them in the upcoming Section \ref{subsec:increase num samples}, where we will see if adding the pseudo-samples to the training split i.e. increasing the number of training samples, does help the model to perform better on the offline action anticipation task.

\subsection{Offline Action Anticipation}
For the offline setup, a video trimmed up to $\tau_a$ seconds before the action occurs is used as input. This section presents an ablation study on the $\tau_a$ parameter, an interpretability analysis comparing the activation maps between models trained at different $\tau_a$, an experiment evaluating the impact of the number of training samples (including pseudo-labeled samples obtained with the previously described spotting model), and a comparison between the performance of the anticipation model and that of a group of human experts. Remember that in this section, only the results of the \textit{baseline} model are presented and analyzed, as adding the encoder layers from \textit{TEAM} did not significantly improve performance. However, the results of the experiments using \textit{TEAM} can be found in the Appendix of this work (Section~\ref{sec:appendix}).

\subsubsection{Changing the anticipation time \texorpdfstring{$\tau_a$}{}} \label{sec:changing t_a}

\begin{figure}[htbp]
    \centering
    \begin{subfigure}[t]{0.5\textwidth}
        \centering
        \includegraphics[width=\textwidth]{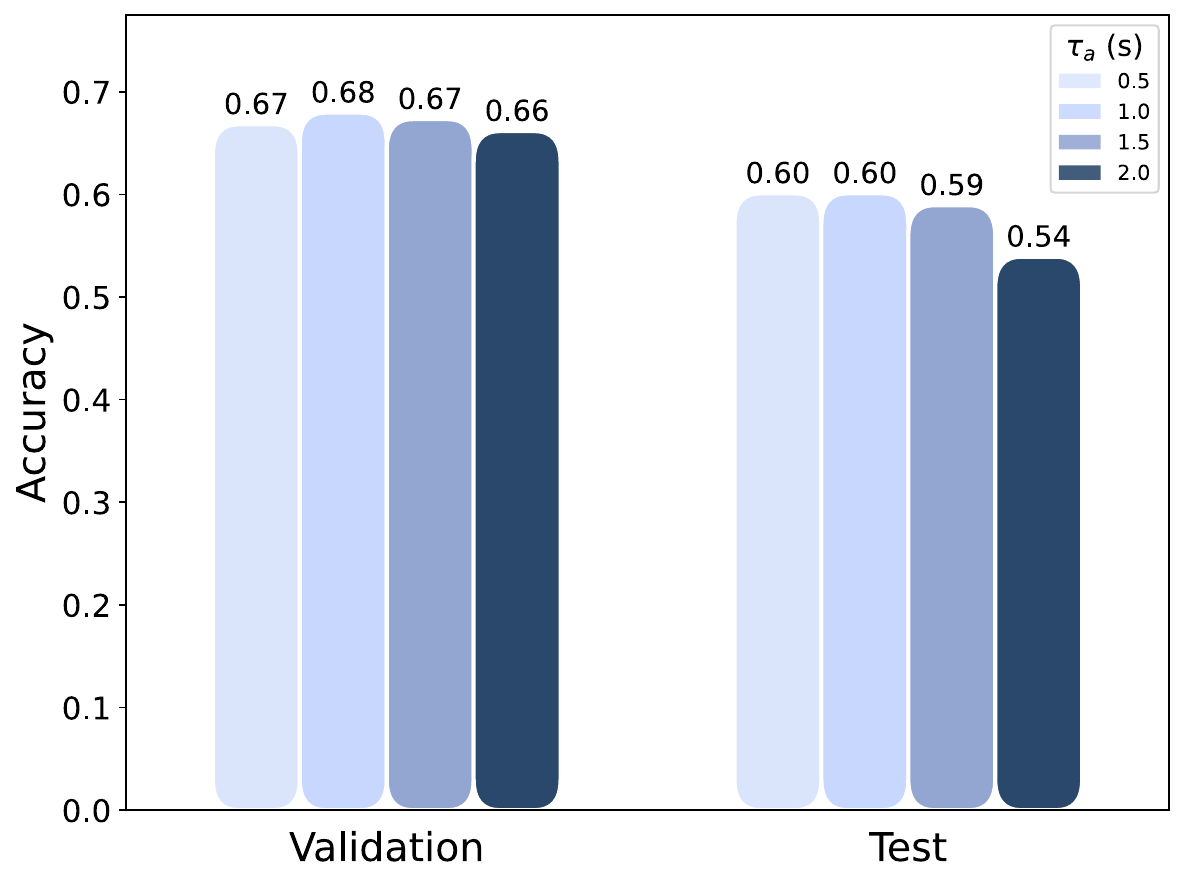}
        \caption{Barplot comparing the accuracies for different anticipation times both in the validation and test splits.}
        \label{fig:offline_anticipation_time_barplot}
    \end{subfigure}%
    \hfill
    \begin{subfigure}[t]{0.44\textwidth}
        \centering
        \includegraphics[width=\textwidth]{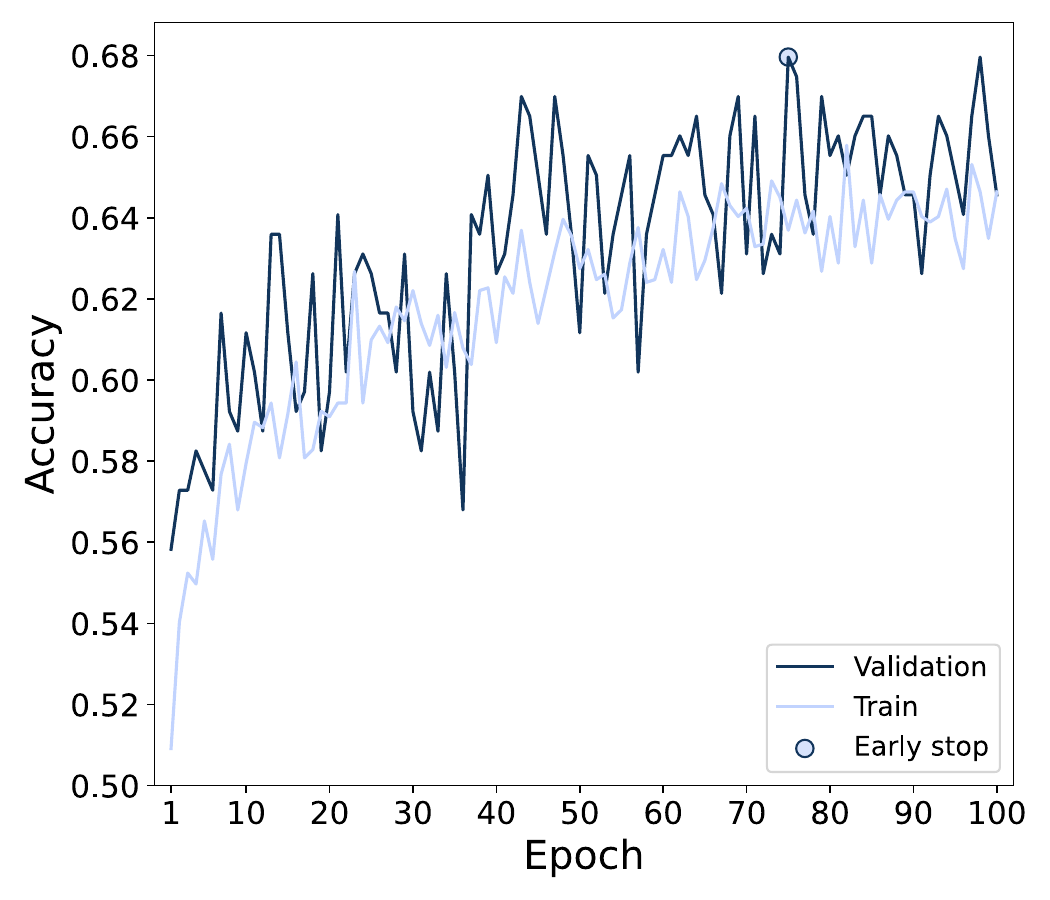}
        \caption{Evolution of accuracy in the training (light blue) and validation (dark blue) splits for the particular case of $\tau_a=0.5$s.}
        \label{fig:offline_anticipation_time_evo}
    \end{subfigure}
    
    \vspace{0.1cm}
    
    \caption{Results for the experiment of varying the anticipation time ($\tau_a$).}
    \label{fig:offline_anticipation_time}
\end{figure}

\begin{figure}[htbp]
    \centering
    \begin{subfigure}[t]{0.5\textwidth}
        \centering
        \includegraphics[width=\textwidth]{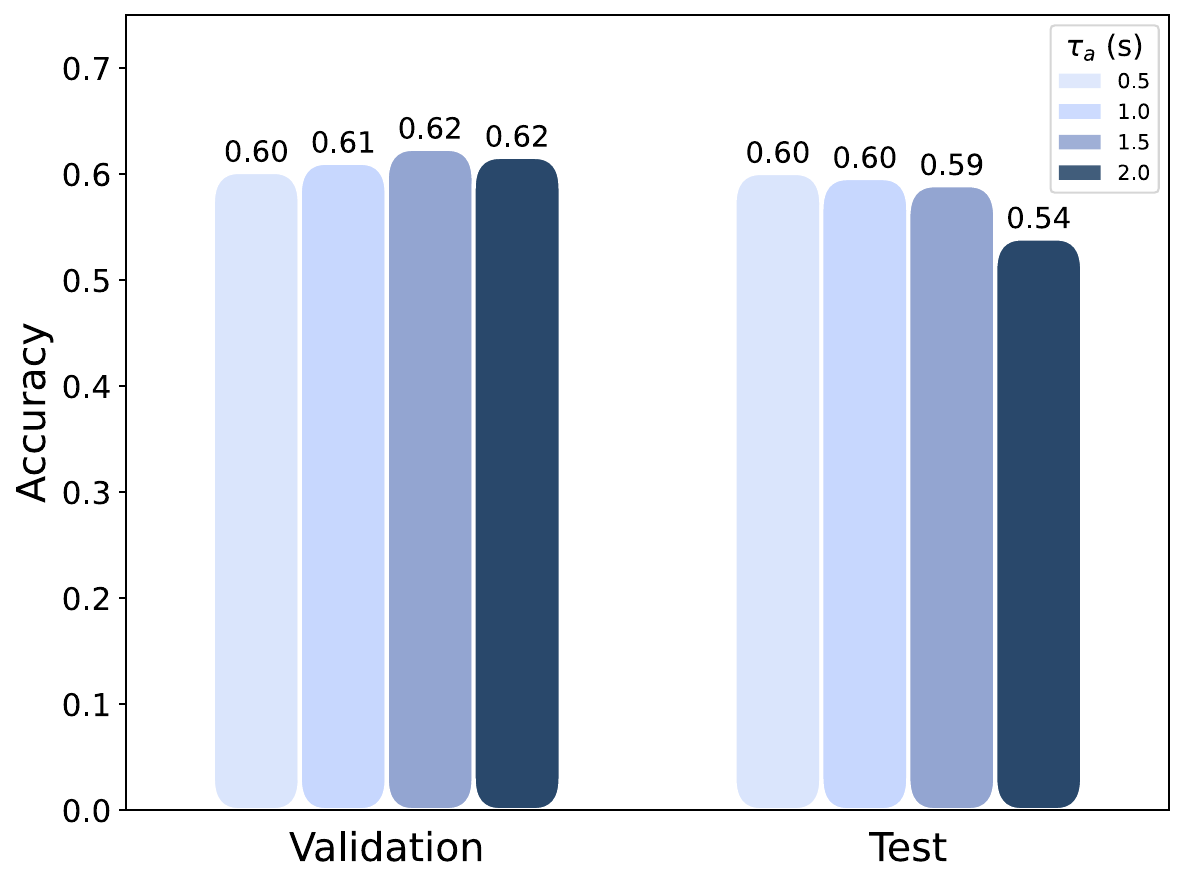}
        \caption{Barplot comparing the accuracies for different anticipation times both in the validation and test splits.}
    \end{subfigure}%
    \hfill
    \begin{subfigure}[t]{0.45\textwidth}
        \centering
        \includegraphics[width=\textwidth]{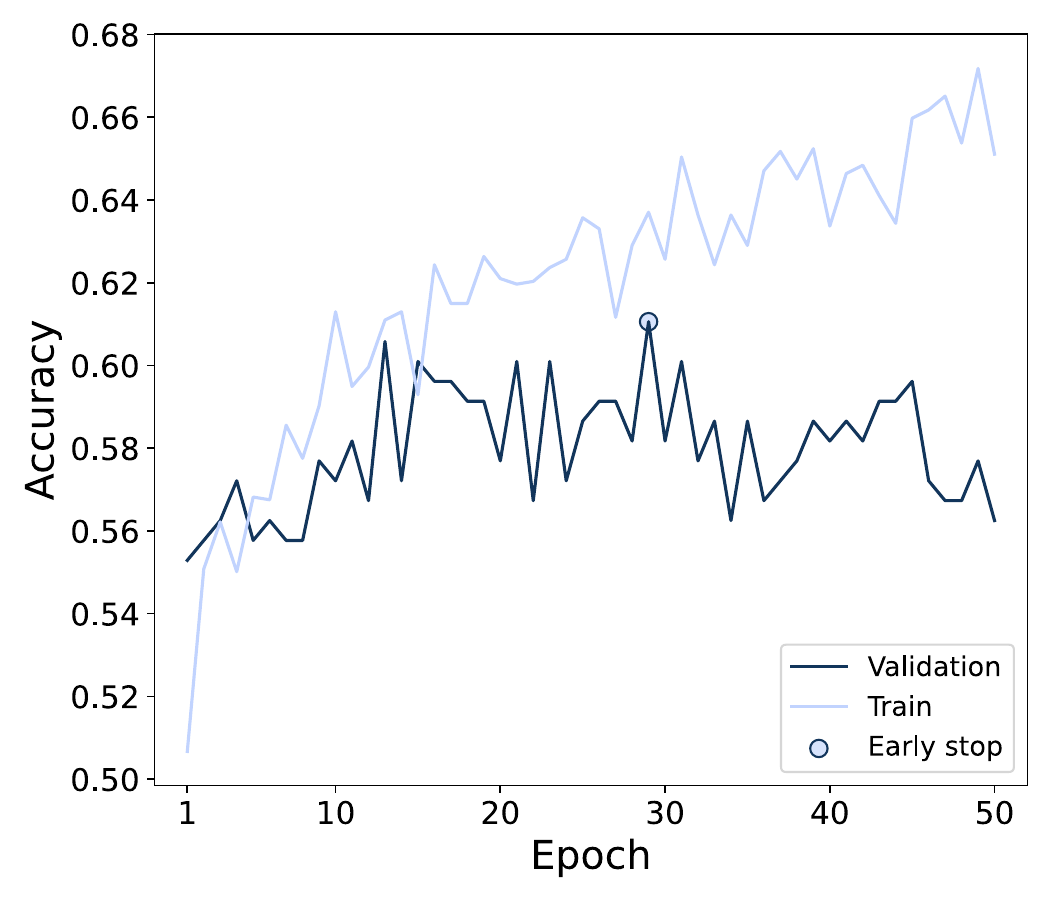}
        \caption{Evolution of accuracy in the training (light blue) and validation (dark blue) splits for the particular case of $\tau_a=0.5$s.}
    \end{subfigure}
    
    \vspace{0.1cm}
    
    \caption{Results for the experiment of varying the anticipation time ($\tau_a$) using \textit{TEAM}.}
    \label{fig:offline_anticipation_time_transformer}
\end{figure}

In Figure~\ref{fig:offline_anticipation_time}, the final results obtained for different anticipation times are presented, using the \textit{baseline} model. Remember that the results for \textit{TEAM} can be found in the Appendix of this work (Section~\ref{sec:appendix}, Figure~\ref{fig:offline_anticipation_time_transformer}). Figure~\ref{fig:offline_anticipation_time_barplot} compares the model's performance at different anticipation times using the fine-tuned X3D model. Validation accuracies are consistently higher than those on the test set, suggesting mild overfitting to the validation data. This is expected given the relatively small size of the validation set (only 250 samples) and the high variability observed across epochs. On the test set, the expected trend is observed: accuracies decrease slightly as the anticipation time increases, reflecting the intuitive notion that predicting further in advance is more challenging due to the broader range of possible developments in the scene.

To address the overfitting observed on the validation set, cross-validation could have been employed. However, this approach was not adopted, as it would have limited other analyses presented in this work. Using a small number of folds, such as 2 or 3, would not substantially increase training time but would reduce the number of training samples, potentially impacting model performance. Conversely, employing a larger number of folds would mitigate the issue of limited training samples and, although the validation set would remain small — resulting in high variability — multiple splits would provide a better estimate of the mean and standard deviation. Nevertheless, this approach would substantially increase the training time for each experiment, as each model would need to be trained 4 or 5 times for each anticipation time $\tau_a$.

As previously discussed, it is expected that increasing the anticipation time $\tau_a$ makes the prediction task more challenging, as the model has access to less temporal context and the uncertainty and variability of plausible future scenarios are greater. To provide intuition on the difficulty of predicting 0.5s versus 1s before the action, Figure~\ref{fig:anticipation_time_example} shows the last frame observed by the model prior to making a prediction. In general, for a 0.5s anticipation, the ball has typically already bounced off the rim or backboard, allowing the model to infer its trajectory. In contrast, 1s before the action, the ball is usually still in the air or even in the hands of the player, requiring the model to rely on additional cues to make an accurate prediction.

Figure~\ref{fig:offline_anticipation_time_evo} shows the training and validation curves for the model trained with $\tau_a=0.5$s. The curves for other anticipation times are omitted due to space constraints and because they exhibit similar behavior. The main observation is the difference in variability between the training and validation curves. While the training curve stabilizes around an accuracy of approximately 0.64 after epoch 60, the validation curve, although showing a clear upward trend, exhibits considerably higher variability. This variability is especially pronounced in the initial epochs and decreases toward the end of training, yet remains significantly larger than in the training curve. Attempts to reduce the learning rate resulted in poorer convergence, with final accuracies of 0.64 for validation and 0.56 for training. Consequently, the higher learning rate was maintained, accepting the associated increase in variability in the validation split. Using a larger batch size was not feasible due to GPU memory limitations.

\subsubsection{Interpretability}
\label{sec:interpretability_experiment}

This section analyzes how activation maps vary with different anticipation times ($\tau_a$) used during model training. A larger $\tau_a$ corresponds to a more restrictive setting, as the model is provided with less temporal context. The activation maps are normalized across the entire video, as shown in Equation~\ref{eq:cam_normalization}, to highlight both the frames most relevant for prediction and the spatial regions where the model focuses.

\begin{equation} \label{eq:cam_normalization}
\text{CAM}_{\text{norm}}[t,h,w] = 
\frac{\text{CAM}[t,h,w] - \min_{t',h',w'} \text{CAM}[t',h',w']}
{\max_{t',h',w'} \text{CAM}[t',h',w'] - \min_{t',h',w'} \text{CAM}[t',h',w'] + \epsilon}
\end{equation}

Here, $\epsilon = 10^{-8}$ is used to avoid division by zero, $t, h, w$ correspond to the time, height and width dimensions and $\min_{t',h',w'} \text{CAM}$ and $\max_{t',h',w'} \text{CAM}$ correspond to the minimum and maximum activation values across the entire video, respectively. 

\begin{figure}[htbp]
    \begin{subfigure}[b]{0.25\textwidth}
        \includegraphics[width=\textwidth]{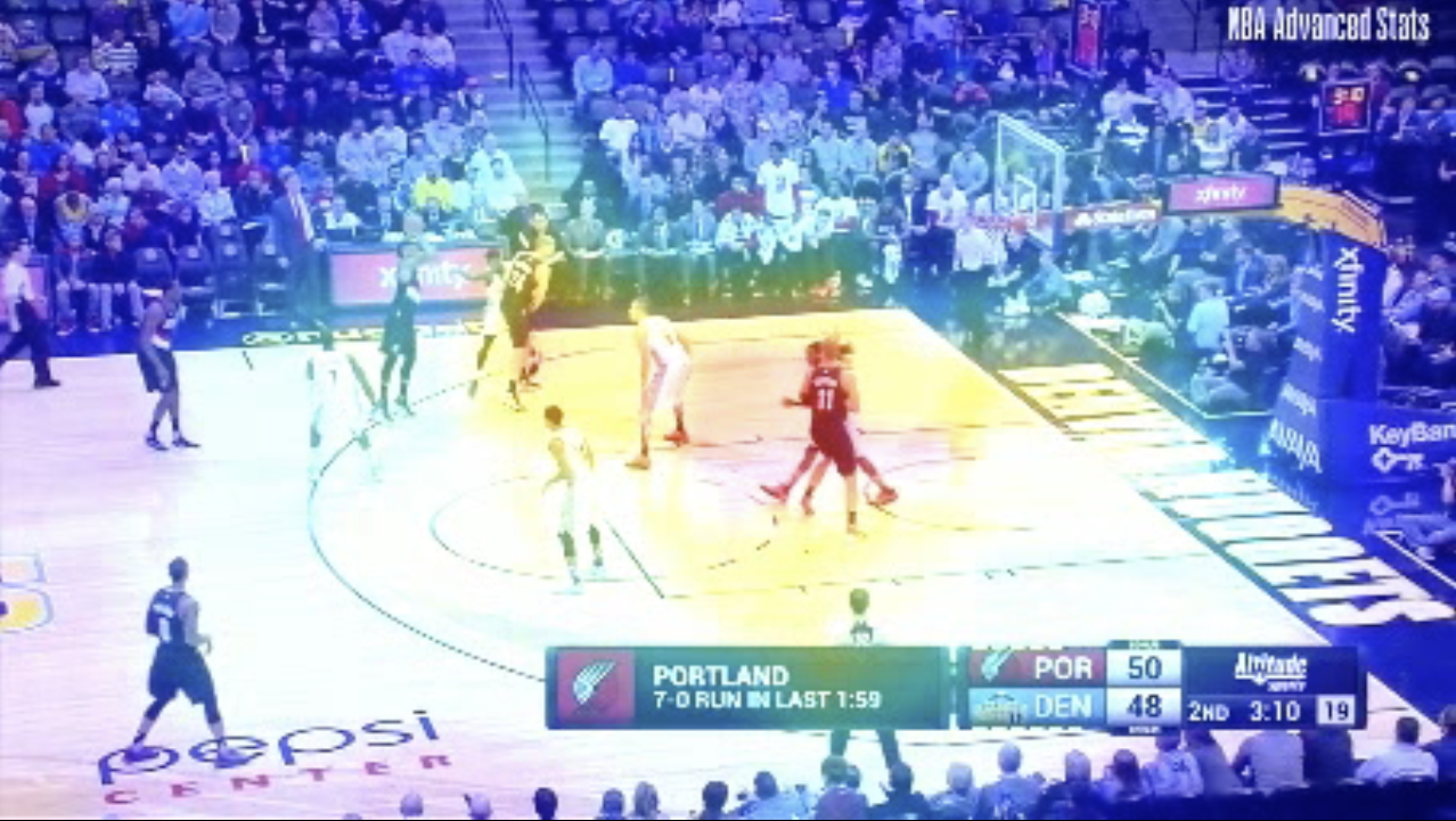}
        \caption{$\tau_a = 2.0$s. Frame 47/48.}
        \label{fig:max_activation_2}
    \end{subfigure}%
    \begin{subfigure}[b]{0.25\textwidth}
        \includegraphics[width=\textwidth]{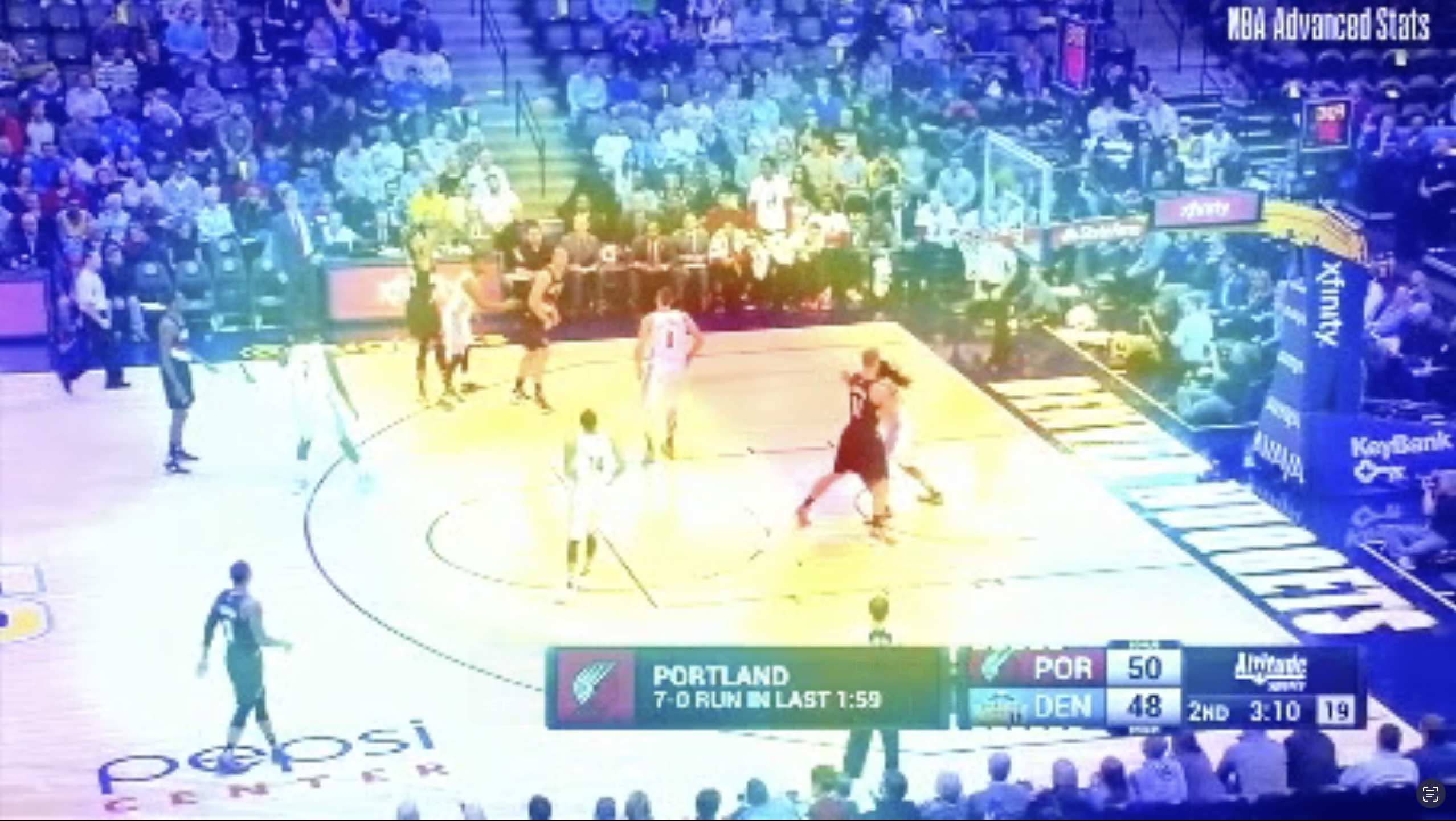}
        \caption{$\tau_a = 1.5$s. Frame 53/54.}
        \label{fig:max_activation_1_5}
    \end{subfigure}%
    \begin{subfigure}[b]{0.25\textwidth}
        \includegraphics[width=\textwidth]{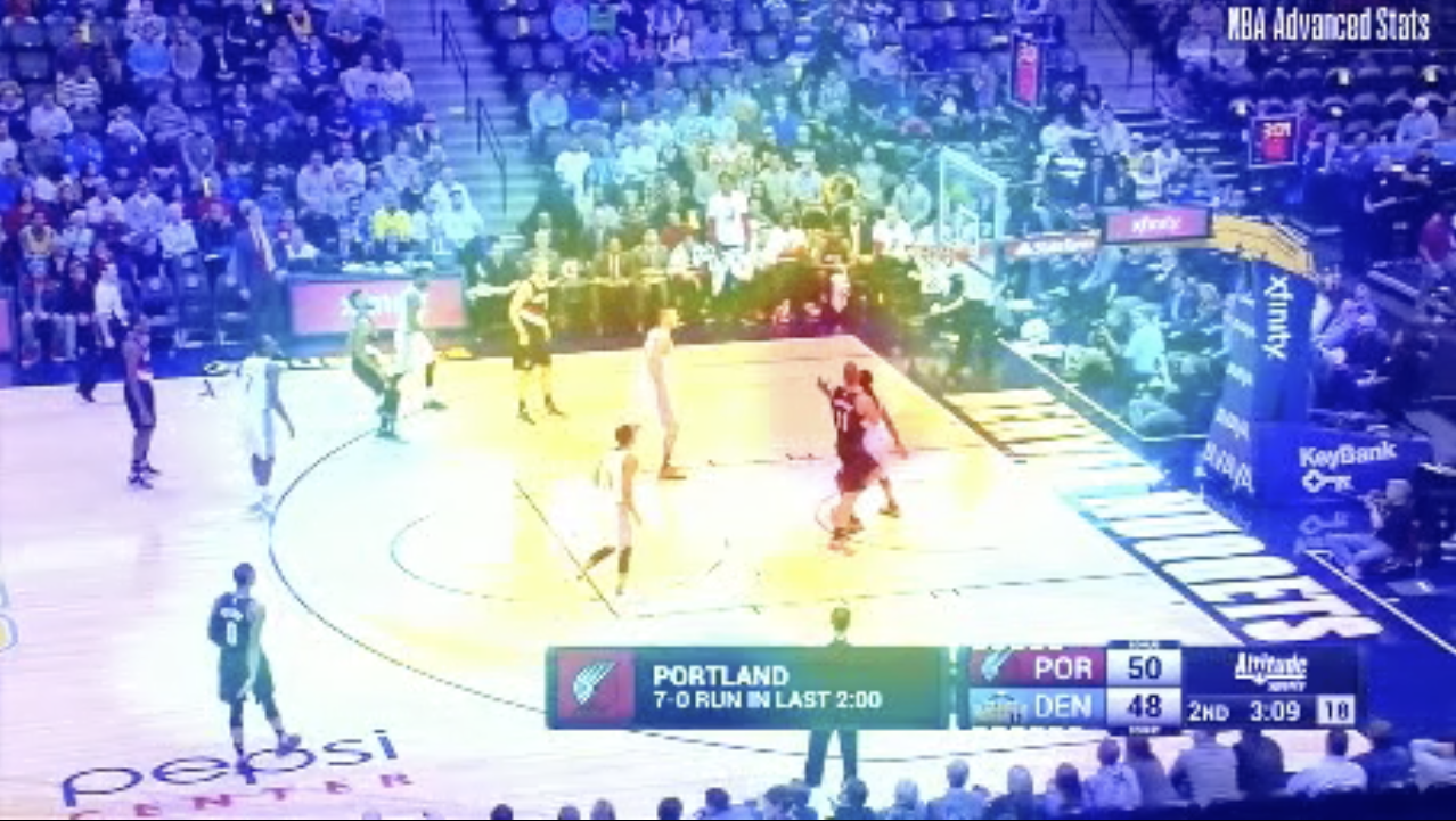}
        \caption{$\tau_a = 1.0$s. Frame 59/60.}
        \label{fig:max_activation_1}
    \end{subfigure}%
    \begin{subfigure}[b]{0.25\textwidth}
        \includegraphics[width=\textwidth]{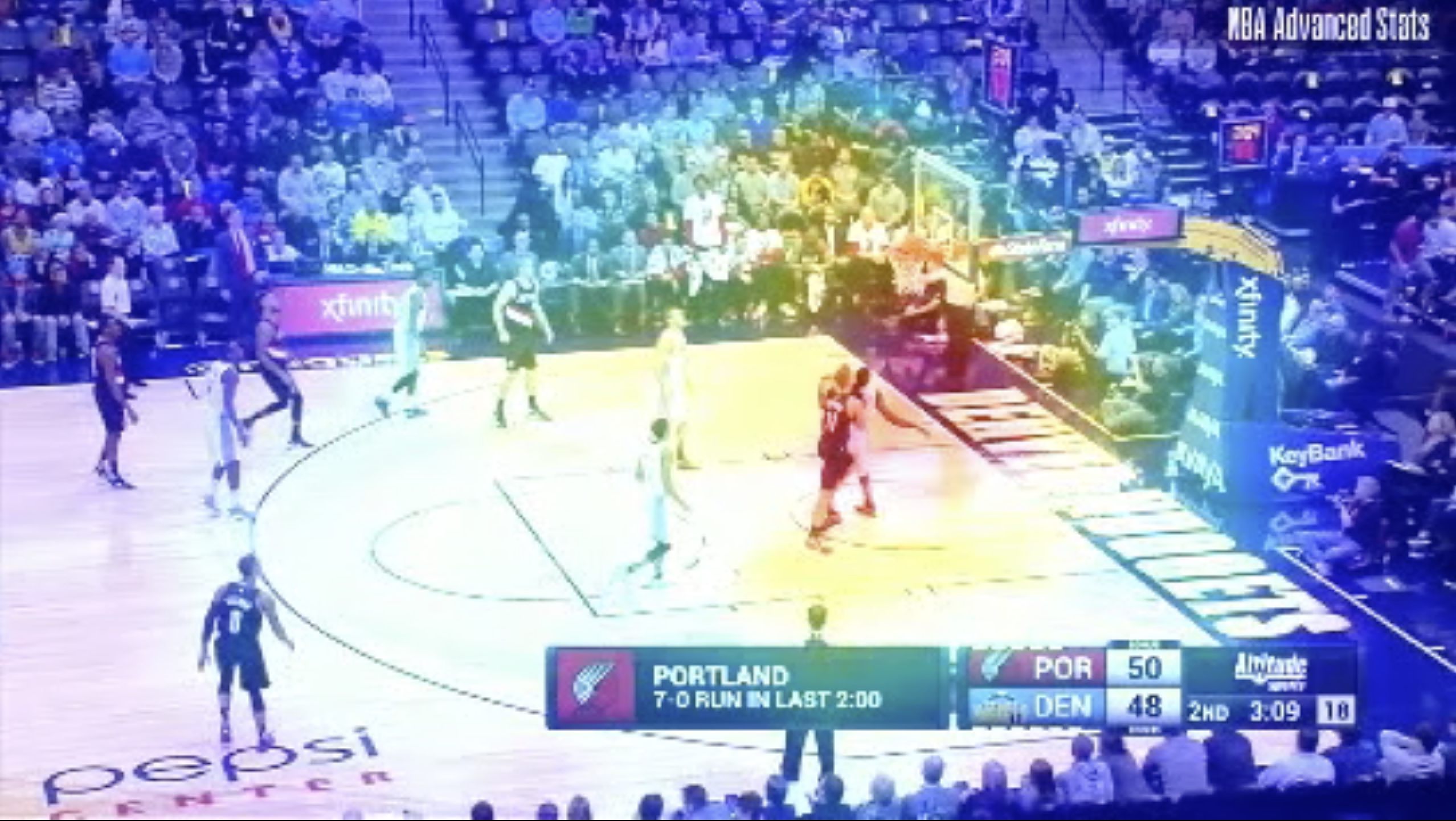}
        \caption{$\tau_a = 0.5$s. Frame 64/66.}
        \label{fig:max_activation_0_5}
    \end{subfigure}
    
    \caption{Comparison of frames with the highest activation for models trained at different $\tau_a$: a) 2.0s, b) 1.5s, c) 1.0s, and d) 0.5s. Each subcaption specifies which is the frame containing the maximum activation with respect to the temporal context available for that specific model. See the complete activation maps across an example sequence \href{https://youtu.be/wOyMdw9xFMg}{here}.}
    \label{fig:gradcam_max_activation}
\end{figure}

After a detailed qualitative analysis of the activation maps on the test set, the first general observation is that the maximum activations within a video tend to occur at the end of the sequence. Figure~\ref{fig:gradcam_max_activation} shows the frame with the highest activation across the entire sequence for different models trained at a different $\tau_a$ for a given example play. We see that, regardless of the anticipation time $\tau_a$, the maximum activations consistently appear in the last two or three frames of the temporal context provided. This suggests that the model relies heavily on the final frames to make its prediction, which is intuitive in a fast-changing and highly uncertain environment where the state of the game (including the positions of attackers, defenders, and the ball) can change dramatically within just a few frames. It should be noted, however, that this does not imply the model exclusively attends to the last frames, since relevant activations also appear earlier in the sequence. Rather, it indicates that the highest-intensity activations are concentrated near the end of the video.

\begin{figure}[htbp]
    \begin{subfigure}[b]{0.25\textwidth}
        \includegraphics[width=\textwidth]{figures/results/offline_anticipation/interpretability/last_frame_2.0s.png}
        \caption{$\tau_a = 2.0$s. Frame 47/48.}
        \label{fig:last_frame_2}
    \end{subfigure}%
    \begin{subfigure}[b]{0.25\textwidth}
        \includegraphics[width=\textwidth]{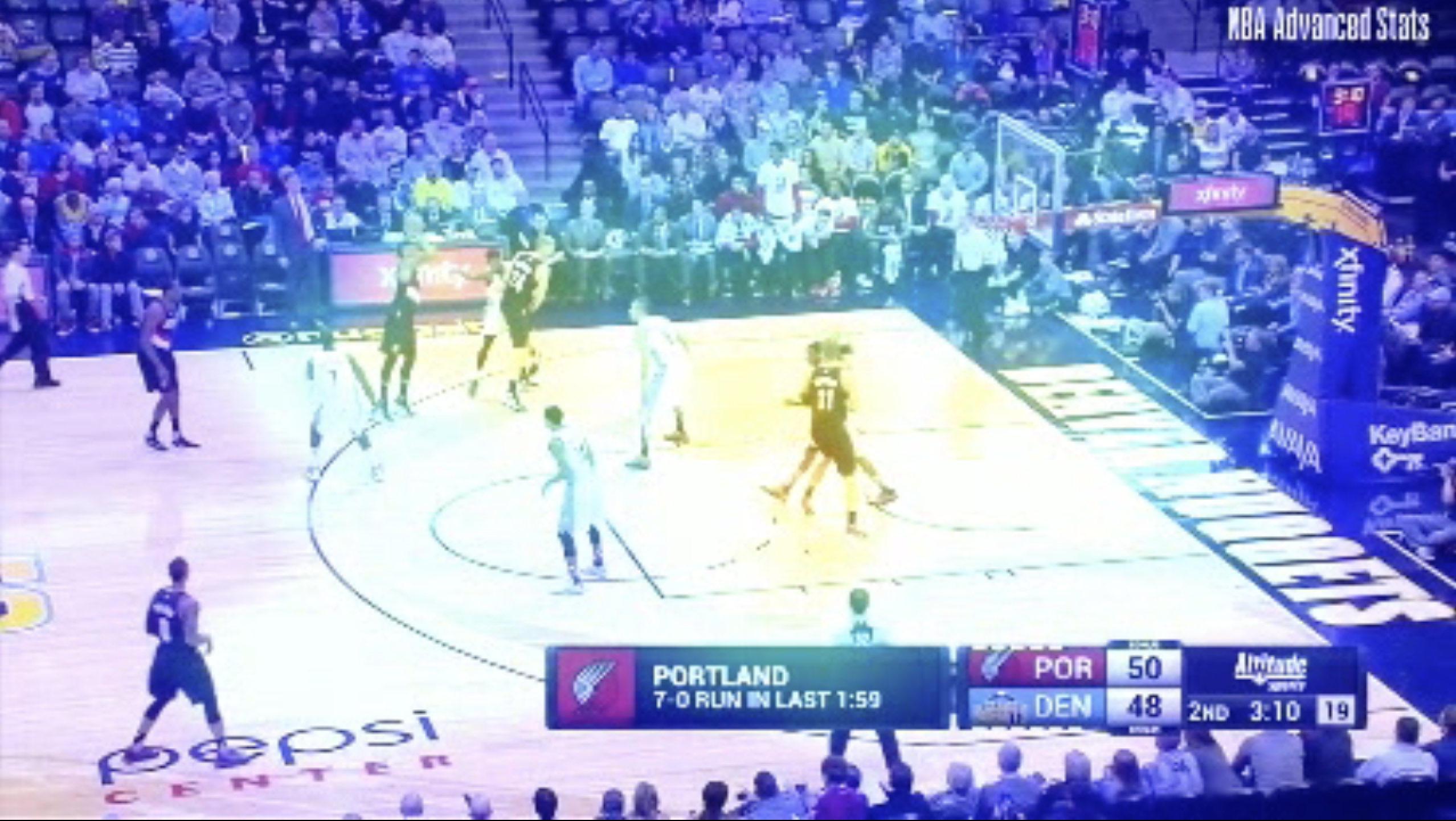}
        \caption{$\tau_a = 1.5$s. Frame 47/54.}
        \label{fig:last_frame_1_5}
    \end{subfigure}%
    \begin{subfigure}[b]{0.25\textwidth}
        \includegraphics[width=\textwidth]{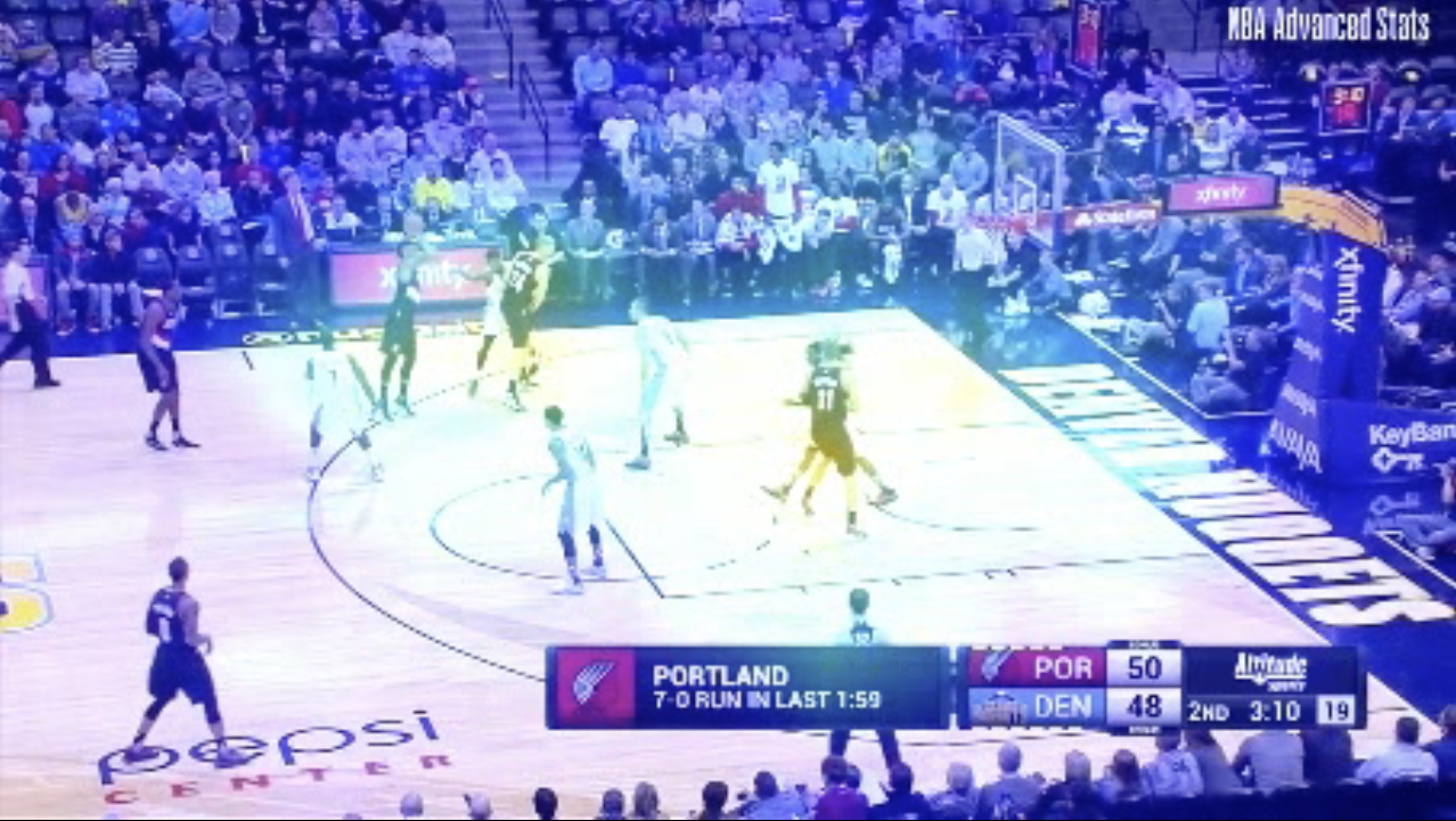}
        \caption{$\tau_a = 1.0$s. Frame 47/60.}
        \label{fig:last_frame_1}
    \end{subfigure}%
    \begin{subfigure}[b]{0.25\textwidth}
        \includegraphics[width=\textwidth]{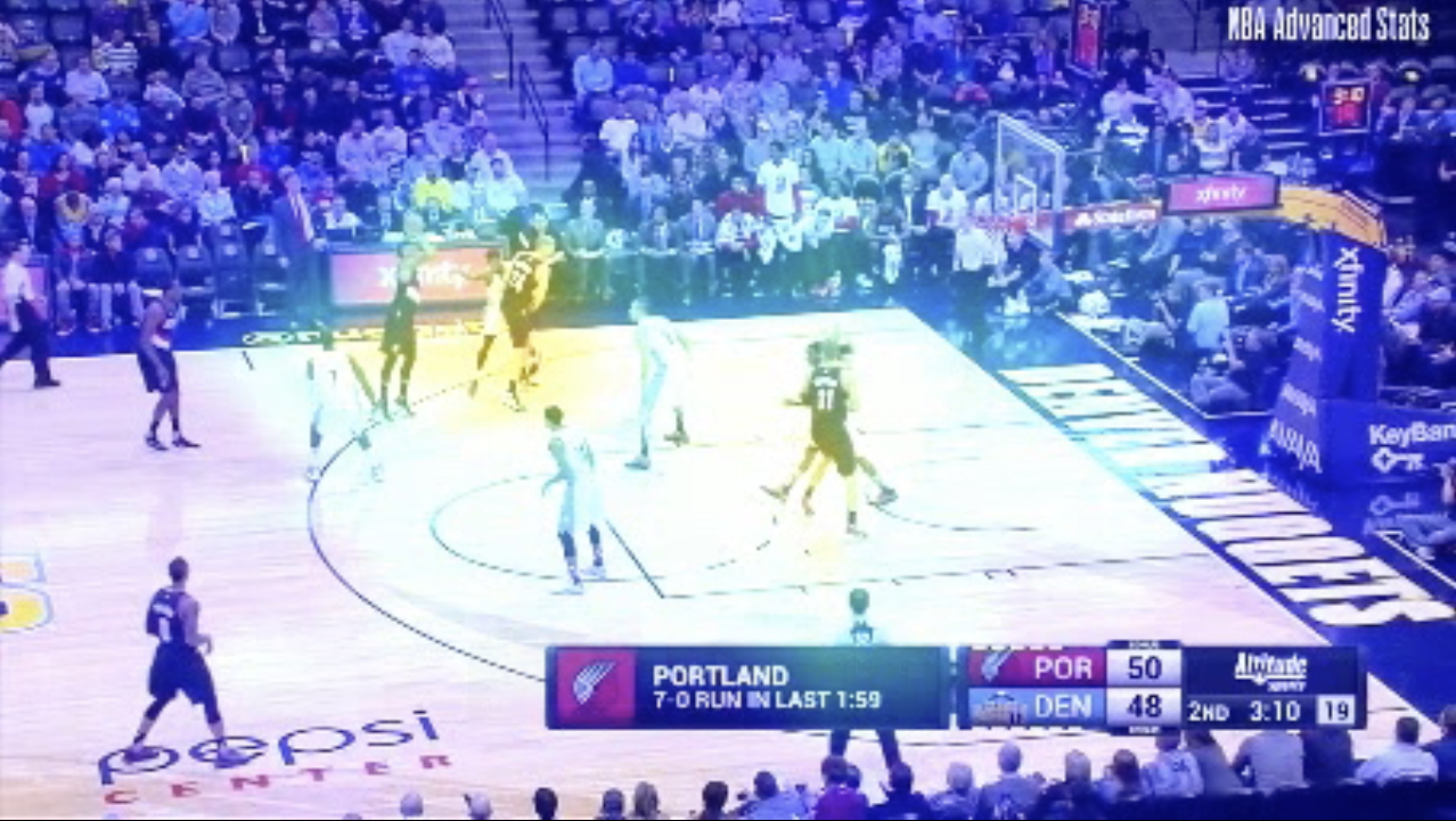}
        \caption{$\tau_a = 0.5$s. Frame 47/66.}
        \label{fig:last_frame_0_5}
    \end{subfigure}
    
    \caption{Comparison of activation maps from models trained with different anticipation times $\tau_a$: a) 2.0s, b) 1.5s, c) 1.0s, and d) 0.5s, all shown on the same frame. The chosen frame corresponds to the last frame within the temporal context of the most restrictive model, i.e., the one trained with $\tau_a=2.0$s.}
    \label{fig:gradcam_last_frame}
\end{figure}

To complement this analysis, Figure~\ref{fig:gradcam_last_frame} shows a comparison of activation maps across models trained with different values of $\tau_a$, but evaluated on the same frame. The selected frame corresponds to the last available frame in the temporal context sequence of the model trained with $\tau_a=2.0$. Although this frame is also included in the temporal context sequences of models trained with smaller $\tau_a$ values, in those cases it does not correspond to the final frame. As a result, the activation on player \#11 (black jersey) appears lower (closer to yellow–blue than to red) when $\tau_a$ decreases, since the model has not yet reached its final decision about which player or team will secure the ball, given that additional temporal context is still available.

Another interesting observation is that training a model with a smaller $\tau_a$, i.e., providing it with more temporal context, does not necessarily result in higher confidence in its predictions compared to a model with less temporal context when evaluating the same video. Given the fast-paced nature of the environment, stopping the video 1.5s before the action may lead to a situation in which the defense appears to be in control, yielding a highly confident prediction. However, subsequent frames may introduce critical changes (e.g., a player repositioning to gain a better chance at securing the rebound), which could lead the model trained with $\tau_a = 0.5$s to produce a less confident prediction.

Finally, when analyzing the visual cues attended by the model, it is particularly notable that, in general, the models do not focus strongly on the ball. Although this can be intuited from Figures~\ref{fig:gradcam_max_activation} and \ref{fig:gradcam_last_frame}, where the activations are not centered on the ball, this behavior can be more clearly observed in the example sequence provided \href{https://youtu.be/wOyMdw9xFMg}{here}. While this is just a single example, the same pattern is observed across the test set, regardless of the sequence analyzed or the anticipation time $\tau_a$ used for training. Therefore, no explicit “tracking” of the ball is observed in the activation maps; instead, the maps seem to capture the player with the ball or other players involved in the play. This behavior will later be further analyzed and compared with that of human experts.

\subsubsection{Augmenting training with pseudo-labeled data} \label{subsec:increase num samples}
\begin{figure}[htbp]
    \begin{subfigure}[t]{0.49\textwidth}
        \centering
        \includegraphics[width=\textwidth]{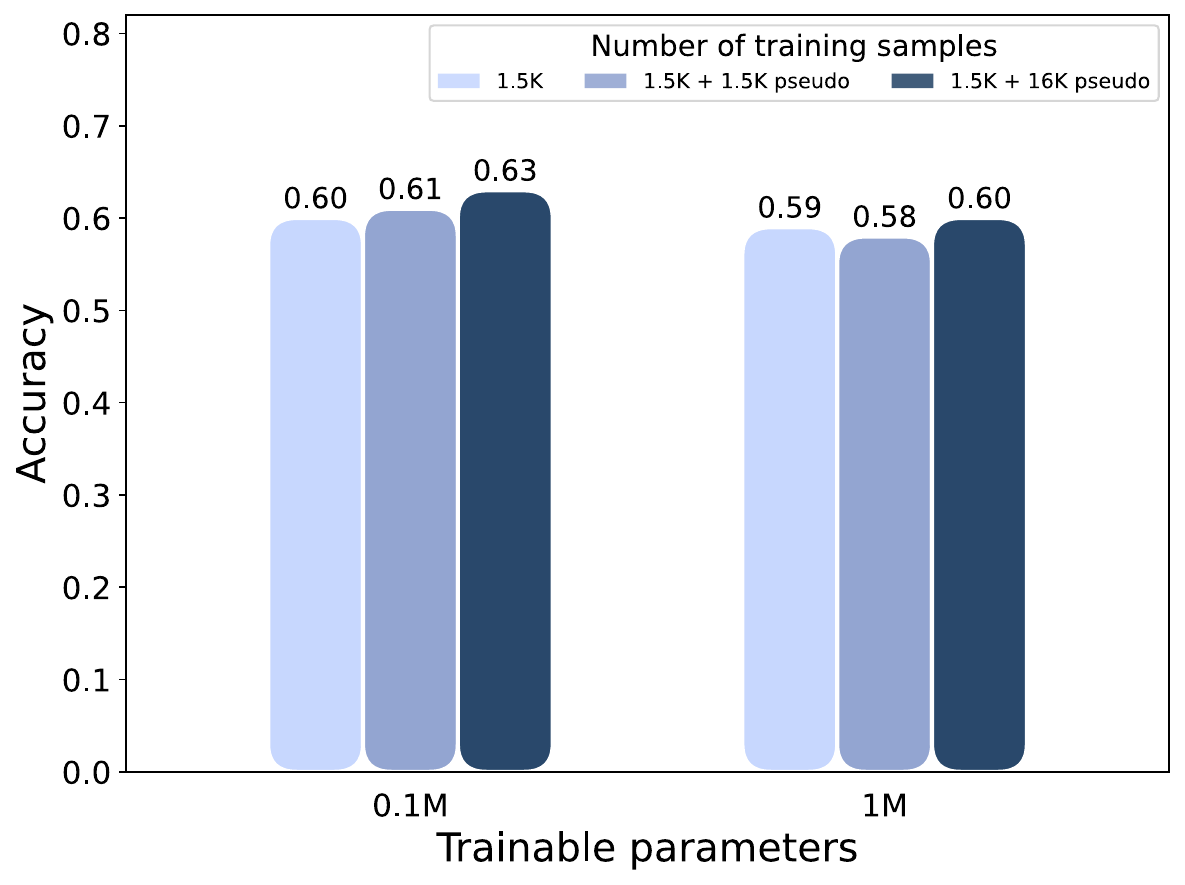}
        \caption{Test split accuracies for different number of training samples and trainable parameters.}
    \end{subfigure}%
    \hfill
    \begin{subfigure}[t]{0.49\textwidth}
        \centering
        \includegraphics[width=\textwidth]{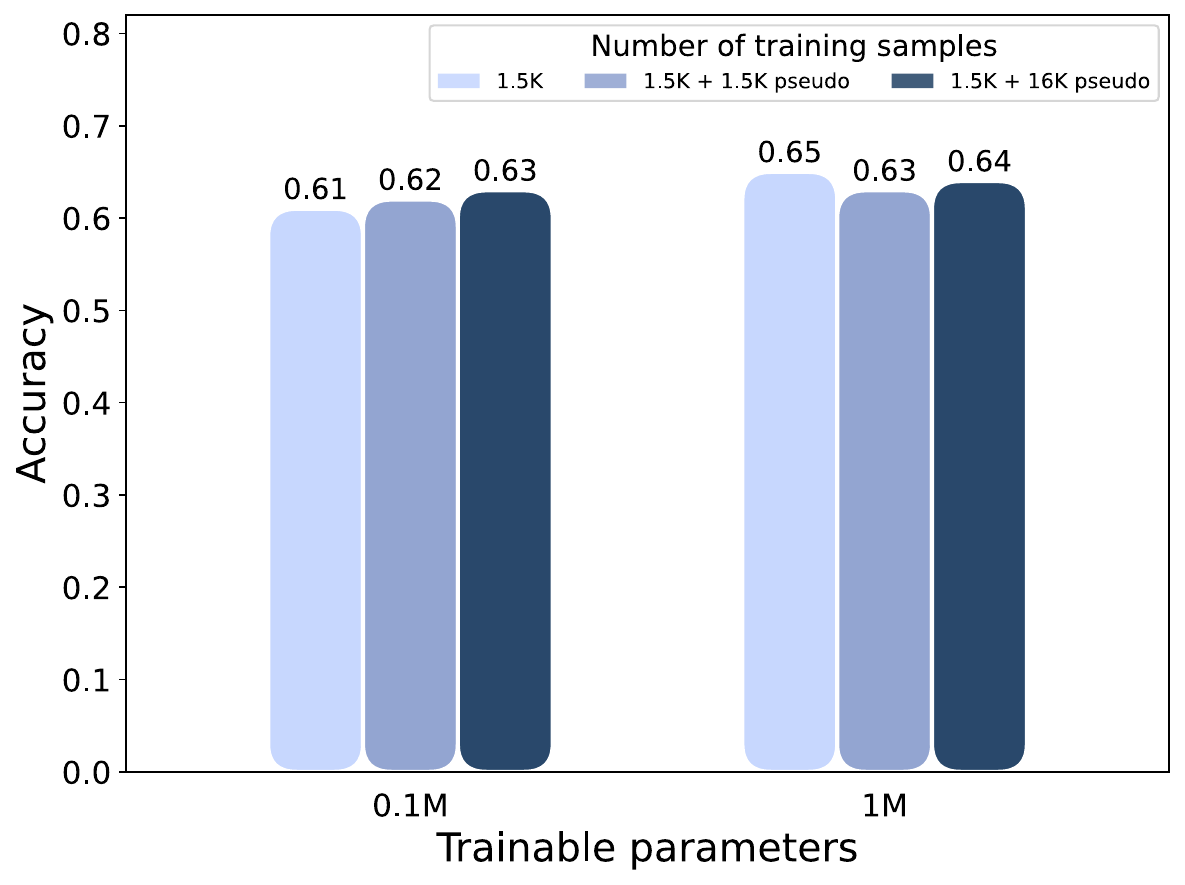}
        \caption{Validation split accuracies for different number of training samples and trainable parameters.}
    \end{subfigure}
    
    \vspace{0.1cm}
    
    \caption{Results using \textit{TEAM} for the experiment of increasing the number of training samples via action spotting pseudo-labeling both on the a) test and b) validation splits. All experiments are done at $\tau_a=1.0$s.}
    \label{fig:offline_anticipation_increasing_number_samples_transformer}
\end{figure}

This experiment investigates whether adding additional training samples through action spotting pseudo-labeling leads to performance improvements using the \textit{baseline} model. Remember that the results for this exact same experiment using \textit{TEAM} can be found in the Appendix of this work (Section~\ref{sec:appendix}, Figure~\ref{fig:offline_anticipation_increasing_number_samples_transformer}). 

\begin{figure}[htbp]
    \centering
    \begin{subfigure}[t]{0.44\textwidth}
        \centering
        \includegraphics[width=\textwidth]{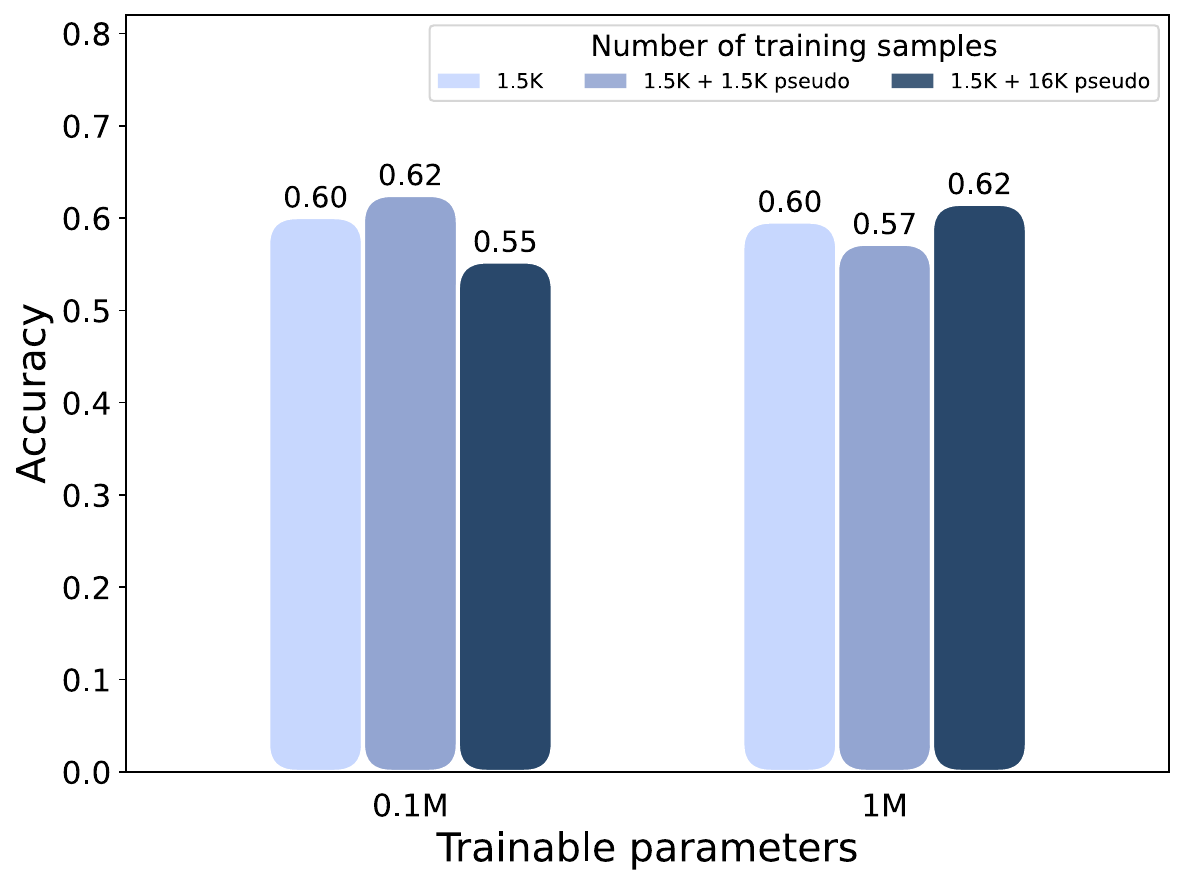}
        \caption{Test split accuracies for different number of training samples and trainable parameters.}
        \label{fig:offline_anticipation_num_samples_test}
    \end{subfigure}%
    \hfill
    \begin{subfigure}[t]{0.44\textwidth}
        \centering
        \includegraphics[width=\textwidth]{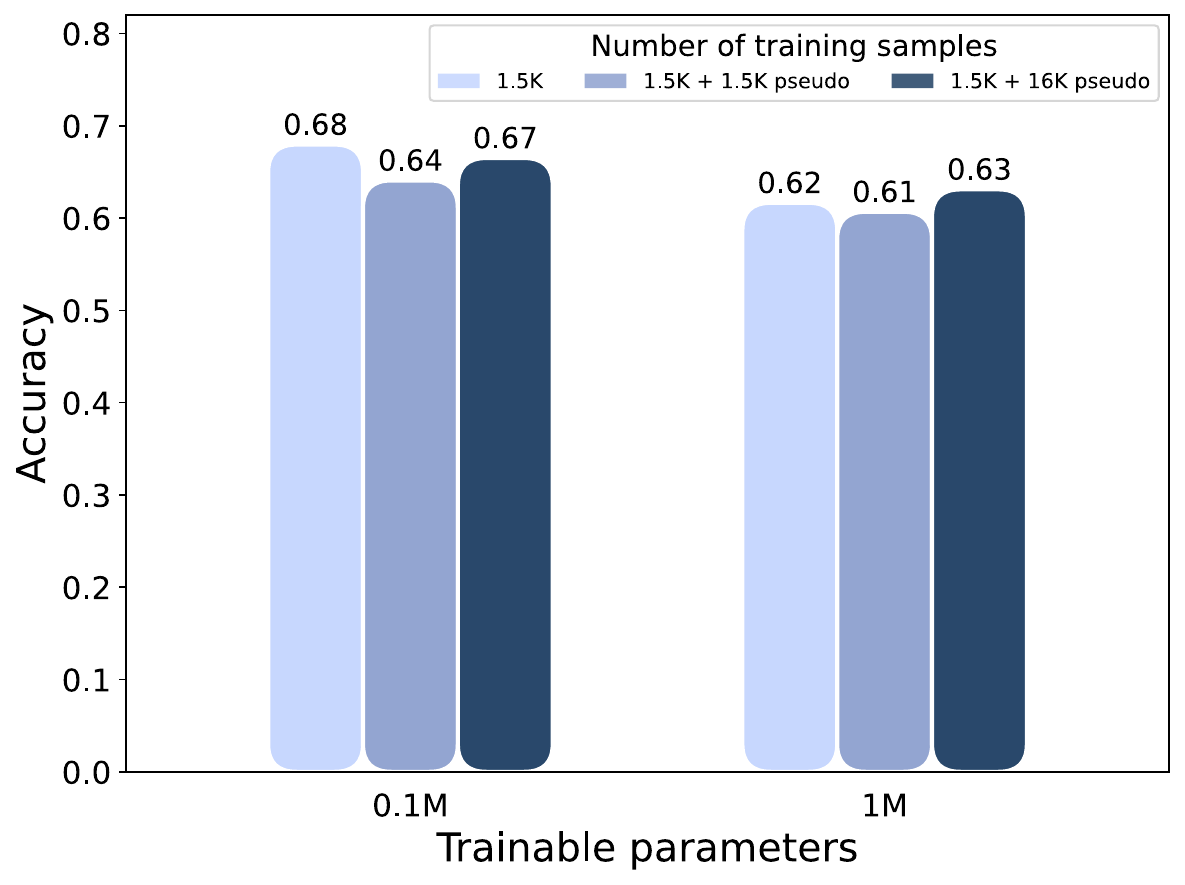}
        \caption{Validation split accuracies for different number of training samples and trainable parameters.}
        \label{fig:offline_anticipation_num_samples_val}
    \end{subfigure}
    
    \vspace{0.1cm}
    
    \caption{Results for the experiment of increasing the number of training samples via action spotting pseudo-labeling both on the a) test and b) validation splits. All experiments are done at $\tau_a=1.0$s.}
    \label{fig:offline_anticipation_increasing_number_samples}
\end{figure}

Figure~\ref{fig:offline_anticipation_increasing_number_samples} reports the results on both the validation and test splits for two different setups. In the first setup, training is conducted with the same number of trainable parameters as in Experiment~\ref{sec:changing t_a}, allowing for a fair comparison. In the second setup, since training with additional samples may require greater capacity to adequately capture the extra information, the experiments are repeated with the number of trainable parameters increased to 1M (approximately 26\% of the total parameters available in the X3D\_m model). The final training configuration remained largely consistent with that of Experiment~\ref{sec:changing t_a}, except for the aforementioned variations in the number of training samples and trainable parameters.

Figure~\ref{fig:offline_anticipation_num_samples_test} provides the main insight: increasing the number of training samples does not necessarily lead to better generalization. In the 0.1M setup, accuracy shows a slight improvement for the model trained with 3K sequences, but drops substantially when using 17.5K training samples. A similar pattern appears in the 1M setup, where accuracy marginally improves for the model trained with 17.5K samples, while it decreases for the model trained with 3K sequences. In parallel, the accuracy of the model trained exclusively with manually annotated samples remains unchanged across both setups. This irregular trend prevents concluding that adding more samples consistently benefits the model, as the observed variations may simply result from chance factors, such as converging to a slightly better checkpoint during training.

Before proceeding, it is important to highlight one major consideration regarding these results, which relates to the quality of the pseudo-labeled samples. Since the spotting model used to generate them was far from perfect, the detected rebounds in these pseudo-labels are not temporally precise. For instance, when trimming a video using an anticipation window of $\tau_a = 1.0$\,s, this window accurately captures the intended temporal context for manually annotated samples, but may not align correctly for pseudo-labeled samples. If the spotting model identifies the rebound 0.5s earlier than it actually occurs, the effective anticipation window becomes 1.5s, inadvertently increasing the difficulty of the task. Conversely, if the detection occurs after the actual rebound, the effective anticipation window becomes shorter than the stated value. Such temporal inconsistencies reduce the reliability of the pseudo-samples, as they may introduce noise or contradictory information, thereby limiting the model’s ability to learn meaningful patterns.

Finally, since it has been shown that the pseudo-samples, at least under the current setup, do not significantly enhance the learning capacity of the models, they will no longer be considered in the subsequent experiments. The remaining models will therefore be trained exclusively on the 1.5K manually annotated samples.

\begin{figure}[htbp]
    \centering
    \begin{subfigure}[t]{0.24\textwidth}
        \centering
        \includegraphics[width=\textwidth]{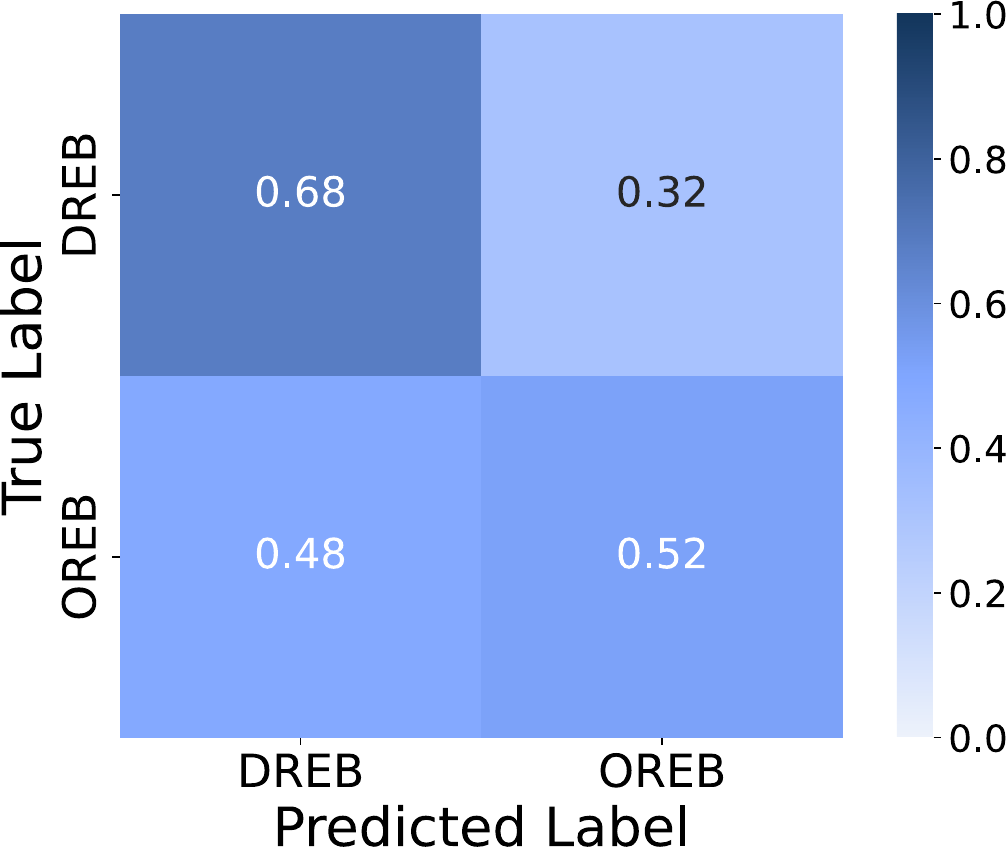}
        \caption{AI at $\tau_a=0.5$s.}
        \label{fig:CM_AI_0_5}
    \end{subfigure}
    \hfill
    \begin{subfigure}[t]{0.24\textwidth}
        \centering
        \includegraphics[width=\textwidth]{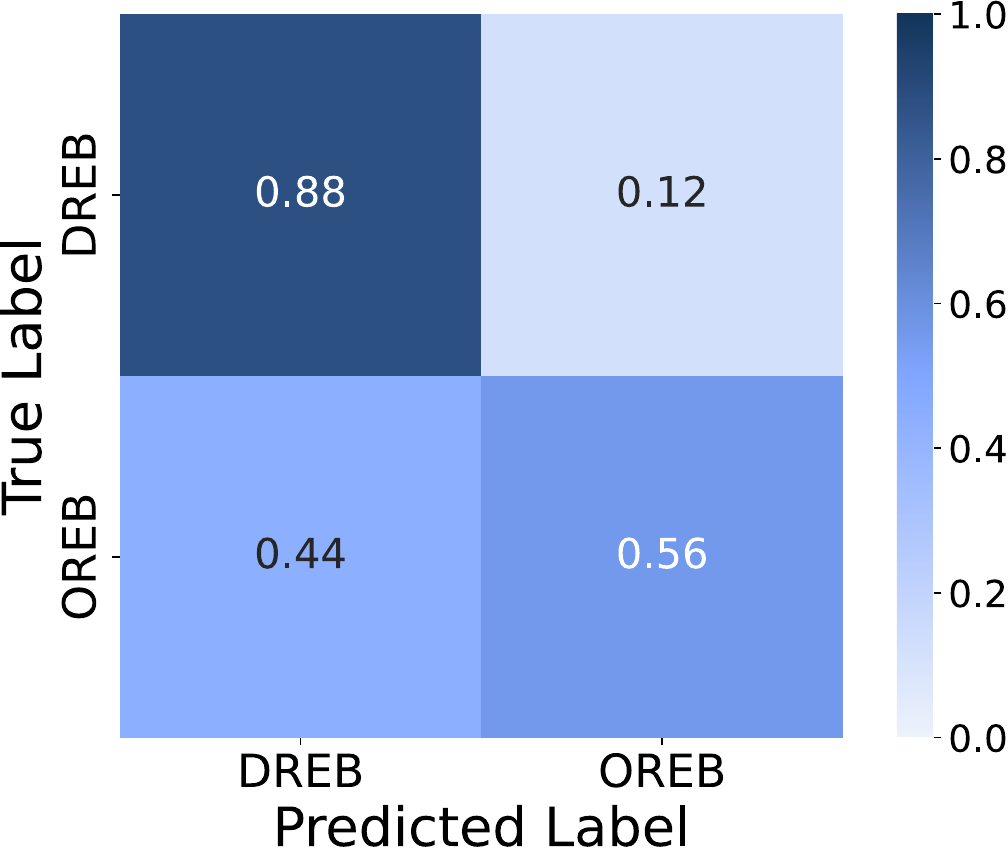}
        \caption{Humans at $\tau_a=0.5$s.}
        \label{fig:CM_HUMAN_0_5}
    \end{subfigure}
    \hfill
    \begin{subfigure}[t]{0.24\textwidth}
        \centering
        \includegraphics[width=\textwidth]{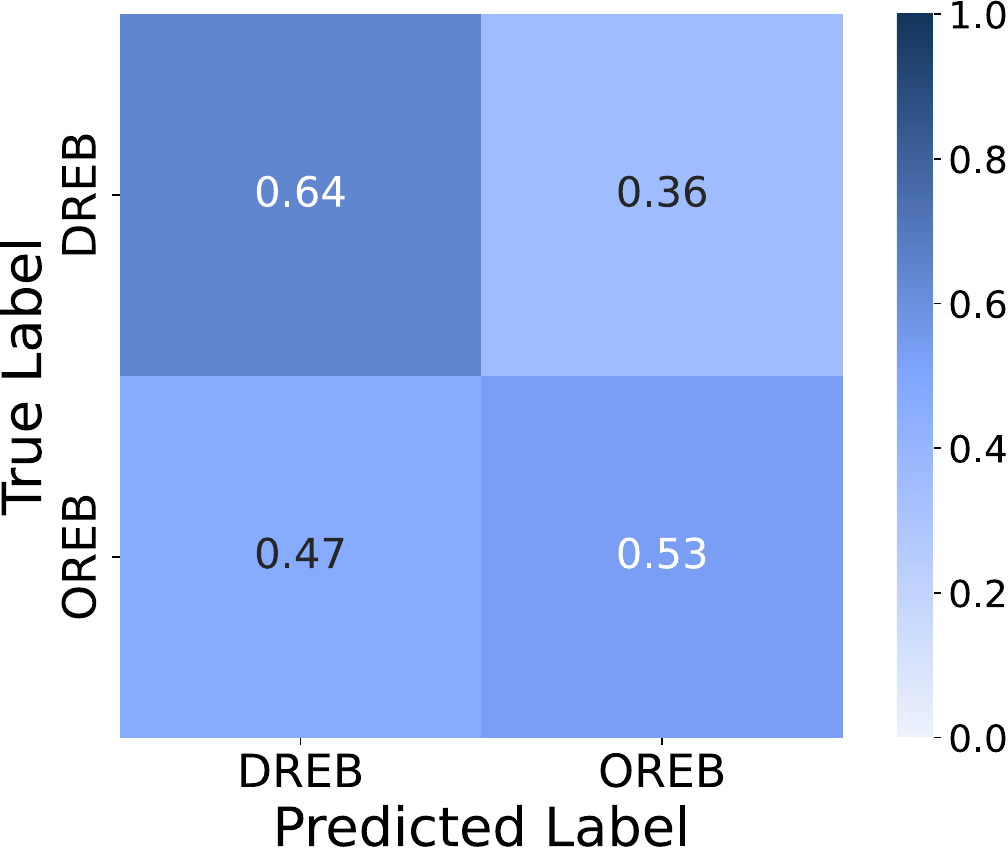}
        \caption{AI at $\tau_a=1.5$s.}
        \label{fig:CM_AI_1_5}
    \end{subfigure}
    \begin{subfigure}[t]{0.24\textwidth}
        \centering
        \includegraphics[width=\textwidth]{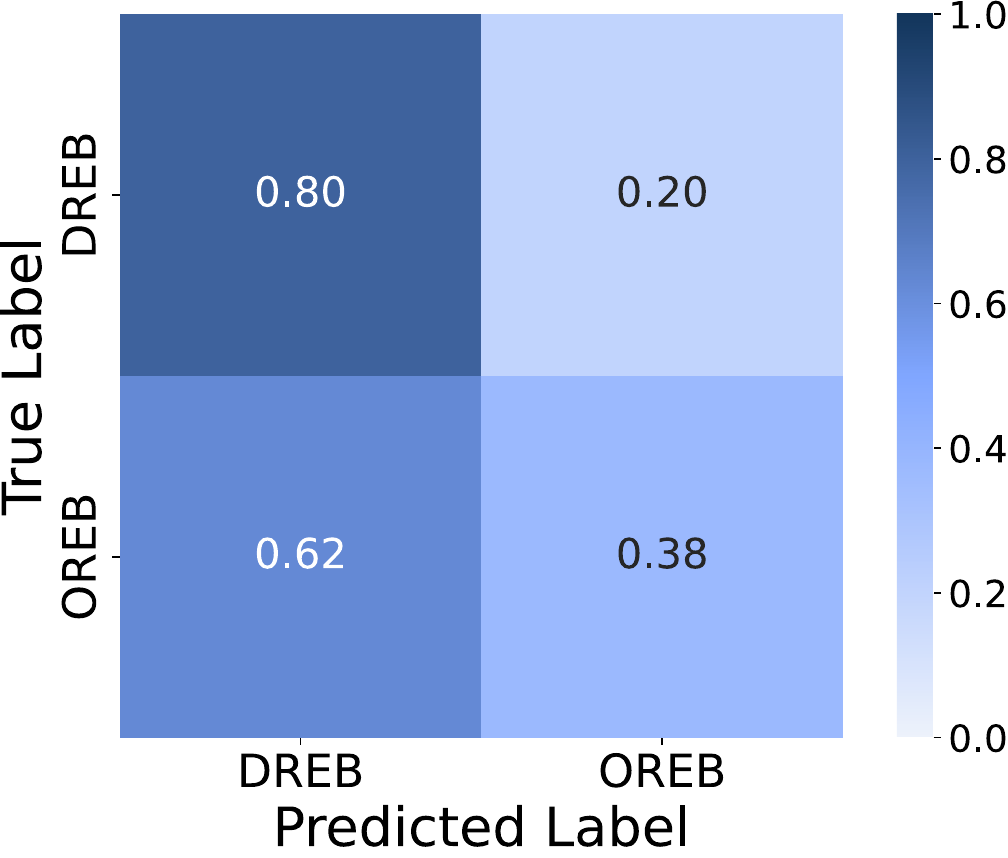}
        \caption{Humans at $\tau_a=1.5$s.}
        \label{fig:CM_HUMAN_1_5}
    \end{subfigure}
    
    \caption{Comparison between confusion matrices from the AI model, a) and c), vs Human, b) and d), at anticipation times $\tau_a=\{0.5, 1.5\}$s. Confusion matrices are normalized across the True Label axis (horizontally).}
    \label{fig:CMs_AI_vs_HUMAN}
\end{figure}

\subsubsection{Human experts vs AI}
The dataset used in this work is being studied for the first time, therfore there are no established expectations regarding what should be considered a ``good" or ``bad" performance. To contextualize the obtained results, the \textit{baseline} model performance is compared against that of a group of five human experts.

For the easier case of $\tau_a = 0.5$s (Figure~\ref{fig:CM_AI_0_5} and Figure~\ref{fig:CM_HUMAN_0_5}), the expected pattern is observed in the confusion matrices: the DREB class consistently outperforms the OREB class for both the AI model and the human evaluators. This outcome is not surprising, since defensive rebounds are generally easier to predict than offensive ones, the latter typically involving more players and greater variability. When comparing AI and human performance, humans show a clear advantage in predicting the DREB class. However, this difference nearly disappears in the case of OREB, where the accuracy is very similar, with only a 4\% higher correct prediction rate for humans.

The more challenging case corresponds to predictions made $\tau_a = 1.5$s before the rebound occurs (Figure~\ref{fig:CM_AI_1_5} and Figure~\ref{fig:CM_HUMAN_1_5}). In this setting, the superiority of DREB predictions over OREB remains evident for the AI model, but not for human predictions. Instead, humans display a strong bias toward the DREB class. This tendency can be explained by the natural distribution of rebounds in real games, where defensive rebounds occur approximately two to three times more frequently than offensive ones. Consequently, when little visual information is available, humans tend to rely on intuition and prior experience, defaulting to the most common outcome. When comparing AI and human performance under these conditions, humans still achieve higher accuracy in the DREB class, although this may be strongly influenced by their bias. In terms of overall prediction accuracy, both humans and AI perform at an identical level, with an accuracy of 0.59 for both of them. Furthermore, when precision and recall are considered, the AI demonstrates an ability to exploit visual cues more effectively than humans, particularly in situations where the decision is less intuitive. 

Table~\ref{tab:experiment_comparison} presents an extended comparison of classification metrics for the AI model and the group of human experts. At $\tau_a = 0.5$s, human experts significantly outperform the AI model. However, at $\tau_a = 1.5$s, the AI model achieves a slightly higher F1-score (0.56) compared to the humans (0.48), indicating more balanced predictions across classes when the anticipation task becomes more uncertain. These results highlight the limitations of human prediction patterns and demonstrate the AI model's advantages over human behavior: while humans excel in short-term anticipation scenarios, they become increasingly unbalanced in their predictions as $\tau_a$ increases, leading to class bias that deteriorates their F1-score. In contrast, the AI model maintains consistent and balanced decision-making across different anticipation windows, showing superior robustness under challenging conditions where visual cues are limited. This comparison demonstrates the promising potential of AI for basketball action anticipation. Even one of the simplest models capable of solving this task shows competitive performance relative to human experts. These results highlight the potential for developing AI tools to assist coaches, analysts, and players in understanding game dynamics and making more informed decisions.

\begin{table}[htbp]
\centering
\renewcommand{\arraystretch}{1.2}
\caption{Comparison of classification metrics for the four experiments.}
\begin{tabular}{lcccc}
\hline
\textbf{Experiment} & \textbf{Precision} & \textbf{Recall} & \textbf{F1-score} & \textbf{Accuracy} \\
\hline
AI at $\tau_a=0.5$s & 0.62 & 0.52 & 0.57 & 0.60 \\
Humans at $\tau_a=0.5$s & 0.83 & 0.56 & 0.67 & \textbf{0.71} \\
\hdashline
AI - Humans at $\tau_a=0.5$ & \textcolor{red}{-0.21} & \textcolor{red}{-0.04} & \textcolor{red}{-0.10} & \textcolor{red}{-0.11}\\
\hline
AI at $\tau_a=1.5$s & 0.60 & 0.53 & \textbf{0.56} & 0.59 \\
Humans at $\tau_a=1.5$s & 0.65 & 0.38 & 0.48 & 0.59 \\
\hdashline
AI - Humans at $\tau_a=1.5$ & \textcolor{red}{-0.05} & \textcolor{green}{+0.15} & \textcolor{green}{+0.08} & \textcolor{gray}{0.00}\\
\hline
\end{tabular}
\label{tab:experiment_comparison}
\end{table}

Finally, to complement the interpretability study presented in Section~\ref{sec:interpretability_experiment}, the experts were asked about the visual cues they relied on to make their predictions. Their answers were largely overlapping and included: 
(1) the location of the shot, since in general the farther the shot is taken, the stronger the ball bounces off the rim or backboard; 
(2) when the information is available in the video, they rely heavily on the trajectory of the ball after it bounces off the rim  (generally it is available when $\tau_A=0.5$s and not available when $\tau_a=1.5$s); 
(3) the number of offensive players inside the three-point line, as players positioned outside this area are very unlikely to secure a rebound;
(4) whether these offensive players are being boxed out by a defender\footnote{Boxing out is a defensive technique in basketball where a player positions themselves between the opponent and the basket, using their body to prevent the opponent from gaining access to the rebound. In case the explanation isn't clear, please find attached a video example \href{https://youtu.be/KohsPS7Etyc}{here}.}, since effective boxing out drastically reduces the chances of obtaining a rebound; and
(5) whether the offensive player has momentum toward the rim, as a player driving with momentum typically outperforms a stationary defender who only jumps vertically for the ball.

Among the visual cues used by the experts and the AI model, the only significant difference lies in the use of the “ball tracking” cue. Both the model and the human experts tend to focus strongly on the players inside the three-point line (i.e., those closer to the rim), which is reasonable since most rebounds occur in that region. In addition, the model exhibits a form of tracking or heightened activation on the player taking the shot, which may be related to humans considering the shot location. However, the model does not appear to focus on the ball after the shot is released or when it bounces off the rim, whereas human experts reported relying heavily on this information when available. This difference may explain why the model performs significantly worse than humans at $\tau_a = 0.5$s, when the ball bounce off the rim is visible in most videos. Humans exploit this information, while the model appears to rely solely on player positioning. Conversely, when the ball bounce information is not available (e.g., $\tau_a = 1.5$s), human performance drops significantly and becomes comparable to that of the AI model, as both rely primarily on player positioning, speed, and momentum.

Based on these results, a potential improvement could be to explicitly encourage the model to track the ball, for example, by introducing an auxiliary loss. Such a modification would introduce an inductive bias toward ball-awareness, potentially enhancing the model’s anticipation capabilities. This idea will be further discussed in Section~\ref{sec:limitations}.

\subsection{Online Action Anticipation}
In this section the results for the experiments on the online anticipation setup will be presented for both the \textit{baseline} and \textit{TEAM} models. 

\subsubsection{Changing the clip length}
\begin{figure}[htbp]
    \centering
    \begin{subfigure}[t]{0.47\textwidth}
        \centering
        \includegraphics[width=\textwidth]{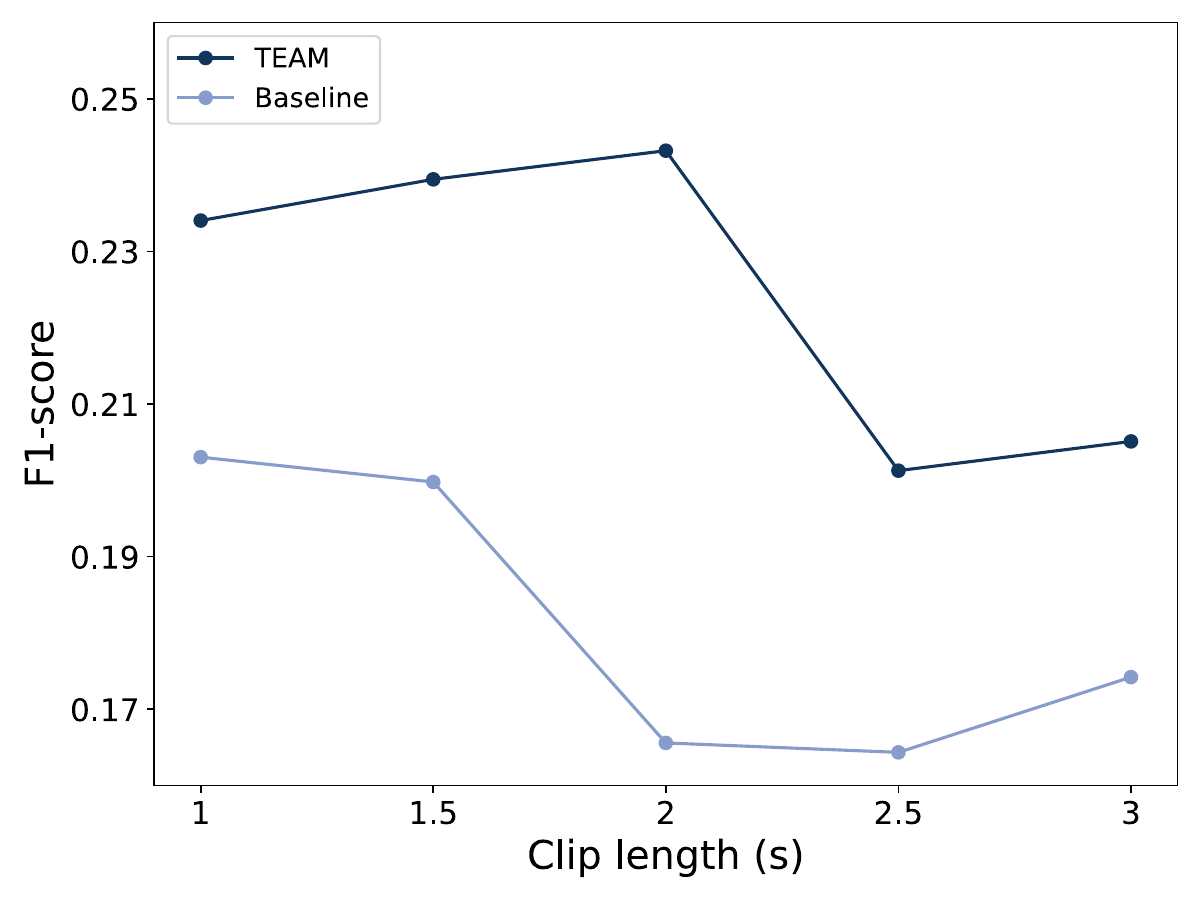}
        \caption{Comparison of validation F1-scores at different clip lengths for the \textit{baseline} (light blue) and \textit{TEAM} (dark blue) models.}
        \label{fig:online_anticipation_clip_len_val}
    \end{subfigure}%
    \hfill
    \begin{subfigure}[t]{0.47\textwidth}
        \centering
        \includegraphics[width=\textwidth]{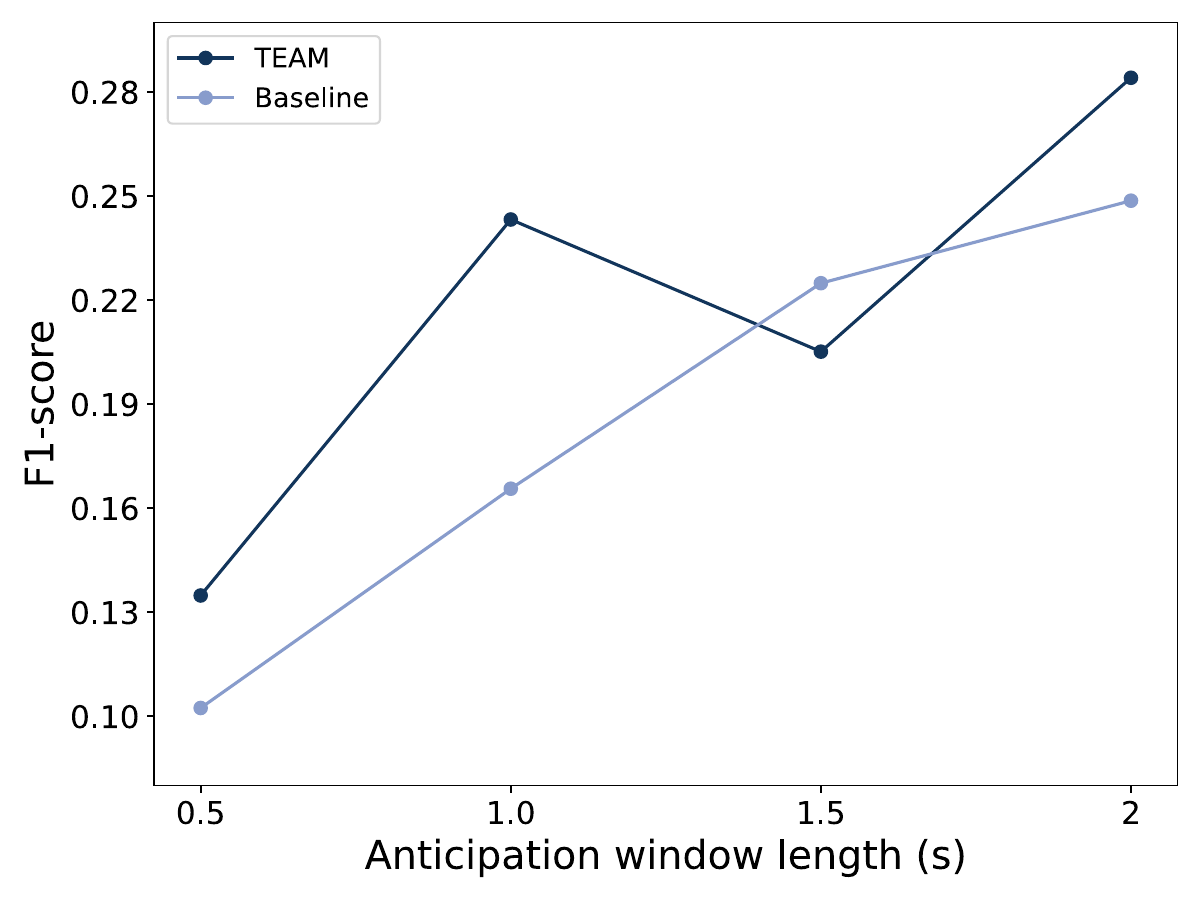}
        \caption{Comparison of validation F1-scores at different $AW$ lengths for the \textit{baseline} (light blue) and \textit{TEAM} (dark blue) models.}
        \label{fig:online_anticipation_AW_val}
    \end{subfigure}
    
    \vspace{0.1cm}
    
    \caption{Results of the experiments varying (a) the clip length and (b) the anticipation window length in the online anticipation setup, shown on the validation split.}
    \label{fig:online_anticipation_ablations_clip_and_AW_lens}
\end{figure}

The first experiment evaluated the effect of increasing or decreasing the temporal context while keeping a fixed anticipation window at 1.0s, considering both the \textit{baseline} and \textit{TEAM} models. The results are shown in Figure~\ref{fig:online_anticipation_clip_len_val}.

The first observation is that both models exhibit similar behavior, with \textit{TEAM} consistently outperforming the \textit{baseline} across all experiments. This result is expected, as \textit{TEAM} builds upon the baseline's backbone and further incorporates a temporal module, adding more parameters and enabling it to better capture long-range temporal dependencies.

Regarding the specific focus of this experiment — varying the amount of temporal context provided to the model — the initial hypothesis stated in Section \ref{sec:experiments} does not fully hold: a longer temporal context does not necessarily lead to improved performance, and, in general, the results suggest a decreasing trend. For the baseline model, the trend is clear: the larger the clip length, the lower the performance. The only exception is the last experiment (with clip length = $3$s), which slightly outperforms the setups with clip lengths of 2.0s and 2.5s. Additionally, there is a pronounced inflection point between 1.5s and 2s, where the drop in performance is noticeably larger than in other intervals. In contrast, for \textit{TEAM}, we observe a slight upward trend for the first three experiments (clip lenght 1.0, 1.5 and 2.0s, with performance peaking at a clip length of 2.0s. After this maximum, performance drops significantly for the longer clip lengths.

\subsubsection{Changing the anticipation window length}

For the second experiment, the clip length is fixed at $2$s (since it was the best-performing value for \textit{TEAM} on the validation set), while the anticipation window is varied. The results are shown in Figure~\ref{fig:online_anticipation_AW_val}.

In this case, the expected behavior described in the initial hypothesis is observed: the larger the anticipation window, the higher the performance for both models (except for a slight drop in \textit{TEAM} from 1.0s to 1.5s.

Comparing both models, \textit{TEAM} outperforms the \textit{baseline} in nearly all experiments. However, as the anticipation window increases, the performance gap narrows; in some cases ($AW=1.5$s) the \textit{baseline} model even slightly surpasses \textit{TEAM}. This can be partly explained by the task becoming easier with longer $AW$, as the clip is more likely to contain an action (OREB or DREB), which alleviates the severe class imbalance with the negative (no-rebound) clips. In contrast, for shorter $AW$ lengths, the class imbalance is stronger, making it harder for the simpler \textit{baseline} — with fewer parameters — to optimize effectively. For longer $AW$, the added complexity of the transformer becomes less necessary, and the backbone alone can capture the temporal dynamics, allowing the \textit{baseline} to perform relatively well.

\subsubsection{Complementary ablations}
Finally, Table~\ref{tab:final_ablations} reports ablation studies on hyperparameters related to the task, the model configuration, and the overall architecture. The base parameters used for these experiments were presented in Section~\ref{sec:TEAM_online_training_config}.

For the task-related ablations, the optimal stride is observed to be around 3. Increasing the stride beyond this value results in poorer performance, as the reduction in temporal resolution may lead to the loss of fine-grained details. Conversely, reducing the stride to 2 produces slightly better results than larger strides but still worse than the optimal value. Moreover, a smaller stride increases the number of frames to be processed, leading to higher computational overhead. Regarding the overlap parameter, results on the test set show an upward trend when the overlap is increased. This behavior may be explained by the fact that denser clip sampling introduces greater variability in the data, since the anticipated actions can occur at different positions within the anticipation window, thereby potentially improving the model’s generalization capability.

The second set of experiments ablates different settings of \textit{TEAM}. In experiment (c), it is observed that increasing the input resolution of the frames improves performance. However, higher resolution also leads to increased memory usage and longer computation times, which is why a lower resolution that balances efficiency and performance was chosen as the baseline. Experiment (d) evaluates the impact of varying the number of MHSA heads in the transformer-encoder layer of \textit{TEAM}. The chosen baseline corresponds to the optimal setting on the validation set, although in the test set the highest number of heads (12) yields the best performance. The third experiment (e) investigates the effect of changing the number of transformer-encoder layers. A mismatch between validation and test results is observed: while two encoder layers yield the best performance on the test set by a significant margin, three encoder layers slightly outperform this configuration on the validation set.

The final set of ablations concerns the overall architecture of the method. Experiment (f) highlights the benefits of initializing the backbone with weights pre-trained on a task closely related to the downstream objective (rebound classification $\rightarrow$ rebound anticipation) using the same dataset, as opposed to the default weights pre-trained on a general-purpose dataset such as Kinetics-400. Experiment (g) compares the performance of the model with two different backbones: X3D\_m (3D CNN) and ConvNeXt\_Tiny (2D CNN). The results are nearly identical; however, it is important to note that X3D\_m contains only $3.79$M parameters, whereas ConvNeXt\_Tiny has $28.5$M, a considerable difference. This illustrates the efficiency and strong performance of X3D\_M despite its significantly smaller size. Finally, experiment (h) evaluates alternative temporal modules. The transformer-encoder module achieves substantially better performance compared to both the 3D CNN and the LSTM. The 3D CNN model corresponds to the \textit{baseline} X3D\_m, the transformer-based model follows the \textit{TEAM}implementation detailed in Section~\ref{sec:TEAM_online_training_config}, and the LSTM-based variant builds upon the X3D backbone (exactly the same used in \textit{TEAM}) by adding two unidirectional LSTM layers with a hidden state dimension of 512. The resulting hidden representation of the last layer is then fed into the final classification head, taking inspiration from~\cite{furnari2020rolling}.

Finally, it should be explicitly noted that it is no coincidence that the best-performing parameters align with the default settings established for the task and the model's hyperparameters. Preliminary experiments were conducted to identify the optimal hyperparameters, and subsequent experiments were performed to assess how variations in specific parameters affected performance.

\begin{table}[t]
  \centering
  \begin{minipage}[t]{0.48\linewidth}
    \centering
    \vspace{0pt}
    \begin{tabular}{llccc}
      \toprule
      \multicolumn{2}{l}{\textbf{Experiment}} & \multicolumn{3}{c}{\textbf{Validation F1-score}} \\
      \cmidrule(lr){3-5}
      \multicolumn{2}{l}{} & OREB & DREB & Avg. \\
      \midrule
      (a) & stride=2   & 0.305 & 0.247 & 0.276 \\
          & stride=3   & 0.290 & 0.284 & \textbf{0.287} \\
          & stride=6   & 0.236 & 0.232 & 0.234 \\
      \midrule
      (b) & overlap=0   & 0.200 & 0.275 & 0.238 \\
          & overlap=0.5 & 0.290 & 0.284 & \textbf{0.287} \\
          & overlap=0.7 & 0.286 & 0.271 & 0.279 \\
      \midrule
      (c) & resolution=128$\times$225   & 0.266 & 0.271 & 0.269 \\
          & resolution=256$\times$455   & 0.290 & 0.284 & \textbf{0.287} \\
      \midrule
      (d) & MHSA heads=4  & 0.252 & 0.289 & 0.271 \\
          & MHSA heads=8  & 0.290 & 0.284 & \textbf{0.287} \\
          & MHSA heads=12 & 0.296 & 0.254 & 0.275 \\
      \bottomrule
    \end{tabular}
  \end{minipage}%
  \hfill
  \begin{minipage}[t]{0.48\linewidth}
    \centering
    \vspace{0pt}
    \begin{tabular}{llccc}
      \toprule
      \multicolumn{2}{l}{\textbf{Experiment}} & \multicolumn{3}{c}{\textbf{Validation F1-score}} \\
      \cmidrule(lr){3-5}
      \multicolumn{2}{l}{} & OREB & DREB & Avg. \\
      \midrule
      (e) & encoder layers=1 & 0.221 & 0.236 & 0.229 \\ 
          & encoder layers=2 & 0.290 & 0.284 & \textbf{0.287} \\ 
          & encoder layers=3 & 0.265 & 0.252 & 0.259 \\
      \midrule
      (f) & pretrain=Kinetics       & 0.289 & 0.253 & 0.271 \\
          & pretrain=Classification & 0.290 & 0.284 & \textbf{0.287} \\
      \midrule
      (g) & backbone=X3D             & 0.290 & 0.284 & \textbf{0.287} \\
          & backbone=ConvNeXt\_Tiny  & 0.277 & 0.292 & 0.285 \\
      \midrule
      (h) & temporal=X3D       & 0.198 & 0.205 & 0.202 \\
          & temporal=Transformer  & 0.290 & 0.284 & \textbf{0.287} \\
          & temporal=LSTM         & 0.154 & 0.184 & 0.169 \\
      \bottomrule
    \end{tabular}
  \end{minipage}
  \caption{Complementary ablation study on task settings: (a) stride and (b) overlap; \textit{TEAM}'s settings: (c) input frame resolution, (d) number of MHSA heads, and (e) number of transformer-encoder layers; and architecture settings: (f) backbone pre-training dataset, (g) backbone, and (h) temporal module choice.}
  \label{tab:final_ablations}
\end{table}

%
%

\section{Limitations and future work} \label{sec:limitations}

In this section, the main limitations of the present work are discussed, together with potential directions for future research. These include improvements to the dataset, generalization of the tasks, as well as architectural and methodological enhancements.

\subsection{Annotated data for action spotting and anticipation}

Having more, and high-quality, data is always beneficial for deep learning models. In this work, the limited number of annotated samples for rebound spotting and rebound anticipation resulted in relatively small validation and test sets, leading to a relatively high variability in the results and making comparisons between them less reliable. Nevertheless, no major signs of overfitting were observed during training, suggesting that the amount of training data was sufficient. Therefore, only a moderate increase in annotated samples may be needed, primarily to stabilize evaluation on the validation and test sets.

Future work on this dataset and the associated tasks could focus on refining the semi-automatic methods used for pseudo-labeling action timestamps. First, state-of-the-art methods for action spotting could be employed instead of the simple baseline X3D used in this work. Second, more sophisticated training strategies could be explored, such as incrementally expanding the training dataset with the most confident pseudo-labeled samples, assigning them a lower weight in the loss function, rather than relying solely on a limited set of ground-truth annotations followed by inference.

\subsection{Number of classes considered}

Another important limitation of this work lies in the type of actions included in the dataset, which is restricted to only two categories of rebounds. The \href{https://github.com/arnalytics/labeled_plays_NBA}{repository} associated with this work provides code to scrape the NBA Stats website~\cite{nba_stats}, enabling the collection of up to — at least — 8 different types of actions: \textit{Off. rebound}, \textit{Def. rebound}, \textit{2-point shot}, \textit{3-point shot}, \textit{Assist}, \textit{Turnover}, \textit{Steal}, and \textit{Block}. For the purposes of this thesis, only two classes were considered in order to maintain the problem tractable, as anticipation is already a highly challenging task without further increasing the number of classes. Nevertheless, methods intended for real-world or industry applications should be more robust and able to anticipate a wider variety of actions beyond rebounds. Despite this restriction, the present work provides a solid starting point, and the extension of the approach to additional action categories is expected to be feasible without major modifications.

\subsection{Methodological improvements}

\subsubsection{Incorporating a \textit{time-to-event} auxiliary loss}
In the online action anticipation setting, the time between the end of a given clip and the subsequent action to be anticipated (\textit{time-to-event}) is variable. Incorporating an additional time-to-event prediction, optimized through an auxiliary loss, could provide complementary supervision and help the model better address the main online anticipation task.

\subsubsection{Incorporating a ball-tracking auxiliary loss}
The interpretability study revealed that the model does not focus strongly to the ball trajectory or bounces when producing its predictions. Although this was shown in the offline setting, a similar behavior is expected in the online anticipation scenario. In contrast, human observers consistently relied on the ball trajectory, which was associated with better performance, particularly in the simpler setup with an anticipation time of $t_a = 0.5$s. Incorporating an explicit ball-tracking auxiliary loss, where the model is required to predict the bounding box of the ball for each frame, would introduce an inductive bias toward attending to the ball's trajectory and could thereby improve predictive performance in line with human behavior.

\subsubsection{Architecture – Modeling uncertainty in the prediction head}

Finally, a key avenue for improvement lies in the prediction head. The backbone and the temporal module closely mirror components used in related video understanding tasks and offer limited room for further gains, whereas the anticipation-specific head can be refined. Prior work \cite{zhong2023diffant, bball_predictions} explicitly models the inherent uncertainty of forecasting with diffusion models or other non-deterministic formulations. Replacing the simple MLP head with a module capable of capturing such non-deterministic dynamics would constitute a significant improvement.

%
%

\section{Conclusions} \label{sec:conclusions}
This work introduces a novel dataset in the domain of basketball, addressing the lack of existing datasets in this area. The dataset is unique in that it focuses on two previously unexplored types of actions: offensive and defensive rebounds. In addition to the 100,000 videos labeled with the action occurring in the video, 2,000 manual annotations were provided to support a range of video understanding tasks, as demonstrated in this study, including action classification, action spotting, and action anticipation.

Regarding the main results obtained, the offline anticipation task addressed yielded insightful outcomes. The simple baseline model achieved performance comparable to human anticipation for times around $\sim$1.5s, motivating further research in this field. If a model attains strong forecasting performance — potentially surpassing that of human experts — analyzing the regions of the video that the model focuses on during prediction may reveal useful patterns. These patterns could be used to develop post-game analysis tools that coaches and trainers can leverage to improve player performance and gain deeper insights into game dynamics. Moreover, this approach can be extended beyond rebounds to other types of actions, such as shots or passes, and can even be applied for action valuation or performance assessment.

After analyzing the visual cues attended by the model, it was observed that, in cases where humans clearly outperformed the AI, some of these cues resembled those used by humans, such as the location of the shot and the positioning and momentum of off-ball players. A key discrepancy, however, was that the model did not explicitly track the ball after the shot. This finding offers valuable guidance for improving future anticipation models, particularly when applied to this dataset, by informing both architectural design and training strategies to better align with human-like behavior.

For the online anticipation setup, the most extensively studied in the literature, benchmark results were provided for the current state-of-the-art method in online action anticipation within multi-person sports scenarios, \textit{FAANTRA}~\cite{dalal2025fantra}. Additionally, ablation studies were conducted on various task-related parameters — highlighting optimal configurations — as well as on alternative architectural choices. In particular, different backbones (X3D and ConvNeXt\_tiny) and temporal encoders (Transformer and RNN) were evaluated, providing a more comprehensive analysis of the factors influencing model performance.

Finally, potential improvements to the proposed method and dataset were outlined, specially the auxiliary ball-tracking loss, which could inform future research directions.

%
%



\newpage
%
%
\section*{Appendix}\label{sec:appendix}

\end{document}